\documentclass[11pt,a4paper]{article}
\usepackage[margin=2.3cm]{geometry}
\usepackage[T1]{fontenc}
\usepackage[utf8]{inputenc}
\usepackage{mathptmx}
\usepackage{microtype}
\usepackage{amsmath,amssymb}
\usepackage{graphicx}
\usepackage{booktabs}
\usepackage{multirow}
\usepackage{array}
\usepackage{longtable}
\usepackage{caption}
\usepackage{subcaption}
\usepackage{siunitx}
\usepackage{authblk}
\usepackage[numbers,sort&compress]{natbib}
\usepackage{xcolor}
\usepackage{lineno}
\usepackage[hidelinks]{hyperref}
\usepackage{setspace}
\usepackage{pdflscape}
\usepackage{enumitem}
\usepackage{adjustbox}

\graphicspath{{figures/}}
\newcommand{\degree}{\ensuremath{^\circ}}

\title{\textbf{Satellite-derived bathymetry of Arabian Gulf ports under leakage-free validation: a multi-site benchmark of empirical, machine-learning and deep-learning models against 285,000 multibeam soundings, with an outlook for UAE waters}}

\author[1]{Muhammad Saeed\thanks{Corresponding author. E-mail: muhammad.saeed@adports.ae}}
\author[1]{Mohamed H. Alhosani}
\author[1]{Dmitry Mikhaylov}
\author[1]{Yasser F. Al Wahedi}
\affil[1]{Abu Dhabi Maritime Academy, P.O. Box 54477, Abu Dhabi, United Arab Emirates}
\date{Preprint, \today}

\begin{document}
\maketitle

\begin{abstract}
\noindent
Satellite-derived bathymetry (SDB) is often proposed as a low-cost complement to multibeam surveys of ports, but its accuracy is seldom measured the way the method is used, on an area the model has never seen. This study tests SDB in the turbid, shallow and tidal waters of the Abu Dhabi coast against 285,619 multibeam soundings from twelve surveys at six port areas, reduced to Lowest Astronomical Tide, and then applies the same chain to the whole coast of the United Arab Emirates. Two empirical methods (the Stumpf ratio and the Lyzenga transform), a gradient-boosted tree model with monotonic constraints (LightGBM) and a U-Net ensemble were trained on 72 Sentinel-2 images and tested in transfer to an unseen site and with local control, in which each image is calibrated with about 20\,\% of the site's soundings and tested on spatial blocks beyond a 225\,m buffer. The validation design changed the result more than the model: the same LightGBM model gave 0.57\,m RMSE under a random pixel split, 1.30\,m under spatial blocks and 1.66\,m at an unseen site. In transfer the Stumpf ratio (1.45\,m) was not beaten by LightGBM (1.66\,m), the U-Net (1.77\,m) or Lyzenga (2.00\,m). With local control a hybrid of the tree output and the band ratio reached 1.11\,m over 0--11\,m, 67\,\% of pixels within the IHO S-44 Order~2 tolerance and a conformal 95\,\% interval with 95\,\% observed coverage; in clear water it reached 0.29--0.53\,m. The seabed could not be read beyond 11\,m, and summer images were five to six times more turbid than winter ones. ICESat-2 ATL24 agreed with the surveys to 0.25\,m after an offset of $+0.28$\,m $+3.5\,\%$ of depth was removed. Along the UAE coast a 60\,m reconnaissance mosaic of 17,200\,km$^2$, checked against 110,000 ICESat-2 pixels, showed the same error pattern with depth as the survey sites (2.9\,m RMSE, 3.4\,m too deep on the flats); calibrated on the lidar cell by cell, with a held-out block test, it reached 1.42\,m with a bias of $+0.2$\,m over half of the product, without any vessel survey. Sentinel-2 SDB with about a hundred control depths per image is a usable reconnaissance and change-monitoring tool in these waters to about 10\,m; it does not meet S-44 Order~1 without local control, and accuracies reported from random splits should not be taken as operational accuracy.
\end{abstract}


\section*{Nomenclature}
\begin{small}
\begin{adjustbox}{max width=\textwidth}
\begin{tabular}{@{}llll@{}}
\toprule
Symbol & Meaning & Symbol & Meaning \\
\midrule
$\rho_i$ & surface reflectance of band $i$ & $d$ & depth below LAT (m) \\
$\rho_{\infty,i}$ & deep-water reflectance of band $i$ & $h$ & water column at overpass, $h = d + \eta$ (m) \\
$X_i = \ln(\rho_i - \rho_{\infty,i})$ & Lyzenga transform & $\eta$ & tide height above LAT at overpass (m) \\
$p_{ij} = \ln(1000\rho_i)/\ln(1000\rho_j)$ & Stumpf log-ratio & $\hat d$ & satellite-derived depth (m) \\
$K_d$ & diffuse attenuation coefficient (m$^{-1}$) & U95 & 95\,\% uncertainty of $\hat d$ (m) \\
$\beta_i$ & glint slope of band $i$ against B11 & TVU & S-44 total vertical uncertainty (m) \\
$\sigma_b$ & between-image standard deviation (m) & $a, b$ & S-44 TVU coefficients \\
$\sigma_\infty$ & spread of the deep-water set (green band) & $\kappa$ & ratio value of the shallow-water hinge \\
$\sigma_n$ & sensor noise estimate (green band) & $\gamma_0$ & floor of the conformal score (m) \\
SNR & seabed signal-to-noise ratio & $q_{95}$ & conformal 95\,\% quantile \\
$S$ & scene score of an image & $d_{\max}$ & transfer depth limit (11\,m) \\
$w_s$ & fusion weight of image $s$ & $e_k$ & error of pixel $k$, $\hat d_k - d_k$ (m) \\
\midrule
\multicolumn{4}{@{}l@{}}{\textit{Abbreviations:} ADMA, Abu Dhabi Maritime Academy; ATL24, ICESat-2 bathymetry product; BOA, bottom of atmosphere; CV, cross-validation;}\\
\multicolumn{4}{@{}l@{}}{EOT20, Empirical Ocean Tide model 2020; GBDT, gradient-boosted decision trees; IHO, International Hydrographic Organization; LAT, Lowest}\\
\multicolumn{4}{@{}l@{}}{Astronomical Tide; LOSO, leave-one-site-out; MAE, mean absolute error; MBES, multibeam echo sounder; NDTI, normalised difference}\\
\multicolumn{4}{@{}l@{}}{turbidity index; NDWI, normalised difference water index; NMAD, normalised median absolute deviation; RMSE, root mean square error;}\\
\multicolumn{4}{@{}l@{}}{S-44, IHO Standards for Hydrographic Surveys (6th ed.); SCL, scene classification layer; SDB, satellite-derived bathymetry; SHAP,}\\
\multicolumn{4}{@{}l@{}}{Shapley additive explanations; SWIR, short-wave infrared; VHR, very high resolution.}\\
\bottomrule
\end{tabular}
\end{adjustbox}
\end{small}

\section{Introduction}
\label{sec:intro}

Reliable knowledge of water depth in ports, approach channels and the surrounding shallow water is a prerequisite for safe navigation, dredging management, coastal construction and the maintenance of nautical charts. Multibeam echo-sounder (MBES) surveys provide this knowledge at the accuracy demanded by the International Hydrographic Organization (IHO) Standards for Hydrographic Surveys, S-44 \cite{IHO2020}, but they are expensive, vessel-bound and slow, so that port authorities typically re-survey a given area only every few years. Between surveys, the depth of dredged basins and the seabed of natural shoals may change by amounts that are relevant to under-keel clearance, and no comparably accurate information is available. Satellite-derived bathymetry (SDB), which infers water depth from the colour of the water in optical satellite imagery, has therefore been studied for decades as a cheap and repeatable complement to acoustic surveying \cite{Gao2009,Jawak2015,Ashphaq2021,Kutser2020}.

The physical basis of optical SDB is well established. The upwelling radiance above shallow water contains a seabed contribution that decays approximately exponentially with depth at a rate set by the diffuse attenuation coefficient of the water column, $K_d$, which differs between spectral bands \cite{Lyzenga1978,Philpot1989}. Two empirical formulations exploit this relation. The linear multi-band transform of Lyzenga \cite{Lyzenga1978,Lyzenga1985,Lyzenga2006} regresses depth on the logarithms of the deep-water-corrected reflectances, whereas the ratio transform of Stumpf et al.\ \cite{Stumpf2003} regresses depth on the ratio of the logarithms of two bands, which cancels much of the variation in bottom albedo at the cost of a less direct physical interpretation. Both methods require calibration against known depths in the image, and both degrade in turbid water, where suspended sediment both brightens the water column and shortens the depth to which the seabed remains visible \cite{Caballero2020,Casal2020}. The availability of the Sentinel-2 constellation, with 10\,m pixels in the blue, green and red bands and a revisit of five days, has made multi-temporal SDB practicable at regional scale \cite{Caballero2019,Traganos2018,Li2021}, and machine-learning regressors (random forests, support vector machines, gradient-boosted trees and, more recently, convolutional neural networks) have been reported to outperform the empirical transforms when abundant reference depths are available \cite{Sagawa2019,Mudiyanselage2022,LumbanGaol2021}. In parallel, the photon-counting lidar of ICESat-2 has provided a spaceborne source of reference depths \cite{Parrish2019,Babbel2021}, which has been combined with Sentinel-2 to produce SDB without in-situ soundings \cite{Ma2020,Thomas2021}, and NASA's ATL24 product now distributes classified seabed photons globally.

Three problems stand between the accuracies published for SDB and their use in port surveying. The first is the validation design. Most studies report accuracies from a random split of the available reference depths into training and testing subsets. Because neighbouring pixels share both depth and colour, a random split places a training pixel within a few tens of metres of nearly every test pixel, and the resulting error reflects interpolation rather than prediction \cite{Roberts2017,Ploton2020}. The error that matters operationally, namely that of a model applied to an area or to an image on which it was not calibrated, is rarely reported. The second is the environment. The Arabian Gulf is shallow, hypersaline, strongly tidal in its southern embayments and, along the Abu Dhabi coast, frequently turbid because of resuspension by the shamal wind, dredging and coastal construction \cite{Sheppard2010,Vaughan2019}. Dredged port basins reach 18--21\,m in water that is optically deep to a passive sensor, so that an SDB model must not only estimate depth but also recognise where no estimate is possible. Published SDB results for the Gulf are few, and none, to the authors' knowledge, has been validated against dense MBES surveys at several ports under a leakage-free design. The third is that models and sensors are rarely compared on equal terms. Comparisons of empirical, machine-learning and deep-learning SDB models are usually made at one site, with one image, on one random split (a recent exception, for coral-reef waters, is the cross-regional comparison of \cite{Hsu2026}), which does not reveal whether a more complex model transfers better, whether it benefits from very-high-resolution (VHR) imagery, how many control soundings it needs, how its accuracy changes with the season of acquisition, or whether two maps of the same area made a year apart agree.

The present study addresses these issues with a benchmark built on the hydrographic survey archive of the Abu Dhabi ports. Twelve MBES survey datasets from six port areas on the Abu Dhabi coast, totalling 285,619 soundings referenced to Lowest Astronomical Tide (LAT), were paired with 72 Sentinel-2 L2A images, one Pl\'eiades-1A image and 97 ICESat-2 ATL24 granules. Four model families were compared: the Stumpf ratio, the Lyzenga transform, a gradient-boosted decision-tree model (LightGBM) with monotonic constraints that encode the attenuation physics, and an ensemble of U-Nets. Each model was evaluated in two modes that correspond to two operational situations: a \emph{transfer} mode, in which the model trained at the other sites is applied to an unseen site, and a \emph{local-control} mode, in which every image is calibrated with a small subset of the site's own soundings, as in hydrographic SDB practice \cite{Mavraeidopoulos2017}. In both modes the test pixels are separated from any calibration pixel by spatial blocks sized from the variogram of the residuals and by a buffer, two sites were held back as fully independent tests, and a third survey from a different year tested temporal stability. The same framework was then used to quantify the effect of the validation design itself, the achievable depth and its relation to turbidity, the seasonal variation of accuracy and coverage, the number of fused images and of control soundings required, the effect of pixel size from 2 to 60\,m, the agreement between ICESat-2 lidar and MBES, and the possibility of calibrating with lidar alone. Every depth carries a 95\,\% uncertainty obtained by split-conformal prediction and is graded against the S-44 total vertical uncertainty. The chain was then applied, unchanged, beyond the surveyed sites to show what SDB can deliver for the waters of the United Arab Emirates as a whole: a coast-wide feasibility and water-clarity climatology from the Sentinel-2 archive of one year, a reconnaissance depth mosaic of the coast checked against and then calibrated on ICESat-2 cell by cell, depth maps of two areas that have never been surveyed, calibrated with ICESat-2 alone, and the year-to-year repeatability of the maps over five winters.

The specific objectives of the study were: (i) to establish, under a leakage-free validation, the accuracy and coverage of SDB in Gulf port waters for transfer and local-control use; (ii) to determine whether machine-learning and deep-learning models transfer better than the empirical transforms, and whether local control changes that ranking; (iii) to quantify the depth to which Sentinel-2 can read the seabed in these waters and how that depth and the error vary with season and turbidity; (iv) to measure the sensitivity of the accuracy to the number of images, the number of control soundings and the pixel size, including a VHR image; (v) to assess ICESat-2 ATL24 as an independent check and as a substitute for soundings; (vi) to provide operational guidance, including the IHO S-44 order that can be claimed, for the use of SDB by a port authority; and (vii) to establish, from the coast-wide climatology, the unsurveyed areas and the repeatability experiment, the outlook for an SDB service in UAE waters. The paper is organised as follows. Section~\ref{sec:data} describes the study area, the surveys, the imagery, the lidar and the tide data. Section~\ref{sec:methods} presents the processing chain, the models, the validation protocol and the experiments. Section~\ref{sec:results} reports the results, Section~\ref{sec:discussion} discusses their implications and limitations, and Section~\ref{sec:conclusions} concludes.

\section{Study area and data}
\label{sec:data}

\subsection{Study area and survey datasets}
\label{sec:surveys}

\begin{figure}[htbp]
\centering
\includegraphics[width=0.9\textwidth]{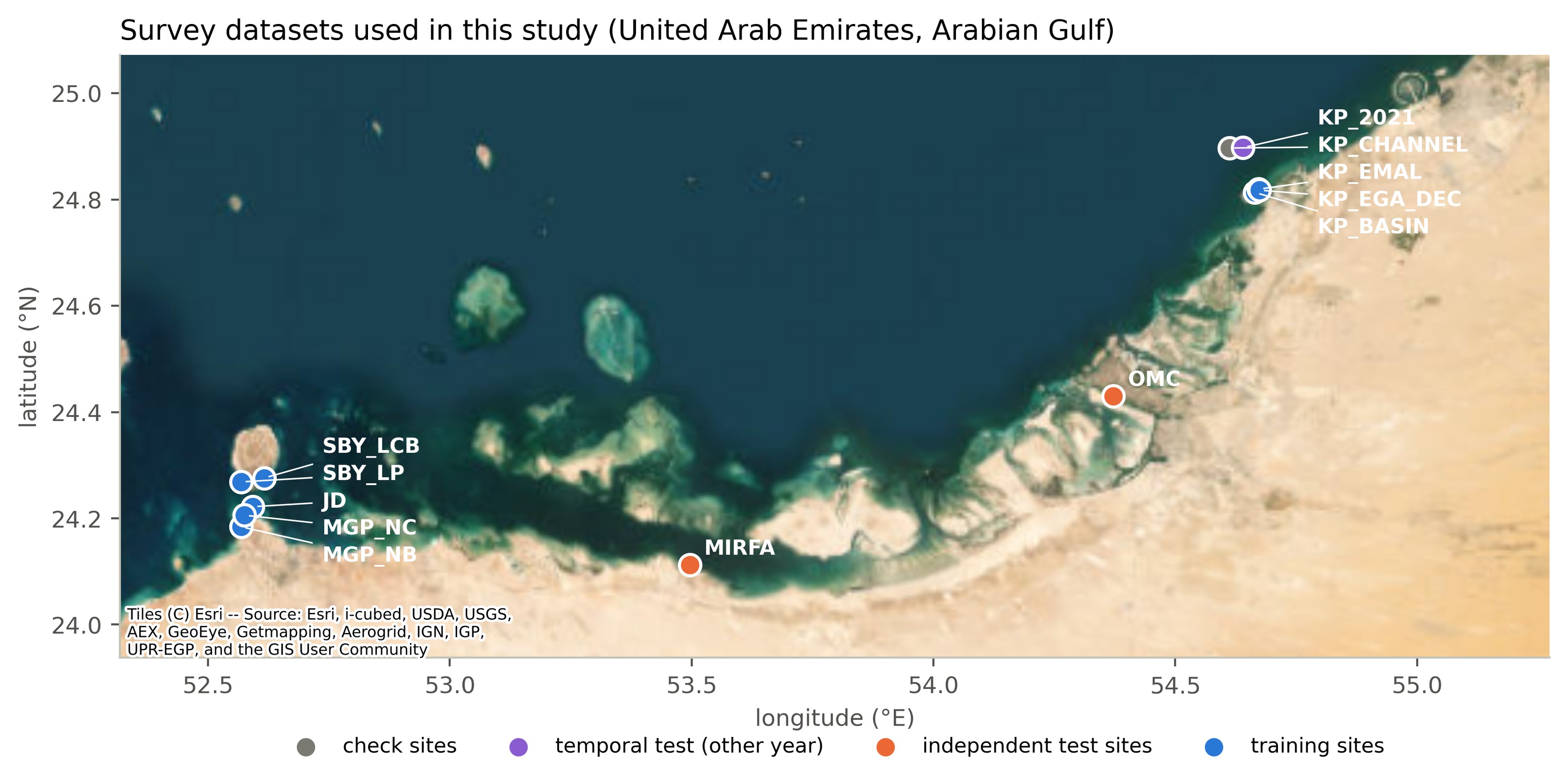}\\[2mm]
\includegraphics[width=\textwidth]{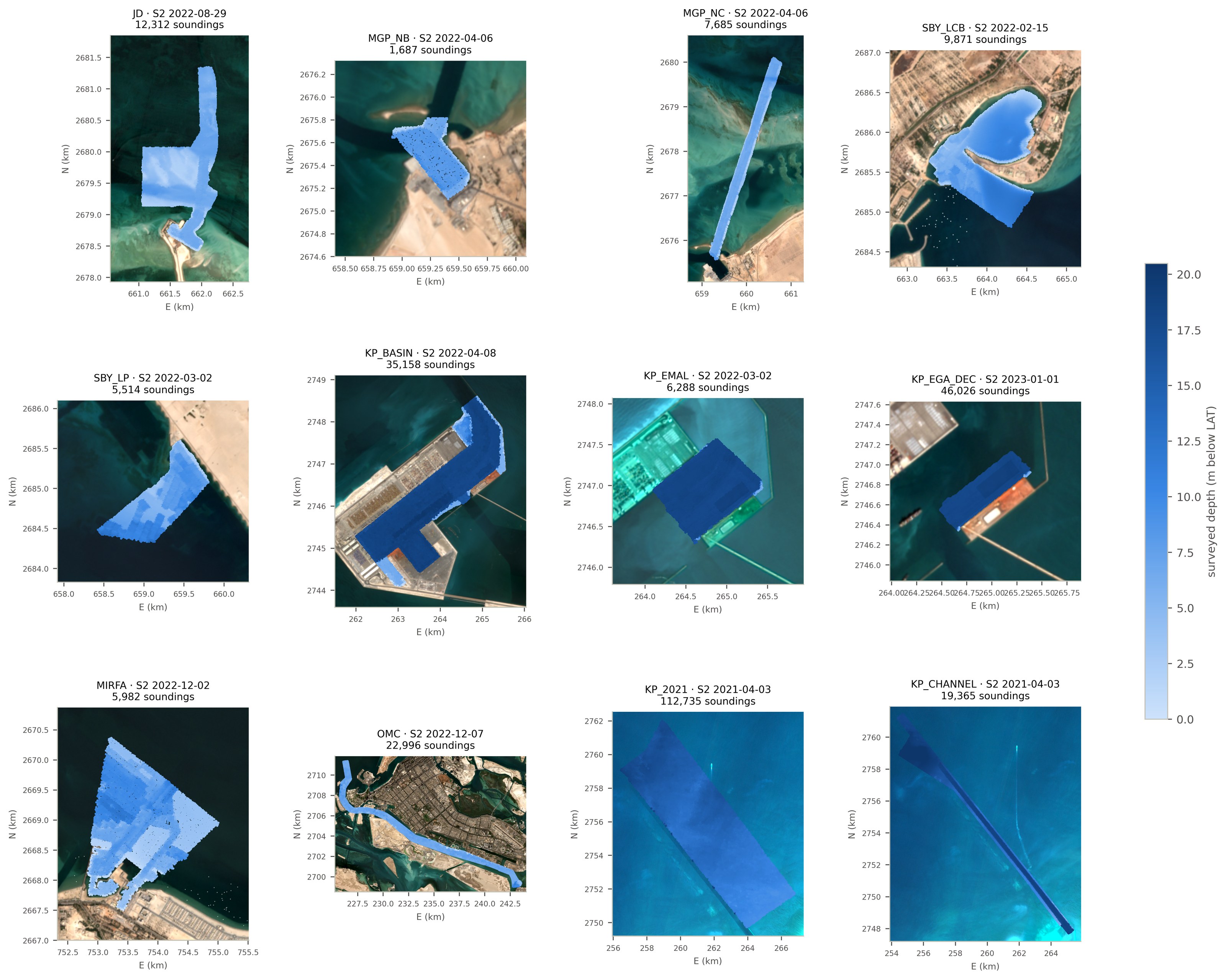}
\caption{(a) Location of the survey datasets along the Abu Dhabi coast, coloured by their role in the validation design (background: Esri World Imagery; sources: Esri, i-cubed, USDA, USGS, AEX, GeoEye, Getmapping, Aerogrid, IGN, IGP, UPR-EGP and the GIS User Community). (b) Each dataset on the Sentinel-2 true-colour image with the highest scene score after sun-glint removal, with the soundings (10\,m pixel medians) coloured by depth below LAT. Coordinates in the dataset's UTM zone (39N for Al Dhafra, 40N for Abu Dhabi). Contains modified Copernicus Sentinel data (2021--2023).}
\label{fig:studyarea}
\end{figure}

The study area extends along about 250\,km of the Abu Dhabi coast of the Arabian Gulf, from the Al Dhafra ports in the west (Jebel Dhanna, Mugharraq and Sir Bani Yas, UTM zone 39N) to Khalifa Port and the Musaffah channel in the east (UTM zone 40N) (Fig.~\ref{fig:studyarea}a). The coast is low-lying, fringed by sabkha and by extensive sand and carbonate flats shallower than 5\,m, and interrupted by dredged basins and approach channels whose depths reach 18--21\,m. Tides are mixed, predominantly semi-diurnal, with a spring range of about 2\,m at Khalifa Port and about 1.5\,m in the western embayments, so that the water column at the time of a satellite overpass may differ from the charted depth by more than 1\,m. The water is clear offshore in winter but is frequently turbid near the coast, in dredged basins and after shamal events, with suspended sediment and, in summer, phytoplankton reducing the visibility of the seabed.

Twelve MBES survey datasets from six port areas were made available to the Abu Dhabi Maritime Academy by the port operator (Table~\ref{tab:datasets}; Fig.~\ref{fig:studyarea}b). They comprise 285,619 soundings acquired between April 2021 and January 2023 and delivered as gridded ASCII files (10\,m or 20\,m cells, or a 2\,m grid in one case) referenced to WGS~84/UTM and to LAT. The survey standard stated in the survey reports was IHO Exclusive Order for the dredged basins and Order~1a elsewhere. The datasets cover natural shallow water (Mirfa, Sir Bani Yas logistics port and landing-craft basin, the Khalifa Port limits), dredged channels (Mugharraq north channel, Old Musaffah channel, Khalifa Port channel) and dredged basins (Jebel Dhanna, Mugharraq north basin, Khalifa Port basin and the EGA/EMAL berths). Depths range from $-0.26$\,m to 21.67\,m below LAT. The 2\,m EGA basin grid of December 2022 was thinned to the 10\,m analysis grid (46,026 soundings to 2,612 pixels), and the Khalifa Port channel file, which stores the shoalest sounding of each 20\,m cell rather than the mean, was used only as a check dataset.

Each dataset was assigned one role in the validation design, fixed before any model was fitted. Eight datasets (Jebel Dhanna, the two Mugharraq surveys, the two Sir Bani Yas surveys and the three Khalifa Port basin surveys of 2022) were used for training and cross-validation. Two datasets were held back as fully independent test sites that were never used for training, tuning, calibration of the uncertainty or choice of method: Mirfa, a natural shallow bay with sand flats, and the Old Musaffah channel (OMC), a dredged channel in turbid water. The 2021 survey of the Khalifa Port limits (KP\_2021), acquired a year before the Khalifa Port training surveys and over natural seabed outside the basins, served as a temporal test. The shoal-biased Khalifa Port channel grid served as a check only. In total 239,420 Sentinel-2 pixels of 10\,m contain at least one sounding, of which 35,762 lie at the training sites, within the depth limit derived in Section~\ref{sec:depthlimit}, and were seen by at least one usable image.

\begin{table}[htbp]
\centering
\caption{Multibeam survey datasets. Depths in m below LAT; \emph{pixels} = 10\,m Sentinel-2 pixels holding at least one sounding; \emph{deeper} = share of pixels deeper than the 11\,m depth limit of Section~\ref{sec:depthlimit}.}
\label{tab:datasets}
\small
\begin{adjustbox}{max width=\textwidth}
\begin{tabular}{@{}llllrrrrrr@{}}
\toprule
Dataset & Role & Survey dates & Seabed & Soundings & Pixels & $d_{\min}$ & $d_{\mathrm{med}}$ & $d_{\max}$ & Deeper (\%) \\
\midrule
JD (Jebel Dhanna basin and channel) & train & 30 Aug--1 Sep 2022 & dredged & 12,312 & 11,866 & 0.18 & 5.89 & 9.00 & 0 \\
MGP\_NB (Mugharraq north basin) & train & Apr--Aug 2022 & dredged & 1,687 & 1,622 & 1.27 & 8.18 & 10.55 & 0 \\
MGP\_NC (Mugharraq north channel) & train & Apr--Aug 2022 & dredged & 7,685 & 7,400 & 0.93 & 6.64 & 9.56 & 0 \\
SBY\_LCB (Sir Bani Yas landing-craft basin) & train & 16--27 Feb 2022 & natural & 9,871 & 9,441 & 0.82 & 8.62 & 13.58 & 18.9 \\
SBY\_LP (Sir Bani Yas logistics port) & train & 16--27 Feb 2022 & natural & 5,514 & 5,289 & 1.37 & 8.15 & 13.59 & 10.2 \\
KP\_BASIN (Khalifa Port basin) & train & 11--25 Apr 2022 & dredged & 35,158 & 34,040 & $-0.22$ & 18.52 & 20.98 & 88.1 \\
KP\_EMAL (Khalifa Port EGA/EMAL berths) & train & 1--2 Mar 2022 & dredged & 6,288 & 6,072 & 5.62 & 18.27 & 19.51 & 99.4 \\
KP\_EGA\_DEC (EGA basin, 2\,m grid) & train & 31 Dec 2022 & dredged & 46,026 & 2,612 & 3.56 & 18.19 & 19.72 & 99.4 \\
MIRFA (Mirfa bay) & independent test & 25 Oct--4 Nov 2022 & natural & 5,982 & 5,982 & $-0.26$ & 7.11 & 12.51 & 4.2 \\
OMC (Old Musaffah channel) & independent test & 7 Dec 2022--20 Jan 2023 & dredged & 22,996 & 22,996 & 0.02 & 7.04 & 11.53 & 0.3 \\
KP\_2021 (Khalifa Port limits) & temporal test & 1 Apr--26 May 2021 & natural & 112,735 & 112,735 & 8.51 & 13.74 & 19.58 & 84.0 \\
KP\_CHANNEL (Khalifa Port channel, shoal-biased) & check & Apr--May 2021 & dredged & 19,365 & 19,365 & 4.71 & 18.55 & 21.67 & 95.1 \\
\midrule
Total & & Apr 2021--Jan 2023 & & 285,619 & 239,420 & $-0.26$ & & 21.67 & \\
\bottomrule
\end{tabular}
\end{adjustbox}
\end{table}

\subsection{Sentinel-2 imagery}
\label{sec:s2}

Sentinel-2 Level-2A (bottom-of-atmosphere, Sen2Cor \cite{MainKnorn2017}) products were obtained from the Microsoft Planetary Computer STAC catalogue \cite{PlanetaryComputer}. Bands B01 (443\,nm, 60\,m), B02 (490\,nm), B03 (560\,nm), B04 (665\,nm), B05 (705\,nm, 20\,m), B08 (842\,nm), B11 (1610\,nm, 20\,m) and the scene classification layer were read as windows covering each survey polygon with a 500\,m buffer, plus a 8\,km context window at 60\,m used for the deep-water reference. All bands were resampled to the 10\,m grid of the site. For every dataset, up to 20 candidate images with less than 40\,\% tile cloud cover were examined within $\pm$365 days of the survey for natural seabeds and $\pm$60 days for dredged areas, where the seabed may change, and the six best images were kept according to a scene score that combines the clear-water fraction, the measured glint, a turbidity index, the time gap to the survey and the clear-water season (Section~\ref{sec:preproc}). In total 240 images were examined and 72 kept (Table~\ref{tab:images}; supplementary Fig.~\ref{fig:timeline}). The median time gap between the kept images and the survey was 0--20 days at eleven datasets and 53 days at Mirfa, whose autumn survey was paired with the clearer images of the following winter; the tide at overpass ranged from 0.12 to 2.24\,m above LAT. The seasonal experiment of Section~\ref{sec:seasons} examined a further 122 images and the outlook experiments of Section~\ref{sec:outlook} a further 900 or so (720 monthly images of the coastal cells and the candidates of the repeatability and unsurveyed-area experiments; Section~\ref{sec:outlookmethods}), so that about 1,300 Sentinel-2 images were examined in all.

\begin{table}[htbp]
\centering
\caption{Sentinel-2 L2A images kept for each dataset (six per dataset, from 20 examined): acquisition dates, median absolute time gap to the survey, median turbidity index (B05 reflectance above deep water, Section~\ref{sec:preproc}), number of images whose tide at overpass was taken from the survey gauge rather than the EOT20 model, and range of the tide above LAT at overpass.}
\label{tab:images}
\small
\begin{adjustbox}{max width=\textwidth}
\begin{tabular}{@{}lllrrcrr@{}}
\toprule
Dataset & First image & Last image & Gap (d) & Turbidity index & Gauge tides & $\eta_{\min}$ (m) & $\eta_{\max}$ (m) \\
\midrule
JD & 2022-08-29 & 2022-10-13 & 19.5 & 0.0105 & 1 & 0.65 & 1.50 \\
MGP\_NB & 2022-04-01 & 2022-06-20 & 0.0 & 0.0007 & 1 & 0.31 & 1.19 \\
MGP\_NC & 2022-04-01 & 2022-06-10 & 0.0 & 0.0064 & 1 & 0.30 & 1.19 \\
SBY\_LCB & 2022-01-11 & 2022-04-06 & 4.5 & 0.0002 & 1 & 0.12 & 1.04 \\
SBY\_LP & 2022-01-26 & 2022-04-06 & 8.5 & 0.0052 & 1 & 0.15 & 1.16 \\
KP\_BASIN & 2022-03-24 & 2022-05-13 & 8.0 & 0.0071 & 0 & 0.52 & 1.49 \\
KP\_EMAL & 2022-02-15 & 2022-03-14 & 8.5 & 0.0027 & 1 & 0.77 & 1.88 \\
KP\_EGA\_DEC & 2022-12-19 & 2023-01-16 & 8.5 & 0.0013 & 0 & 1.17 & 2.06 \\
MIRFA & 2022-12-02 & 2023-01-31 & 53.0 & 0.0017 & 0 & 1.34 & 1.93 \\
OMC & 2022-12-07 & 2023-01-21 & 0.0 & 0.0071 & 4 & 1.28 & 2.24 \\
KP\_2021 & 2021-03-27 & 2021-05-23 & 0.0 & 0.0006 & 5 & 0.34 & 1.84 \\
KP\_CHANNEL & 2021-04-03 & 2021-05-23 & 0.0 & 0.0007 & 6 & 0.34 & 1.69 \\
\bottomrule
\end{tabular}
\end{adjustbox}
\end{table}

\subsection{Very-high-resolution imagery}
\label{sec:vhrdata}

One Pl\'eiades-1A pan-sharpened multispectral image (0.5\,m, four bands, 8-bit display product) acquired over Khalifa Port on 3 April 2026 at 07:10 UTC was available. Because it is a display product rather than a radiometrically calibrated reflectance product, and because it was acquired 3.3--4.0 years after the Khalifa Port surveys, its use was restricted to a resolution comparison at the pixels of the KP\_BASIN and KP\_EGA\_DEC surveys that are shallower than the depth limit, and the results are interpreted with that caveat (Section~\ref{sec:vhr}). No VHR image was available over the clear, shallow western sites.

\subsection{ICESat-2 ATL24 lidar}
\label{sec:atl24data}

The ICESat-2 ATL24 product \cite{ATL24} classifies each photon of the ATLAS green laser as sea surface or seabed, corrects the seabed photons for refraction and provides a confidence and an uncertainty per photon. All 97 ATL24 granules (version 2, processed from ATL03 release 006) intersecting the twelve survey polygons between October 2018 and November 2024 were obtained from the NASA Earthdata archive, giving 170 site--pass combinations. The lidar tracks are thin (six beams of a few metres width, about 3.3\,km apart between beam pairs), so that each pass crosses a site along one or a few lines, and only 13 passes overlap a survey by three or more 10\,m pixels. The lidar was used for two purposes that were kept strictly separate by a fixed, name-based split of the passes: half of the passes (\emph{check}) served only as an independent test of the maps away from the soundings, and the other half (\emph{calibration}) were allowed to provide control points where no soundings exist, to join the control set of the local-control test (Section~\ref{sec:localcontrol}), to calibrate images in a lidar-only experiment and to calibrate the maps of two unsurveyed areas (Section~\ref{sec:outlookmethods}), for which the granules over those areas were obtained in the same way. For the coast-wide experiment a further 291 passes from October 2022 to December 2025 were obtained over the 49 coastal cells, within a download budget of 6\,GB and at most eight passes per cell.

\subsection{Tide model and gauges}
\label{sec:tidedata}

Water levels at the time of each overpass were predicted with the EOT20 global ocean tide model \cite{HartDavis2021} through the \texttt{eo-tides} library, which was also used to compute a model-consistent LAT at each site as the minimum of a 18.6-year prediction. The survey tide-gauge records that accompanied each survey (Khalifa Port, Jebel Dhanna/Mugharraq, Sir Bani Yas, Mirfa and Musaffah, all referenced to LAT) were used to check the model and to derive, for each gauge group, a constant datum offset between the model LAT and the gauge LAT (Section~\ref{sec:tides}). Where a gauge record covered an overpass, the gauge value was used directly (21 of the 72 images); otherwise the offset-corrected model was used.

\section{Methods}
\label{sec:methods}

\begin{figure}[htbp]
\centering
\includegraphics[width=\textwidth]{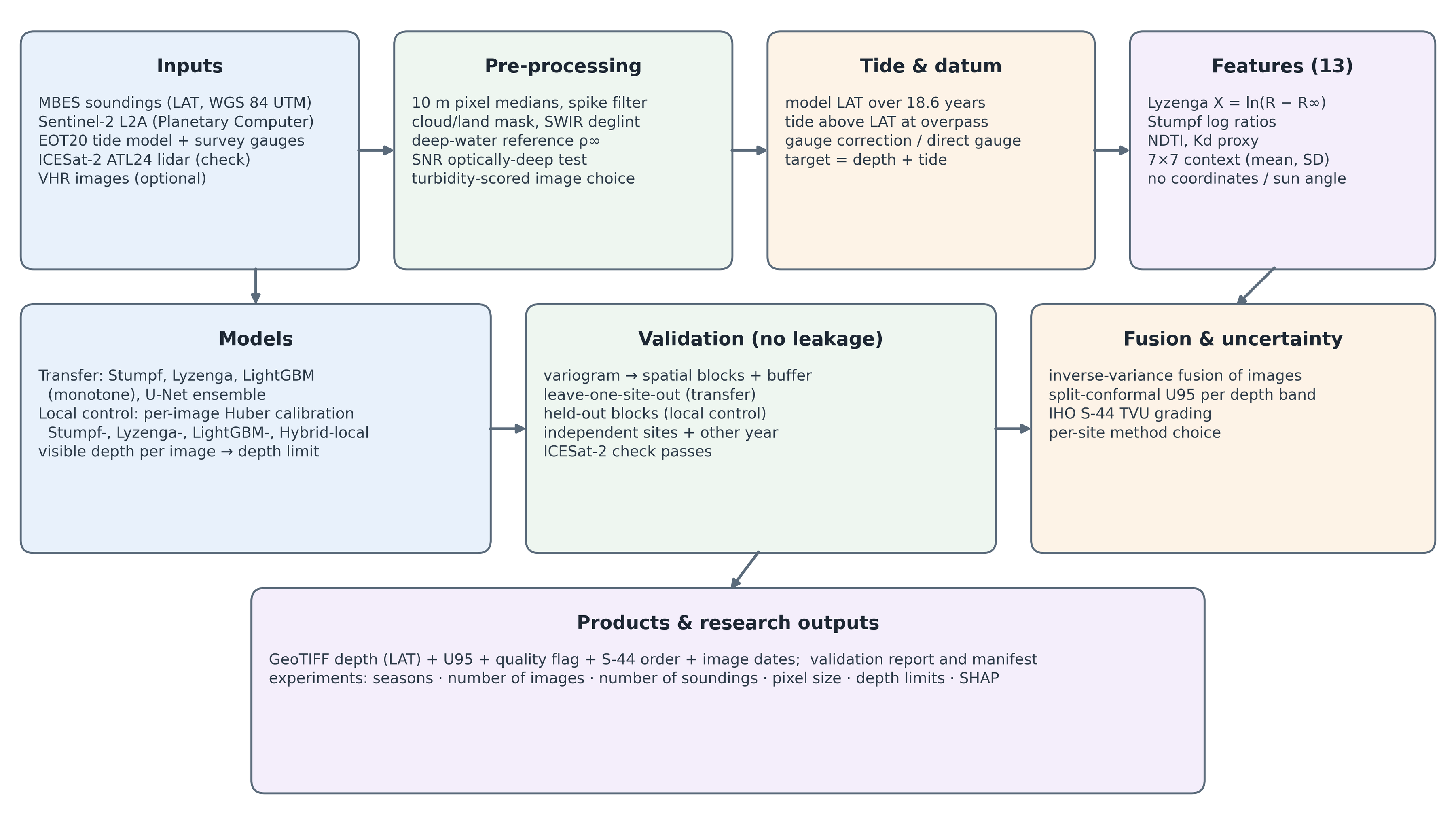}
\caption{Workflow of the study: inputs, pre-processing, tidal reduction to LAT, physics-based features, transfer and local-control models, leakage-free validation, fusion with conformal uncertainty and IHO S-44 grading, products and research experiments.}
\label{fig:workflow}
\end{figure}

Fig.~\ref{fig:workflow} summarises the processing chain. It was written as one documented Python notebook that runs on Google Colab from the raw soundings to the validation report. Every intermediate quantity (glint slope, deep-water reflectance, noise level, tide height, visible depth, scene score) is written to a run manifest, so that any published depth can be traced back to its inputs. The notation used below is listed in the Nomenclature.

\subsection{Pre-processing of the Sentinel-2 images}
\label{sec:preproc}

Each image was processed in six steps.

\begin{enumerate}[leftmargin=*,label=(\arabic*)]
\item Pixels classed as cloud shadow, medium or high cloud or cirrus in the scene classification layer were removed and the mask was grown by 50\,m, because cloud edges are under-detected.
\item Water was defined by a positive normalised difference water index together with a dark short-wave infrared band,
\begin{equation}
\mathrm{NDWI}=\frac{\rho_{B03}-\rho_{B08}}{\rho_{B03}+\rho_{B08}}>0 \quad\text{and}\quad \rho_{B11}<0.06,
\label{eq:ndwi}
\end{equation}
since land, quays, vessels and breakwaters are bright in the SWIR.
\item A deep-water set $\mathcal{D}$ was defined as the darkest 20\,\% of the water pixels in the green band within the 8\,km context window of the site.
\item Sun glint was removed with the SWIR method of Hedley et al.\ \cite{Hedley2005}, in the form reviewed by Kay et al.\ \cite{Kay2009}. Over $\mathcal{D}$ each visible band $i$ was regressed on B11,
\begin{equation}
\rho_i = a_i + \beta_i\,\rho_{B11}, \qquad i\in\{B01,\dots,B04\},
\label{eq:glintfit}
\end{equation}
and where the regression explained at least 30\,\% of the variance the glint term was subtracted from every water pixel,
\begin{equation}
\rho_i' = \rho_i - \beta_i\left(\rho_{B11}-\rho_{B11,\min}\right).
\label{eq:deglint}
\end{equation}
Otherwise no measurable glint was present and the band was left unchanged. The primes are dropped below; all later quantities use the deglinted reflectances.
\item The deep-water reflectance $\rho_{\infty,i}$ of each band, which combines the residual atmospheric path reflectance after Sen2Cor and the reflectance of the water column itself, was taken as the darkest 2\,\% of the deglinted water pixels. It is the reference of the Lyzenga transform and of the test that follows.
\item An \emph{optically deep} test was applied per pixel. The seabed signal in the green band was measured against the sensor noise,
\begin{equation}
\mathrm{SNR}=\frac{\overline{\rho}_{B03}^{\,3\times3}-\rho_{\infty,B03}}{\sigma_n}, \qquad
\sigma_n=\max\!\left(1.4826\,\operatorname{med}\left|\Delta\rho_{B03}\right|,\;0.0005\right),
\label{eq:snr}
\end{equation}
where $\overline{\rho}_{B03}^{\,3\times3}$ is the $3\times3$ mean and $\Delta\rho_{B03}$ the pixel-to-pixel difference of the site's water pixels, a high-pass estimate in which depth gradients do not count as noise. Pixels with $\mathrm{SNR}<3$ were flagged optically deep and given no depth.
\end{enumerate}

A normalised difference turbidity index,
\begin{equation}
\mathrm{NDTI}=\frac{\rho_{B04}-\rho_{B03}}{\rho_{B04}+\rho_{B03}},
\label{eq:ndti}
\end{equation}
flagged pixels above 0.05 as turbid; turbid pixels were excluded from training.

The candidate images of each dataset were ranked by a scene score, the product of five terms,
\begin{equation}
S = \underbrace{\left(f_{\mathrm{clear}}-f_{\mathrm{turb}}\right)e^{-g/0.01}}_{\text{clear water}}
\;\underbrace{e^{-(T-T_{\mathrm{ref}})/0.01}}_{\text{clarity}}
\;\underbrace{\max\!\left(0.3,\,r_s\right)}_{\text{depth signal}}
\;\underbrace{e^{-\Delta t/\tau}}_{\text{date}}
\;\underbrace{s_{\mathrm{season}}}_{\text{season}},
\label{eq:score}
\end{equation}
where $f_{\mathrm{clear}}$ and $f_{\mathrm{turb}}$ are the clear-water and turbid fractions of the sounding pixels and $g$ the mean glint; $T$ is the turbidity index of the image, defined as the median red-edge (B05) reflectance of the site's water above its deep-water reference (suspended sediment raises B05, whereas the seabed hardly does) and $T_{\mathrm{ref}}$ the 25th percentile of $T$ over the site's candidates; $r_s$ is the Spearman correlation between the blue/green ratio and the soundings, used at the training sites only and set to 1 elsewhere; $\Delta t$ is the time between image and survey, with $\tau=20$ days for dredged and 240 days for natural seabeds; and $s_{\mathrm{season}}=1$ from December to May and 0.5 otherwise. The season term was changed during the study. The first version of the chain preferred April to September, following the Secchi-depth climatology of the region, but a seasonal analysis made with that version showed the turbidity index of the images to be several times higher from July to September than in winter. The present version therefore prefers December to May and searches a full year around a natural-seabed survey so that a clear winter is always available; the seasonal experiment of Section~\ref{sec:seasons}, repeated with this version, confirms the choice. The six images with the highest score were kept.

\subsection{Tidal reduction and vertical datum}
\label{sec:tides}

The soundings are referenced to LAT, whereas an image records the water column at the time of the overpass. The tide height above LAT at each overpass, $\eta(t)$, was predicted, the models were trained on the water column rather than on the charted depth, and every satellite depth was reduced to LAT after prediction:
\begin{equation}
h = d + \eta(t), \qquad \hat d = \hat h - \eta(t).
\label{eq:tide}
\end{equation}
The model LAT was computed per site as the minimum of an 18.6-year EOT20 prediction, and the EOT20 series was compared with the survey tide gauges over the full length of each gauge record (supplementary Fig.~\ref{fig:tidecheck}). Phase and amplitude agreed closely, but a constant offset remained between the model LAT and the gauge LAT. It was estimated from the water-level records alone, as the median of the difference between gauge and model: $0.00$\,m at Khalifa Port, $+0.15$\,m at Jebel Dhanna/Mugharraq, $-0.39$\,m at Sir Bani Yas, $-0.43$\,m at Mirfa and $-0.28$\,m at Musaffah. These five offsets are the only site-specific quantities in the chain, and they come from water levels, not from the soundings. Where a gauge record covered an overpass, the gauge value was used instead of the model.

\subsection{Features}
\label{sec:features}

Thirteen features per pixel were computed from the corrected reflectances (Table~\ref{tab:features}). They fall into four groups with a physical reading. The first is the Lyzenga transform \cite{Lyzenga1978,Lyzenga1985}, which follows from the single-scattering model of the water-leaving reflectance over a bottom of albedo $A_i$,
\begin{equation}
\rho_i = \rho_{\infty,i} + \left(A_i-\rho_{\infty,i}\right)e^{-2K_{d,i}h},
\qquad
X_i=\ln\!\left(\rho_i-\rho_{\infty,i}\right)=\ln\!\left(A_i-\rho_{\infty,i}\right)-2K_{d,i}\,h,
\label{eq:lyzenga}
\end{equation}
so that $X_i$ is linear in the water column for a uniform bottom, with a slope set by the diffuse attenuation coefficient $K_{d,i}$. The transforms of bands B01--B04 were used. The second group is the Stumpf log-ratio \cite{Stumpf2003},
\begin{equation}
p_{ij}=\frac{\ln(n\rho_i)}{\ln(n\rho_j)}, \qquad n=1000,
\label{eq:stumpf}
\end{equation}
for the pairs B02/B03, B02/B04 and B03/B04. Because the band with the stronger attenuation falls faster with depth, the ratio grows with depth, and because a change of bottom albedo moves both logarithms by nearly the same amount, it largely cancels. The third group describes the water column: the turbidity index NDTI of Eq.~\eqref{eq:ndti} and the ratio $\rho_{B02}/\rho_{B03}$ as a proxy for the attenuation. The fourth group is spatial context: the mean of $p_{23}$ over $7\times7$ pixels (70\,m), its standard deviation over $3\times3$ and $7\times7$ pixels, and the $7\times7$ mean of $X_{B03}$, which steady the estimate on slopes and suppress pixel noise.

Three kinds of information were excluded on purpose. Coordinates and site identifiers would let a model memorise where the deep basins are. The sun angle and the absolute reflectances are nearly constant within an image and act as a fingerprint of the image and site; in a preliminary run they received 35\,\% of the model's attention and caused biases of 6--9\,m at unseen sites. The signal-to-noise ratio was used only for the optically deep flag, because its scale depends on how much deep water surrounds a site. Every retained feature is a difference, a ratio or a neighbourhood statistic, so that quantities constant within an image cancel.

\begin{table}[htbp]
\centering
\caption{Features used by the LightGBM and U-Net models (all 13) and by the empirical transforms (subsets). The monotonic constraints were imposed on the LightGBM model.}
\label{tab:features}
\small
\begin{adjustbox}{max width=\textwidth}
\begin{tabular}{@{}llll@{}}
\toprule
Group & Feature(s) & Definition & Monotonic constraint \\
\midrule
Lyzenga transforms & $X_{B01}, X_{B02}, X_{B03}, X_{B04}$ & Eq.~\eqref{eq:lyzenga} & depth decreases with $X_{B02}$, $X_{B03}$ \\
Stumpf log-ratios & $p_{23}, p_{24}, p_{34}$ & Eq.~\eqref{eq:stumpf} & depth increases with $p_{23}$ \\
Water column & NDTI, $K_d$ proxy & Eq.~\eqref{eq:ndti}, $\rho_{B02}/\rho_{B03}$ & none \\
Spatial context & $m_7(p_{23}), s_3(p_{23}), s_7(p_{23}), m_7(X_{B03})$ & mean/SD over $3\times3$ and $7\times7$ pixels & none \\
\bottomrule
\end{tabular}
\end{adjustbox}
\end{table}

\subsection{Models}
\label{sec:models}

Four model families were compared on identical training rows: pixel $\times$ image samples at the training sites, in clear and non-turbid water, with a positive water column and a surveyed depth within the visible depth of the image (Section~\ref{sec:depthlimit}).

\emph{Stumpf ratio.} The water column was regressed linearly on $p_{23}$ by weighted least squares,
\begin{equation}
\hat h = m_0 + m_1\,p_{23},
\label{eq:stumpfmodel}
\end{equation}
the classical two-parameter ratio transform \cite{Stumpf2003}.

\emph{Lyzenga transform.} The water column was regressed on the four standardised transforms,
\begin{equation}
\hat h = c_0 + \sum_{i=1}^{4} c_i\,\tilde X_i, \qquad \tilde X_i=\frac{X_i-\mu_i}{s_i},
\label{eq:lyzengamodel}
\end{equation}
with the Huber loss
\begin{equation}
L_\epsilon(e)=\begin{cases} \tfrac12 e^2, & |e|\le\epsilon \\ \epsilon\left(|e|-\tfrac12\epsilon\right), & |e|>\epsilon\end{cases}
\label{eq:huber}
\end{equation}
and $\epsilon=1.5$\,m, which limits the influence of outliers \cite{Lyzenga2006}.

\emph{LightGBM.} A gradient-boosted decision-tree model \cite{Ke2017} was trained on the 13 features. The prediction is an additive expansion of $M$ regression trees $f_m$ fitted in sequence, each to the negative gradient of the loss of the previous stage,
\begin{equation}
\hat h = F_M(\mathbf{x}) = F_0 + \nu\sum_{m=1}^{M} f_m(\mathbf{x}),
\label{eq:gbdt}
\end{equation}
with a learning rate $\nu=0.05$. Monotonic constraints encoded the attenuation physics of Eqs.~\eqref{eq:lyzenga} and \eqref{eq:stumpf}:
\begin{equation}
\frac{\partial F_M}{\partial p_{23}}\ge 0, \qquad
\frac{\partial F_M}{\partial X_{B02}}\le 0, \qquad
\frac{\partial F_M}{\partial X_{B03}}\le 0,
\label{eq:monotone}
\end{equation}
so that the predicted depth cannot fall when the ratio rises, nor rise when the blue or green transform rises. Three settings (31, 63 and 127 leaves, with minimum child samples of 100, 50 and 30 and L2 regularisation of 1, 2 and 5) were compared by the spatial block cross-validation of Section~\ref{sec:validation}. Within each fold 15\,\% of the training blocks were withheld for early stopping, so that neither the test fold nor the test sites influenced the number of trees. The selected setting (127 leaves, minimum child samples of 30, L2 regularisation of 5, $M=2{,}553$ trees, feature and row subsampling of 0.8) was refitted on all training rows. Boosted trees were preferred to a deep network as the main model because, with about $10^5$ sounding pixels, they train in minutes on a CPU, accept monotonic constraints and can be interpreted with Shapley values.

\emph{U-Net ensemble.} As a deep-learning challenger, eight U-Nets \cite{Ronneberger2015} were trained on $64\times64$-pixel patches of the 13-feature image. The loss was the Huber loss of Eq.~\eqref{eq:huber} averaged over the pixels that hold a sounding,
\begin{equation}
\mathcal{L}=\frac{\sum_{k} \omega_k\, L_\epsilon\!\left(h_k-\hat h_k\right)}{\sum_k \omega_k}, \qquad \omega_k=\begin{cases}1, & \text{pixel $k$ holds a valid sounding}\\ 0, & \text{otherwise,}\end{cases}
\label{eq:unetloss}
\end{equation}
so that unsounded pixels contribute nothing to the gradient. For each of the four regions (Jebel Dhanna, Mugharraq, Sir Bani Yas, Khalifa Port) two networks with different random seeds were trained on the other three regions, with one training site of those regions withheld for early stopping. Training sites were predicted with the two networks that had never seen their region (a leave-region-out test, stricter than leave-one-site-out), and the test, temporal and check sites with the mean of all eight networks.

\subsection{Local control}
\label{sec:localcontrol}

In hydrographic practice an SDB image is calibrated with a small number of control depths within the image, such as old survey lines, a single-beam run or a partial recent survey. This \emph{local-control} mode was implemented for every image and site. The control soundings seen in the image (clear, non-turbid water) were used first to find the visible depth of the image (Section~\ref{sec:depthlimit}) and then to fit, by Huber regression on the controls shallower than that depth, one of five relations:
\begin{align}
\text{Stumpf-local:}\quad & \hat h = a + b\,p_{23}, \label{eq:stumpflocal}\\
\text{Lyzenga-local:}\quad & \hat h = a + \textstyle\sum_{i=1}^{4} b_i\,X_i, \label{eq:lyzengalocal}\\
\text{LightGBM-local:}\quad & \hat h = a + b\,\hat h_{\mathrm{GBDT}}, \label{eq:gbdtlocal}\\
\text{UNet-local:}\quad & \hat h = a + b\,\hat h_{\mathrm{UNet}}, \label{eq:unetlocal}\\
\text{Hybrid-local:}\quad & \hat h = a + b\,\hat h_{\mathrm{GBDT}} + c\,p_{23} + e\max\!\left(0,\,\kappa-p_{23}\right). \label{eq:hybrid}
\end{align}
The first is the classical per-image ratio method. The linear re-scaling of the transfer models in Eqs.~\eqref{eq:gbdtlocal} and \eqref{eq:unetlocal} keeps what the model has learned about bottom types and turbidity while the controls remove the site bias. The hybrid of Eq.~\eqref{eq:hybrid} was introduced because a tree model cannot predict shallower than the shallowest water in its training data (about 2\,m here) and therefore maps every flat shallower than 2\,m at 2--2.5\,m, whereas the ratio term keeps falling as the water shallows, so the hybrid can follow shallow controls. Its last term is a shallow-water hinge: $\kappa$ is the median ratio of the controls within $\pm1$\,m of 2.5\,m, the term is active only where the ratio lies below $\kappa$, i.e.\ in water shallower than about 2.5\,m, and it lets the fit bend there, where a straight line in $p_{23}$ is known to over-estimate the depth of the flats. The hybrid was fitted with depth-balanced weights,
\begin{equation}
\omega_k=\min\!\left(5,\;\frac{\bar n}{n_{b(k)}}\right),
\label{eq:balance}
\end{equation}
where $n_{b(k)}$ is the number of controls in the 1\,m band of control $k$ and $\bar n$ their mean, so that each band counts about equally. The hinge was fitted only when at least 15 controls lay on each side of 2.5\,m and was kept only if its coefficient had the physical sign (not deeper in shallower water); if the ratio coefficient came out with the wrong sign, the ratio and hinge terms were dropped and the hybrid reduced to LightGBM-local.

Where verified ICESat-2 calibration-pass pixels existed at a site (Section~\ref{sec:atl24methods}), they were added to the control set of every calibration as independent control depths: 1,162 lidar pixels, shallower than 8\,m and away from the soundings, at six datasets (405 at Jebel Dhanna, 611 at the Mugharraq channel, 18 and 36 at the two Sir Bani Yas sites, 15 at the EGA/EMAL berths and 77 at Mirfa). They are measurements independent of the surveys, so they do not leak the test, and they sample the shallow margins that the vessel surveys do not. An image was rejected if its local fit explained less than 30\,\% of the variance or had the wrong sign, because the seabed was then not readable in it, and at most 5,000 control pixels (a random subset) were used per image. For the training sites, LightGBM-local and UNet-local start from the model that had never seen the site, so the local test is leakage-free for every site.

\subsection{Visible depth and depth limit}
\label{sec:depthlimit}

The depth to which the colour of the water still follows the seabed, called here the \emph{visible depth} of an image, was found from the controls. The controls were grouped into 1\,m depth bins $b$ and the median ratio $\tilde p_b$ of each bin computed with its standard error $\mathrm{se}_b$. Starting from the shallowest bin, the water was considered visible in bin $b$ as long as
\begin{equation}
\tilde p_b - \tilde p_{b-3} > 2\sqrt{\mathrm{se}_b^2+\mathrm{se}_{b-3}^2},
\label{eq:visible}
\end{equation}
i.e.\ as long as the ratio in a bin still exceeded, by more than two standard errors, that of the bin 3\,m shallower; the visible depth was the deepest bin that passed, two failing bins in a row ending the walk. A channel of nearly uniform depth, or turbid water whose colour does not follow depth, yields no visible depth, and such images were excluded from training and from local calibration. With too few controls for the test, the deepest control was used provided that colour and depth were clearly correlated (Spearman $r_s\geq0.3$). The depth limit of an image in local-control mode was the smallest of its visible depth, a ceiling of 12\,m and the deepest control plus 1\,m. For transfer mode a single data-driven limit was set to the 95th percentile of the training depths retained within their images' visible depths, rounded up to 0.5\,m, which gave $d_{\max}=11$\,m. Accuracy is reported for true depths up to this limit, and transfer predictions deeper than it are flagged rather than published. The 2022 EGA/EMAL berth survey (KP\_EMAL) had no image with a visible depth and therefore contributed no training rows.

A complementary \emph{seabed not readable} test used the controls beyond the visible depth (or all controls where the image had none) to characterise the $7\times7$ blue/green ratio of water whose colour no longer follows depth. A pixel whose ratio exceeded the median of those controls minus one robust standard deviation voted \emph{not readable}, and pixels for which most images voted so were flagged \emph{beyond visible depth} and given no depth. This test is what stops the chain from publishing depths of about 10\,m for the 18--21\,m turbid basins of Khalifa Port, which pass the signal-to-noise test of Eq.~\eqref{eq:snr} because turbid water is brighter than the offshore water used as $\rho_\infty$.

\subsection{Validation protocol}
\label{sec:validation}

The protocol was designed so that no test pixel is ever predicted by a fit that used it or its neighbours. First, a quick Stumpf fit was made on the training sites and the empirical semivariogram of its residuals $\varepsilon$ was computed,
\begin{equation}
\gamma(\ell)=\frac{1}{2N(\ell)}\sum_{(k,l):\,|\mathbf{x}_k-\mathbf{x}_l|\approx\ell}\left(\varepsilon_k-\varepsilon_l\right)^2;
\label{eq:variogram}
\end{equation}
the lag at which $\gamma(\ell)$ reached 95\,\% of its sill, the correlation range, was 450\,m (supplementary Fig.~\ref{fig:variogram}). Each site was cut into square blocks of 450\,m, and whole blocks, in all images, were assigned to five folds, with training pixels closer than half a block (225\,m) to any test pixel also dropped. This \emph{spatial block cross-validation} \cite{Roberts2017} was used for model tuning and gives the accuracy inside a surveyed area. Second, the \emph{leave-one-site-out} (LOSO) test retrained each transfer method on all training sites but one and predicted the omitted site, which gives the accuracy at a new site. Third, the \emph{local-control} test calibrated each image on one of the five block groups of a site (about 20\,\% of the site) and predicted the pixels of the other groups beyond the 225\,m buffer, rotating until every group had been predicted once. Fourth, the two independent sites (Mirfa and OMC) were predicted once, at the end, by the final models and by local calibration, and the temporal site (KP\_2021) and the check site (KP\_CHANNEL) likewise. Finally, to measure the leakage that a conventional design produces, the LightGBM model was also cross-validated with a random 80/20 pixel split on the same rows, with the same settings and no other change.

With $e_k=\hat d_k-d_k$ the error of pixel $k$ (positive when the satellite depth is too deep) over $N$ test pixels, accuracy was reported as
\begin{equation}
\mathrm{RMSE}=\sqrt{\tfrac{1}{N}\textstyle\sum_k e_k^2},\quad
\mathrm{MAE}=\tfrac{1}{N}\textstyle\sum_k |e_k|,\quad
\mathrm{bias}=\tfrac{1}{N}\textstyle\sum_k e_k,\quad
R^2=1-\frac{\sum_k e_k^2}{\sum_k (d_k-\bar d)^2},
\label{eq:metrics}
\end{equation}
together with the normalised median absolute deviation, $\mathrm{NMAD}=1.4826\,\operatorname{med}\left|e_k-\operatorname{med}(e)\right|$, and the 95th percentile of $|e_k|$, $P_{95}$. All were computed overall and by surveyed-depth band (0--2, 2--5, 5--10 and 10--15\,m), with a depth-balanced RMSE in which each band counts equally. Coverage was reported as the share of in-range sounding pixels that received a depth, and the deep-flagged share as the proportion of pixels deeper than the limit that were correctly given no depth. Each pixel error was compared with the IHO S-44 total vertical uncertainty \cite{IHO2020},
\begin{equation}
\mathrm{TVU}(d)=\sqrt{a^2+(b\,d)^2},
\label{eq:tvu}
\end{equation}
with $(a,b)=(0.15,0.0075)$ for the Exclusive Order, $(0.25,0.0075)$ for Special Order, $(0.50,0.013)$ for Order~1a/1b and $(1.00,0.023)$ for Order~2, and the share of pixels with $|e_k|\le\mathrm{TVU}(d_k)$ was reported. An Order was considered met when at least 95\,\% of the pixels were within its TVU, and, per 1\,m depth bin, when $P_{95}$ did not exceed the TVU.

\subsection{Multi-image fusion, uncertainty and S-44 grading}
\label{sec:fusion}

The depths from the $n$ images of a site were fused per pixel by a weighted mean,
\begin{equation}
\hat d=\frac{\sum_{s=1}^{n} w_s\,\hat d_s}{\sum_{s=1}^{n} w_s}, \qquad
w_s=\frac{S_s}{q_s^2+4r_s^2},
\label{eq:fusion}
\end{equation}
where $S_s$ is the scene score of image $s$ from Eq.~\eqref{eq:score}, $q_s$ the typical 95\,\% error of a single-image prediction of that method in that depth band (taken from the LOSO test for the transfer methods and from the held-out local-control test for the local methods) and $r_s$ the residual RMSE of the local calibration of image $s$ ($r_s=0$ for the transfer methods, whose weight reduces to $S_s/q_s^2$). The second term in the denominator makes an image whose calibration left a large residual count less. It was added after a preliminary run had shown the fused local-control error rising beyond three images because poorly calibrated images were weighted like good ones. The weighted standard deviation of the $\hat d_s$ about $\hat d$,
\begin{equation}
\sigma_b=\sqrt{\frac{\sum_s w_s\left(\hat d_s-\hat d\right)^2}{\sum_s w_s}},
\label{eq:sigmab}
\end{equation}
measures the between-image disagreement. The fused depth was then passed through a $3\times3$ median filter where at least five of the nine pixels hold a depth, which removes the pixel-to-pixel noise that the resolution experiment of Section~\ref{sec:resolution} shows to be noise rather than seabed, without blurring the edges of channels and basins. A pixel also received the weighted mean of its images' depth limits and the not-readable votes, and no depth was published beyond the limit. One method was selected per mode: the transfer method with the lowest fused LOSO error, and the local method with the lowest depth-balanced held-out error, Hybrid-local being preferred if within 5\,\% of the best because it alone maps water shallower than 2\,m. Methods covering less than 80\,\% of what the best method covered were not eligible, and flagged pixels were counted at their limit so that a method cannot win by refusing difficult pixels. The U-Net was reported but not eligible as a product. For the maps, the method was additionally chosen per site from the site's own held-out test among the local methods and all transfer methods, and no map was published where no method beat a constant depth ($R^2\leq0$ on the held-out test).

A 95\,\% uncertainty per pixel, U95, was obtained by Mondrian split-conformal prediction \cite{Vovk2005,Angelopoulos2023}. On the fused training-site pixels, each predicted without its own soundings, the normalised nonconformity score
\begin{equation}
\zeta_k=\frac{\left|d_k-\hat d_k\right|}{\sigma_{b,k}+\gamma_0}
\label{eq:score_conf}
\end{equation}
was computed, where $\gamma_0$ is a floor equal to the typical between-image spread of the method. Its 95\,\% quantile $q_{95}$ was taken within each predicted-depth band, and the interval of a new pixel is
\begin{equation}
\mathrm{U95}=q_{95}(\text{band})\times\left(\sigma_b+\gamma_0\right),
\label{eq:u95}
\end{equation}
so that it is wider where the images disagree and in depth bands where the method is less reliable. Its calibration was checked at several nominal levels on the training sites and on the independent sites, which were never used in its construction. Because a pooled quantile cannot anticipate a site whose errors exceed those of the calibration population, a second, \emph{site-conformal} quantile was computed for use at new areas: the quantile was taken per training site $s$ and per depth band (where at least 50 pixels were available) and the largest value over the sites was kept,
\begin{equation}
q_{95}^{\mathrm{new}}(\text{band})=\max_{s}\,q_{95}^{(s)}(\text{band}),
\label{eq:u95new}
\end{equation}
so that the interval is as wide as the worst training site requires. The pooled interval (U95) is used for the maps of the training sites and the site-conformal interval (U95$_{\mathrm{new}}$) for the independent sites, for areas without soundings and for the outlook products; both are reported. U95 was also compared with the S-44 TVU of Eq.~\eqref{eq:tvu} to produce an estimated-Order map, and the products were written as GeoTIFF rasters of depth, U95, quality flag and estimated Order, each carrying the acquisition dates of the images used.

\subsection{ICESat-2 ATL24 processing}
\label{sec:atl24methods}

Seabed photons (class 40) with a confidence of at least 0.6, without low-confidence or sensor-depth-exceeded flags and with a depth between 0.2 and 20\,m below the modelled sea surface were kept, and their median per pass, beam and 10\,m pixel was taken where at least three photons fell in the pixel. Each pass was reduced to LAT with the EOT20 tide (or the gauge) at the pass time, as in Eq.~\eqref{eq:tide}. Every pass that overlapped a survey by at least three pixels was verified against it, and a pass whose median difference exceeded 1\,m was rejected; a site with any rejected pass contributed no lidar controls. Three passes were rejected, all over the deep (12--21\,m) waters of Khalifa Port, where the lidar returned false seabed photons at 1--6\,m (median differences of $-12$ to $-19$\,m). Because every verified pass read deeper than the survey, a robust straight line
\begin{equation}
d_{\mathrm{lidar}}-d_{\mathrm{survey}}=a_L+b_L\,d_{\mathrm{survey}}
\qquad\Longrightarrow\qquad
d_{\mathrm{lidar}}^{\mathrm{corr}}=\frac{d_{\mathrm{lidar}}-a_L}{1+b_L}
\label{eq:lidaroffset}
\end{equation}
was fitted by trimmed least squares to the verified overlap pixels of the non-test sites and removed from every lidar depth; on this run the fit gave $a_L=+0.28$\,m and $b_L=+3.5\,\%$ from 91 pixels and seven passes. The corrected lidar was then compared with the surveys (which validates the lidar and the tide reduction), the maps were compared with the lidar of the check passes away from the soundings (which tests the maps independently), and, in a separate experiment, the images were calibrated with the lidar of the calibration passes alone and scored on survey pixels more than 225\,m from any lidar pixel. Lidar control candidates were limited to 0--8\,m depth, to trusted sites and to pixels without soundings, and each candidate was further checked against an independent first map before use.

\subsection{Research experiments}
\label{sec:experiments}

Six experiments were run with the same models and the same leakage-free tests. (1)~\emph{Seasons}: for seven datasets with a natural or stable seabed (JD, MGP\_NC, SBY\_LCB, SBY\_LP, MIRFA, OMC and KP\_2021), the four seasons (December--February, March--May, June--August, September--November) within the twelve months centred on the survey were each represented by the three best of up to five candidate images, chosen by cloud and glint only (turbidity was not used in the choice), and the turbidity index, visible depth, colour--depth correlation, transfer error (Stumpf and LightGBM) and local-control error and coverage (Stumpf-local and Hybrid-local) were computed per site and season. (2)~\emph{Number of fused images}: random subsets of one to six images were fused (eight repeats) and scored with the held-out predictions. (3)~\emph{Number of control soundings}: the local-control test was repeated with 30, 60, 125, 250, 500, 1,000 and 2,000 random control pixels per image (two seeds) and with all controls of the calibration group. (4)~\emph{Pixel size}: the Sentinel-2 features were aggregated to 20, 30 and 60\,m and the local-control test repeated; the Pl\'eiades image was tested at 2, 5 and 10\,m. (5)~\emph{Error against depth}: the error was binned by 1\,m of true depth and $P_{95}$ compared with the TVU of each Order, and the visible depth of each image was related to its turbidity index. (6)~\emph{Feature contributions}: Shapley additive explanations (SHAP) \cite{Lundberg2017} were computed from the native contributions of the LightGBM model on a random sample of training rows; for an additive tree model the prediction decomposes exactly as
\begin{equation}
\hat h(\mathbf{x})=\phi_0+\sum_{j=1}^{13}\phi_j(\mathbf{x}),
\label{eq:shap}
\end{equation}
where $\phi_0$ is the mean prediction and $\phi_j$ the contribution of feature $j$, and the mean of $|\phi_j|$ over the sample was reported as the importance of feature $j$.

\subsection{Outlook experiments: the UAE coast beyond the surveyed sites}
\label{sec:outlookmethods}

Three further experiments took the chain, unchanged, beyond the surveyed sites, to describe what SDB can deliver for the waters of the United Arab Emirates as a whole.

(1)~\emph{Coast-wide feasibility.} The coast from 51.55\degree E to 56.40\degree E and from 23.95\degree N to 26.30\degree N was divided into cells of $0.25\degree\times0.25\degree$ (about $25\times28$\,km). Each cell was probed with the scene classification layer of its least cloudy Sentinel-2 image of 2024, and cells with 3--97\,\% water were kept as coastal. For each coastal cell the least cloudy image of every month of 2024 (tile cloud cover below 30\,\%) was processed at 60\,m with the pre-processing of Section~\ref{sec:preproc}, including the sun-glint removal, the deep-water reference and the turbidity index, and a month was counted as \emph{clear} when at least 70\,\% of the cell's water was cloud-free and the turbidity index did not exceed 0.010, the value above which the images of the main experiment were seldom calibrable. The three clearest months were then fused with the transfer method (Stumpf, calibrated on the training sites only) to a \emph{reconnaissance depth} with the site-conformal uncertainty, and the cells were mosaicked into one 60\,m GeoTIFF of the coast. Because turbid or deep water offshore of a coastal cell can be brighter than the darkest 2\,\% used as $\rho_\infty$, a stricter optically deep test than Eq.~\eqref{eq:snr} was also tried in this experiment: a pixel was treated as optically deep unless
\begin{equation}
\rho_{B03}-\rho_{\infty,B03}\ge 2\,\sigma_\infty,
\label{eq:sigmainf}
\end{equation}
where $\sigma_\infty$ is the spread of the green band over the deep-water set, and both tests were checked at the surveyed sites by the share of survey pixels deeper and shallower than the depth limit that each flagged. Because that check showed that neither test separates deep bright water from shallow water (Section~\ref{sec:coast}), the mosaic was finally checked against ICESat-2 along the lidar tracks. Every ATL24 pass from October 2022 to December 2025 over the coastal cells (at most eight per cell, within a total download budget of 6\,GB) was read with the photon screening of Section~\ref{sec:atl24methods}, reduced to the model LAT of the cell with the EOT20 tide at the pass time and corrected with Eq.~\eqref{eq:lidaroffset}; pixels in which the photons of a pass were scattered by more than 1\,m were discarded, and one lidar depth per 60\,m mosaic pixel was taken as the median over passes and beams. The mosaic was then compared with the lidar at the same pixels: the RMSE, bias and U95 coverage on lidar-shallow water (lidar depth within the 11\,m limit); the share of the mosaic depths that lie at pixels where the lidar finds the seabed deeper than the limit, which measures the deep-water contamination of the product; and the share of lidar-deep pixels that the mosaic correctly left without a depth. In each cell with at least 50 lidar pixels the mapped area was reduced by its contamination share to give a lidar-verified mapped area.

The lidar was then used not only to check the mosaic but to calibrate it, cell by cell, in the way a local-control image is calibrated on soundings. In each cell with at least 50 lidar pixels in at least five blocks, the visible depth was first determined from the lidar by the walk of Eq.~\eqref{eq:visible}, with the mosaic depth in place of the ratio, and the cell's depth limit set to the smaller of that and 11\,m. The mosaic depth $d_{\mathrm{map}}$ was then regressed on the lidar depth,
\begin{equation}
d_{\mathrm{cal}} = a + b\,d_{\mathrm{map}} + e\max\!\left(0,\,\kappa_c - d_{\mathrm{map}}\right),
\label{eq:coastcal}
\end{equation}
by Huber regression with the depth-balanced weights of Eq.~\eqref{eq:balance}, the hinge at $\kappa_c$, the median mosaic depth of the lidar pixels within $\pm1$\,m of 2.5\,m, being fitted only where at least 15 lidar pixels lay on each side of 2.5\,m and kept only with $e\le0$; a calibration was accepted only with $b>0$ and a weighted $R^2$ of at least 0.3, otherwise the cell kept the transfer depth. The calibration was tested on held-out spatial blocks of 480\,m: the blocks of a cell were assigned to five folds, the fit was made on the blocks of the other folds minus those adjacent to a test block, and every lidar pixel was thus predicted by a fit that never saw it or its neighbours, as in Section~\ref{sec:validation}. The held-out residuals gave the RMSE and bias per cell and per lidar depth band and the conformal interval of the calibrated product, computed as in Eqs.~\eqref{eq:score_conf}--\eqref{eq:u95new} with the largest quantile over the cells, the between-image spread $\sigma_b$ being recovered from the U95 layer of the mosaic. The calibration fitted on all lidar pixels of a cell was then applied to the cell, depths beyond the cell limit were flagged, and the cells were mosaicked into a second GeoTIFF carrying a band that marks the calibrated cells. The products of this experiment are the number of clear months per cell, the monthly turbidity climatology of the coast, the area of water that is optically shallow and that is mapped to the depth limit, the reconnaissance mosaic, its lidar check and the lidar-calibrated mosaic.

(2)~\emph{Unsurveyed areas with ICESat-2 only.} Two areas without any survey, Mirfa West (53.28--53.46\degree E, 24.04--24.17\degree N, a bay with sand flats west of the Mirfa test site) and Ras Ghanada (54.66--54.86\degree E, 24.84--24.98\degree N, the reef and shoal area east of Khalifa Port), were mapped from the images of December 2024 to May 2025 with the lidar-only mode of Section~\ref{sec:atl24methods}: every ATL24 pass over the area was reduced to LAT, corrected for the offset and screened by the plausibility test, the four best images were calibrated with the lidar pixels alone (Hybrid-local, with the visible depth determined from the lidar), and the depth, the site-conformal uncertainty and the quality flag were produced exactly as for a surveyed site.

(3)~\emph{Year-to-year repeatability.} For the five datasets with a natural seabed (the two Sir Bani Yas sites, Mirfa, the Khalifa Port limits and, as a dredged reference, Jebel Dhanna), the three best of the six least cloudy images of each clear season from December 2020 to May 2025 (five winters) were fused, and three quantities were computed per site and winter: the transfer error against the survey (Stumpf and LightGBM), the held-out local-control error of the hybrid, and, for every pair of winters $(u,v)$ at their $N_{uv}$ common pixels,
\begin{equation}
\Delta_{uv}=\sqrt{\frac{1}{N_{uv}}\sum_{k}\left(\hat d_k^{(u)}-\hat d_k^{(v)}\right)^2},
\label{eq:repeat}
\end{equation}
the root-mean-square difference between the fused transfer maps (ten pairs per site), which measures the repeatability of the product independently of any survey.

\subsection{Implementation}

The chain was implemented in Python 3.13 with numpy 2.1, pandas 2.2, rasterio 1.5, geopandas 1.2, odc-stac 0.5, eo-tides 0.10, LightGBM 4.7, scikit-learn and PyTorch. It runs on a Google Colab instance in about 1.5 hours for the main experiments once the processed images are cached (about five hours on first execution, including the seasonal images) and a further two hours for the outlook experiments, which are checkpointed so that an interrupted run resumes. All random seeds are fixed, and the complete configuration, the library versions and the SHA-256 digests of the sounding files are written to the run manifest.

\section{Results}
\label{sec:results}

\subsection{Effect of the validation design}
\label{sec:leakage}

Table~\ref{tab:valdesign} and Fig.~\ref{fig:methods}b show the same LightGBM model, with the same features and settings, scored under four validation designs on the training sites. Under a random 80/20 split of the pixels the RMSE was 0.57\,m with $R^2=0.93$, and 73\,\% of the pixels were within the Order~1a/1b TVU. Under spatial blocks of 450\,m with a 225\,m buffer the RMSE rose to 1.30\,m ($R^2=0.60$, 36\,\% within Order~1a/1b), and under leave-one-site-out to 1.66\,m ($R^2=0.34$, 26\,\%). Local control on held-out blocks gave 1.25\,m. The random split therefore understated the error of the model by a factor of 2.3 within a surveyed area and by a factor of 2.9 at a new site, and overstated the S-44 Order~1a/1b compliance by a factor of 2.8. All results reported hereafter use the spatially separated designs.

\begin{table}[htbp]
\centering
\caption{The same LightGBM model scored under four validation designs at the training sites (depths $\leq$11\,m). Errors in m; S-44 columns give the share of pixels within the TVU of the Order.}
\label{tab:valdesign}
\small
\begin{adjustbox}{max width=\textwidth}
\begin{tabular}{@{}lrrrrrrr@{}}
\toprule
Validation design & $n$ & RMSE & MAE & Bias & $R^2$ & Order 1a/1b (\%) & Order 2 (\%) \\
\midrule
Random pixel split (conventional) & 35,636 & 0.57 & 0.40 & $+0.03$ & 0.93 & 73.1 & 92.8 \\
Spatial blocks (450\,m) + buffer (225\,m) & 35,250 & 1.30 & 0.99 & $+0.13$ & 0.60 & 36.3 & 61.3 \\
Leave-one-site-out & 35,155 & 1.66 & 1.31 & $+0.22$ & 0.34 & 25.8 & 46.6 \\
Local control, held-out blocks & 16,815 & 1.25 & 0.96 & $0.00$ & 0.66 & 34.8 & 63.5 \\
\bottomrule
\end{tabular}
\end{adjustbox}
\end{table}

\subsection{Transfer accuracy of the four model families}
\label{sec:transfer}

\begin{figure}[htbp]
\centering
\includegraphics[width=\textwidth]{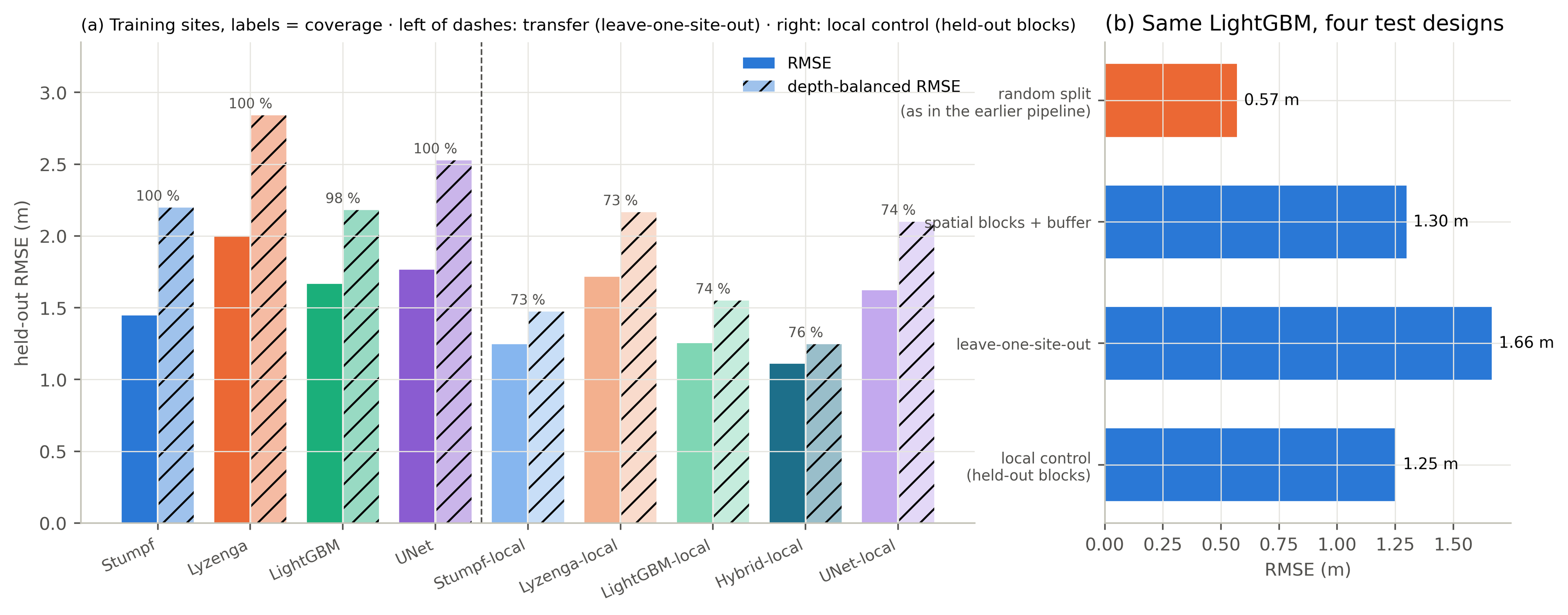}
\caption{(a) Held-out RMSE and depth-balanced RMSE of every method at the training sites; labels give the coverage. Left of the dashed line: transfer methods tested leave-one-site-out; right: local-control methods tested on blocks not used for calibration. (b) The same LightGBM model under four validation designs.}
\label{fig:methods}
\end{figure}

\begin{figure}[htbp]
\centering
\includegraphics[width=\textwidth]{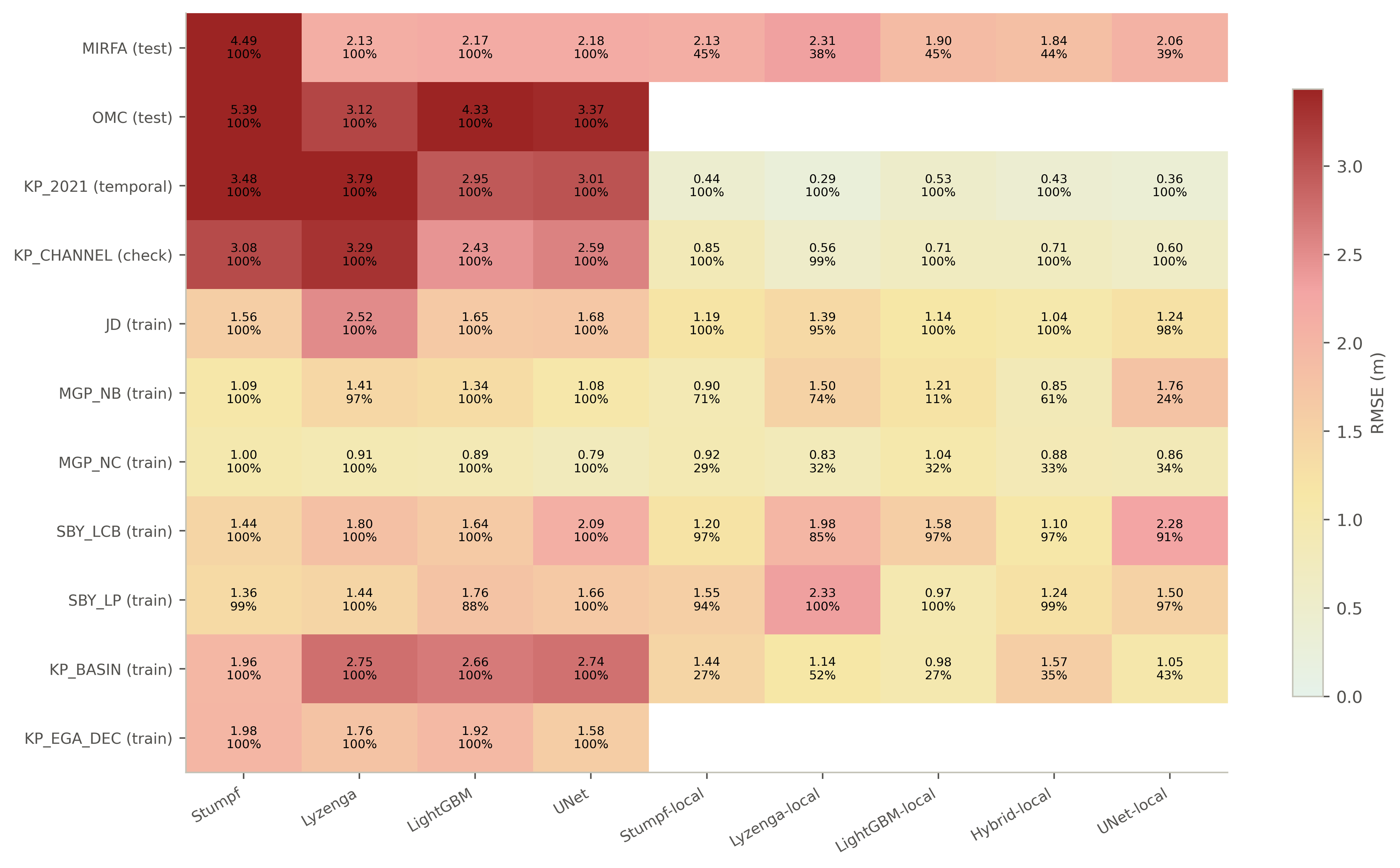}
\caption{Held-out RMSE (m, upper number) and coverage (\%, lower number) of every method at every dataset. Empty cells: no image could be calibrated.}
\label{fig:heatmap}
\end{figure}

In transfer mode, i.e.\ when each model was applied to a training site it had never seen (Table~\ref{tab:methods}; Fig.~\ref{fig:methods}a), the Stumpf ratio gave the lowest pooled RMSE, 1.45\,m (bias $+0.15$\,m, $R^2=0.52$), followed by LightGBM at 1.66\,m, the U-Net ensemble at 1.77\,m (leave-region-out) and the Lyzenga transform at 2.00\,m. All four covered 98--100\,\% of the in-range pixels. When the depth bands were weighted equally the tree model and the ratio were level: the depth-balanced RMSE was 2.18\,m for LightGBM and 2.20\,m for Stumpf, against 2.53\,m for the U-Net and 2.84\,m for Lyzenga, because the ratio transform, which is a straight line in $p_{23}$, over-estimated the depth of the 0--5\,m flats by 2--3.3\,m and under-estimated the 10--15\,m band by 2.1\,m (Table~\ref{tab:bands}), whereas the tree model was less biased at the extremes. Within a surveyed area (spatial block cross-validation), by contrast, LightGBM (1.30\,m) was better than Stumpf (1.45\,m) and Lyzenga (1.76\,m): the additional flexibility of the tree model is exploited when the training data cover the bottom types and water of the area, but does not transfer to an unseen site. At the level of individual sites (Table~\ref{tab:persite}; Fig.~\ref{fig:heatmap}) the transfer RMSE ranged from 0.79\,m (U-Net at the Mugharraq channel) to 2.75\,m (Lyzenga at the Khalifa Port basin); the ratio was the best transfer method at four of the seven calibrable training sites and the U-Net at the other three.

\begin{table}[htbp]
\centering
\caption{Held-out accuracy of every method at the training sites (depths $\leq$11\,m). Transfer methods are tested leave-one-site-out (the U-Net leave-region-out); local-control methods on spatial blocks not used for calibration. Balanced RMSE weights the four depth bands equally; coverage is the share of in-range pixels given a depth; deep flagged the share of pixels deeper than the limit correctly given no depth; block-CV the spatial block cross-validation inside the surveyed areas.}
\label{tab:methods}
\small
\begin{adjustbox}{max width=\textwidth}
\begin{tabular}{@{}llrrrrrrr@{}}
\toprule
Method & Mode & $n$ & RMSE (m) & Bias (m) & Balanced RMSE (m) & Coverage (\%) & Deep flagged (\%) & Block-CV RMSE (m) \\
\midrule
Stumpf & transfer & 35,668 & 1.45 & $+0.15$ & 2.20 & 99.7 & 0 & 1.45 \\
Lyzenga & transfer & 35,641 & 2.00 & $+0.40$ & 2.84 & 99.7 & 0 & 1.76 \\
LightGBM & transfer & 35,155 & 1.66 & $+0.22$ & 2.18 & 98.3 & 1.3 & 1.30 \\
U-Net & transfer & 35,762 & 1.77 & $-0.04$ & 2.53 & 100.0 & 0 & -- \\
\midrule
Stumpf-local & local control & 16,740 & 1.24 & $-0.08$ & 1.47 & 73.4 & 74.8 & -- \\
Lyzenga-local & local control & 16,726 & 1.72 & $-0.31$ & 2.16 & 73.3 & 75.4 & -- \\
LightGBM-local & local control & 16,815 & 1.25 & $0.00$ & 1.55 & 73.7 & 74.9 & -- \\
Hybrid-local & local control & 17,280 & \textbf{1.11} & $-0.13$ & \textbf{1.25} & 75.7 & 75.5 & -- \\
UNet-local & local control & 16,851 & 1.62 & $+0.11$ & 2.10 & 73.8 & 77.8 & -- \\
\bottomrule
\end{tabular}
\end{adjustbox}
\end{table}

\subsection{Accuracy with local control}
\label{sec:local}

\begin{figure}[htbp]
\centering
\includegraphics[width=\textwidth]{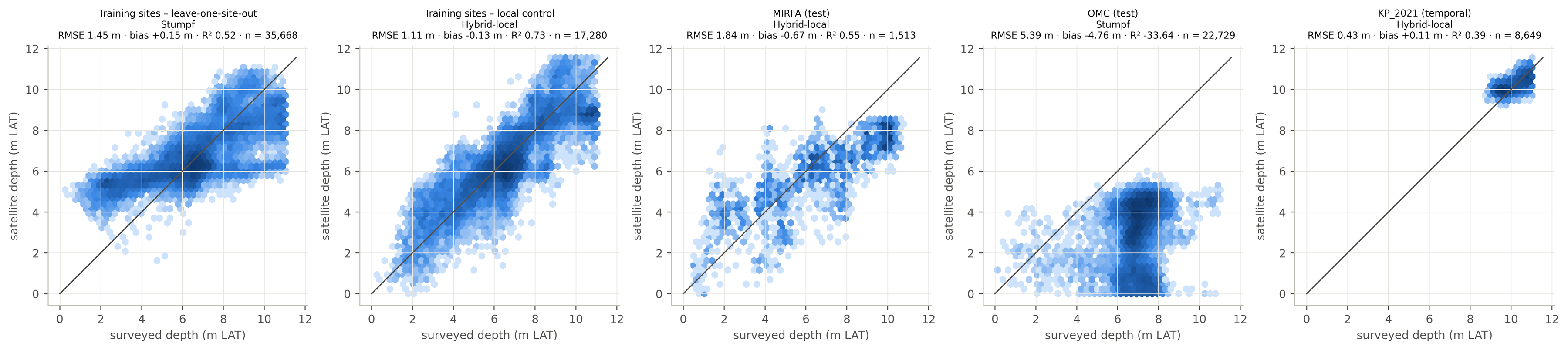}
\caption{Satellite-derived against surveyed depth (10\,m pixels, held-out predictions): training sites in transfer mode (Stumpf) and with local control (Hybrid-local), the two independent sites with their map method, and the temporal site with local control. Colour: number of pixels (logarithmic); line: 1:1.}
\label{fig:scatter}
\end{figure}

Calibrating every image with about 20\,\% of the site's soundings, supplemented by the verified ICESat-2 pixels where they existed, and testing on the remaining blocks beyond the buffer reduced the pooled RMSE of every family (Table~\ref{tab:methods}). Hybrid-local reached 1.11\,m (bias $-0.13$\,m, $R^2=0.73$, NMAD 1.00\,m, $P_{95}=2.23$\,m), Stumpf-local 1.24\,m, LightGBM-local 1.25\,m, UNet-local 1.62\,m and Lyzenga-local 1.72\,m. The hybrid also had the best depth-balanced RMSE, 1.25\,m against 1.47--2.16\,m for the others, because its ratio and hinge terms allowed it to follow the controls on the 0--2\,m flats (RMSE 1.04\,m, bias $+0.51$\,m) where the re-scaled tree output could not. Local control costs coverage: because an image is used only where its colour was shown to follow depth, 73--76\,\% of the in-range pixels received a depth, and 75--78\,\% of the pixels deeper than the limit were correctly refused, against 0--1\,\% in transfer mode, which publishes a depth everywhere within the limit. Per site, the local-control RMSE of the best method ranged from 0.83\,m (Lyzenga-local at the Mugharraq channel) and 0.85\,m (hybrid at the Mugharraq basin) to 1.10\,m (hybrid at the Sir Bani Yas landing-craft basin) at the natural and clear-water sites, and reached 0.98\,m (LightGBM-local) on the 12\,\% of the Khalifa Port basin that lies within the depth limit (Table~\ref{tab:persite}). The scatter of the held-out predictions (Fig.~\ref{fig:scatter}) shows the residual compression of the depth range that remains after local control, with the shallowest water predicted slightly too deep and the 10--11\,m water too shallow.

\begin{table}[htbp]
\centering
\caption{Held-out RMSE (m) per dataset and method. Transfer results at the training sites are leave-one-site-out (U-Net: leave-region-out); local-control results use blocks not used for calibration. Empty cells: no image could be calibrated (seabed not readable). Bold: lowest RMSE per dataset.}
\label{tab:persite}
\small
\setlength{\tabcolsep}{4pt}
\begin{adjustbox}{max width=\textwidth}
\begin{tabular}{@{}lrrrrrrrrr@{}}
\toprule
 & \multicolumn{4}{c}{Transfer} & \multicolumn{5}{c}{Local control} \\
\cmidrule(lr){2-5}\cmidrule(lr){6-10}
Dataset & Stumpf & Lyzenga & LightGBM & U-Net & Stumpf-l. & Lyzenga-l. & LightGBM-l. & Hybrid-l. & UNet-l. \\
\midrule
MIRFA (independent) & 4.49 & 2.13 & 2.17 & 2.18 & 2.13 & 2.31 & 1.90 & \textbf{1.84} & 2.06 \\
OMC (independent) & 5.39 & \textbf{3.12} & 4.33 & 3.37 & -- & -- & -- & -- & -- \\
KP\_2021 (temporal) & 3.48 & 3.79 & 2.95 & 3.01 & 0.44 & \textbf{0.29} & 0.53 & 0.43 & 0.36 \\
KP\_CHANNEL (check) & 3.08 & 3.29 & 2.43 & 2.59 & 0.85 & \textbf{0.56} & 0.71 & 0.71 & 0.60 \\
JD (train) & 1.56 & 2.52 & 1.65 & 1.68 & 1.19 & 1.39 & 1.14 & \textbf{1.04} & 1.24 \\
MGP\_NB (train) & 1.09 & 1.41 & 1.34 & 1.08 & 0.90 & 1.50 & 1.21 & \textbf{0.85} & 1.76 \\
MGP\_NC (train) & 1.00 & 0.91 & 0.89 & \textbf{0.79} & 0.92 & 0.83 & 1.04 & 0.88 & 0.86 \\
SBY\_LCB (train) & 1.44 & 1.80 & 1.64 & 2.09 & 1.20 & 1.98 & 1.58 & \textbf{1.10} & 2.28 \\
SBY\_LP (train) & 1.36 & 1.44 & 1.76 & 1.66 & 1.55 & 2.33 & \textbf{0.97} & 1.24 & 1.50 \\
KP\_BASIN (train) & 1.96 & 2.75 & 2.66 & 2.74 & 1.44 & 1.14 & \textbf{0.98} & 1.57 & 1.05 \\
KP\_EGA\_DEC (train) & 1.98 & 1.76 & 1.92 & \textbf{1.58} & -- & -- & -- & -- & -- \\
\bottomrule
\end{tabular}
\end{adjustbox}
\end{table}

\subsection{Error by depth band and the achievable depth}
\label{sec:depthbands}

\begin{figure}[htbp]
\centering
\includegraphics[width=\textwidth]{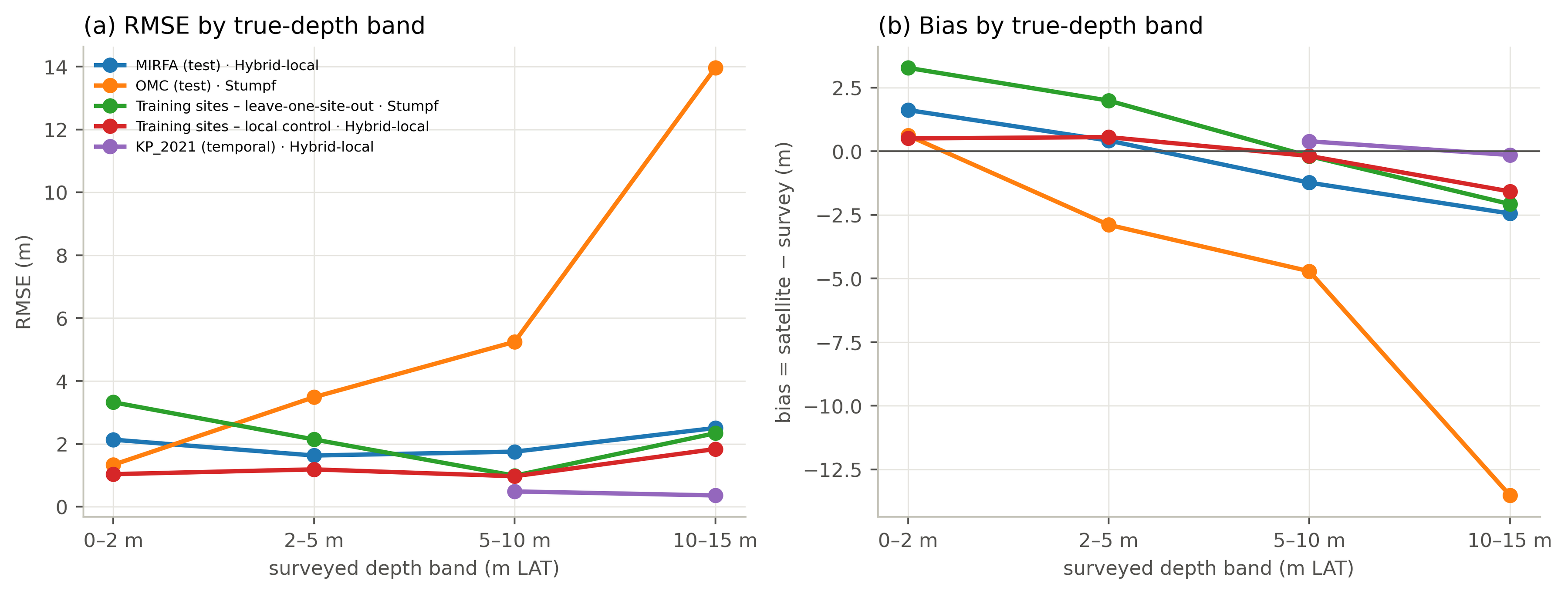}
\caption{RMSE (a) and bias (b) by surveyed-depth band for the transfer and local-control map methods at the training sites and at the independent and temporal sites. Positive bias: satellite depth too deep.}
\label{fig:bands}
\end{figure}

\begin{figure}[htbp]
\centering
\includegraphics[width=\textwidth]{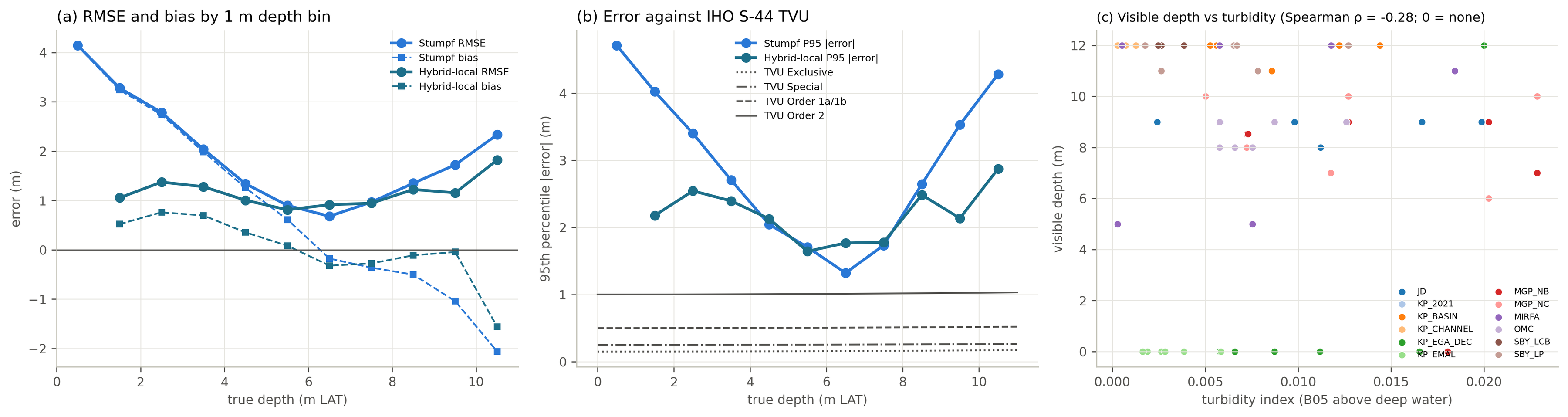}
\caption{(a) RMSE and bias by 1\,m bin of true depth at the training sites for the Stumpf transfer and the Hybrid-local methods; (b) 95th percentile of the absolute error against the IHO S-44 TVU of each Order; (c) visible depth of each image against its turbidity index (images with no colour--depth relation plotted at zero).}
\label{fig:depthlimits}
\end{figure}

Table~\ref{tab:bands} and Fig.~\ref{fig:bands} give the error by surveyed-depth band. In transfer mode the Stumpf ratio was accurate only in the 5--10\,m band, where 73\,\% of the training pixels lie (RMSE 0.99\,m, bias $-0.19$\,m, 43\,\% within Order~1a/1b and 74\,\% within Order~2); in the 0--2 and 2--5\,m bands it was biased by $+3.3$ and $+2.0$\,m, and in the 10--15\,m band by $-2.1$\,m. With local control the hybrid was nearly uniform across the bands to 10\,m, with RMSE values of 1.04, 1.19, 0.97 and 1.84\,m and biases of $+0.51$, $+0.56$, $-0.18$ and $-1.58$\,m in the 0--2, 2--5, 5--10 and 10--15\,m bands, and 71, 61, 74 and 19\,\% of the pixels within Order~2. The finer analysis by 1\,m bins (Fig.~\ref{fig:depthlimits}a) shows the transfer error reaching its minimum of 0.68\,m at 6--7\,m and increasing on both sides, whereas the hybrid remained between 0.81 and 1.37\,m from 1 to 10\,m and rose to 1.82\,m in the 10--11\,m bin, where its bias was $-1.56$\,m. Against the S-44 criterion of $P_{95}\leq\mathrm{TVU}$ (Fig.~\ref{fig:depthlimits}b), no method met any Order in any 1\,m bin: the lowest $P_{95}$ was 1.32\,m for Stumpf at 6--7\,m and 1.64\,m for the hybrid at 5--6\,m, against a TVU of 1.01\,m for Order~2 at those depths.

The visible depth of the images, i.e.\ the depth to which the colour still followed the soundings, was 8--9\,m at Jebel Dhanna and Old Musaffah, 6--10\,m at Mugharraq, 11--12\,m at Sir Bani Yas, at Mirfa in four of six images and at the Khalifa Port limits, and undefined (no colour--depth relation) in every image of the EGA/EMAL berths and in five of six images of the December 2022 EGA basin survey (Fig.~\ref{fig:depthlimits}c); at the Khalifa Port basin the walk reached 12\,m on the shallow margin of the basin, which holds the 12\,\% of its pixels within the limit. Across the 72 images the visible depth was only weakly related to the turbidity index (Spearman $\rho=-0.28$), because the images that could not be calibrated were those over basins whose entire depth range exceeds the optical reach rather than the most turbid ones. The 95th percentile of the training depths within the visible depths, 11\,m, was adopted as the transfer depth limit; 88--99\,\% of the pixels of the Khalifa Port basin and berth surveys and 84\,\% of the 2021 port-limit survey lie beyond it (Table~\ref{tab:datasets}).

\begin{table}[htbp]
\centering
\caption{Accuracy by surveyed-depth band for the transfer map method (Stumpf) and the local-control map method (Hybrid-local) at the training sites, and at the independent and temporal sites. $n$ = held-out pixels; S-44 columns = share within the TVU.}
\label{tab:bands}
\small
\begin{adjustbox}{max width=\textwidth}
\begin{tabular}{@{}llrrrrrr@{}}
\toprule
Set & Method & Band (m) & $n$ & RMSE (m) & MAE (m) & Bias (m) & Order 1a/1b / 2 (\%) \\
\midrule
Training, leave-one-site-out & Stumpf & 0--2 & 477 & 3.32 & 3.28 & $+3.28$ & 0.0 / 0.0 \\
 & & 2--5 & 6,825 & 2.14 & 2.00 & $+1.99$ & 1.2 / 10.1 \\
 & & 5--10 & 26,061 & 0.99 & 0.76 & $-0.19$ & 42.6 / 73.5 \\
 & & 10--15 & 2,305 & 2.34 & 2.09 & $-2.07$ & 3.4 / 13.7 \\
Training, local control & Hybrid-local & 0--2 & 212 & 1.04 & 0.78 & $+0.51$ & 46.2 / 70.8 \\
 & & 2--5 & 3,474 & 1.19 & 0.95 & $+0.56$ & 33.9 / 60.9 \\
 & & 5--10 & 12,251 & 0.97 & 0.75 & $-0.18$ & 43.2 / 73.6 \\
 & & 10--15 & 1,343 & 1.84 & 1.68 & $-1.58$ & 10.0 / 18.6 \\
MIRFA (independent) & Hybrid-local & 0--2 & 134 & 2.13 & 1.79 & $+1.62$ & 14.9 / 31.3 \\
 & & 2--5 & 374 & 1.63 & 1.37 & $+0.42$ & 19.5 / 40.9 \\
 & & 5--10 & 876 & 1.75 & 1.44 & $-1.23$ & 22.6 / 37.4 \\
 & & 10--15 & 129 & 2.51 & 2.44 & $-2.44$ & 0.0 / 0.0 \\
OMC (independent) & Stumpf & 0--2 & 67 & 1.34 & 1.10 & $+0.62$ & 23.9 / 49.3 \\
 & & 2--5 & 491 & 3.49 & 2.94 & $-2.89$ & 6.7 / 17.1 \\
 & & 5--10 & 21,914 & 5.25 & 4.71 & $-4.71$ & 0.0 / 0.2 \\
 & & 10--15 & 257 & 13.96 & 13.52 & $-13.52$ & 0.0 / 0.0 \\
KP\_2021 (temporal) & Hybrid-local & 5--10 & 4,171 & 0.49 & 0.41 & $+0.39$ & 67.0 / 98.1 \\
 & & 10--15 & 4,478 & 0.36 & 0.29 & $-0.14$ & 84.8 / 99.7 \\
\bottomrule
\end{tabular}
\end{adjustbox}
\end{table}

\subsection{Independent and temporal test sites}
\label{sec:independent}

At Mirfa, a natural bay with extensive sand flats shallower than 5\,m, the six images were taken from the clear season (December 2022 to January 2023, 53 days after the survey) and the hybrid with local control gave 1.84\,m (bias $-0.67$\,m, $R^2=0.55$), the best result of any method at the site, with LightGBM-local at 1.90\,m, UNet-local at 2.06\,m, Stumpf-local at 2.13\,m and Lyzenga-local at 2.31\,m (Table~\ref{tab:persite}); two of the six images, whose visible depth fell to 5\,m, could not be calibrated, and the local methods covered 44\,\% of the in-range pixels. In transfer mode the Lyzenga transform (2.13\,m, bias $+0.49$\,m), LightGBM (2.17\,m) and the U-Net (2.18\,m) were close to each other, whereas the Stumpf ratio failed with 4.49\,m and a bias of $-3.74$\,m: in the very clear winter water of this bay the blue/green ratio of the 5--12\,m water is lower than at any training site, and the single straight line fitted at the training sites reads it as shallow. The hybrid's error was concentrated at the extremes of the depth range, $+1.6$\,m in the 0--2\,m band and $-2.4$\,m in the 10--15\,m band, with 1.6--1.8\,m in between (Table~\ref{tab:bands}). The seasonal experiment (Section~\ref{sec:seasons}) shows that the same site calibrated on its autumn images, i.e.\ those closest to the survey, could not be calibrated at all, and on summer images gave 3.0\,m. At Old Musaffah, a turbid dredged channel of 5--10\,m, every method failed: the transfer methods were biased by $-2.8$ to $-4.8$\,m (Lyzenga 3.12\,m RMSE, the others 3.4--5.4\,m), the held-out $R^2$ of the best method was $-10.6$, no image could be locally calibrated (five fits below the $R^2$ criterion and one with the wrong sign), and, correctly, the pipeline withheld the map (Section~\ref{sec:products}). The colour--depth rank correlation of the OMC images was 0.13--0.20, against 0.5--0.8 at the clear sites; the seabed of the channel was not visible in any image although the pixels passed the signal-to-noise test.

The temporal test gave the opposite picture. At the Khalifa Port limits, surveyed in April--May 2021 over natural seabed of 8.5--11\,m within the depth limit, the transfer methods were biased by $-2.9$ to $-3.8$\,m (RMSE 2.95--3.79\,m), because the clear, deep offshore water of this site is unlike the water of any training basin; but every local-control method gave an RMSE between 0.29\,m (Lyzenga-local, $R^2=0.72$) and 0.53\,m, with 100\,\% coverage of the tested pixels, 92\,\% of the pixels within Order~1a/1b and 99.8\,\% within Order~2 for the best method, and 76 and 99\,\% for the hybrid. The shoal-biased Khalifa Port channel grid, used as a check only, gave 0.56\,m for Lyzenga-local, 0.60\,m for UNet-local, 0.71\,m for LightGBM-local and the hybrid and 0.85\,m for Stumpf-local.

\subsection{Per-site method choice and map products}
\label{sec:products}

\begin{figure}[htbp]
\centering
\includegraphics[width=\textwidth]{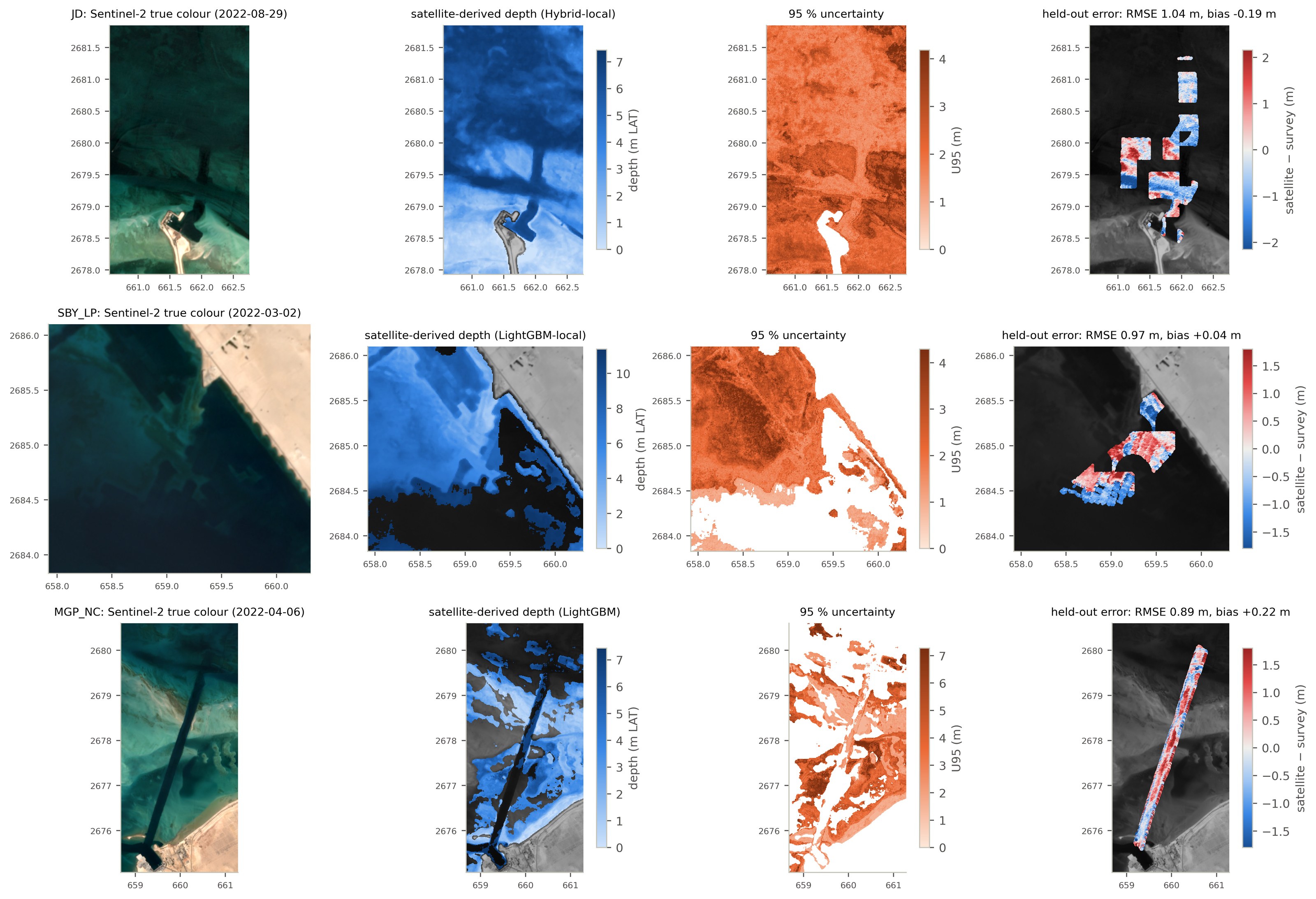}
\caption{Showcase products at Jebel Dhanna (JD), the Sir Bani Yas logistics port (SBY\_LP) and the Mugharraq channel (MGP\_NC). Columns: Sentinel-2 true-colour image; satellite-derived depth below LAT (grey: land, optically deep or beyond the visible depth); 95\,\% uncertainty; held-out error at the survey pixels. Coordinates in km (UTM). Contains modified Copernicus Sentinel data (2022).}
\label{fig:maps}
\end{figure}

\begin{figure}[htbp]
\centering
\includegraphics[width=0.75\textwidth]{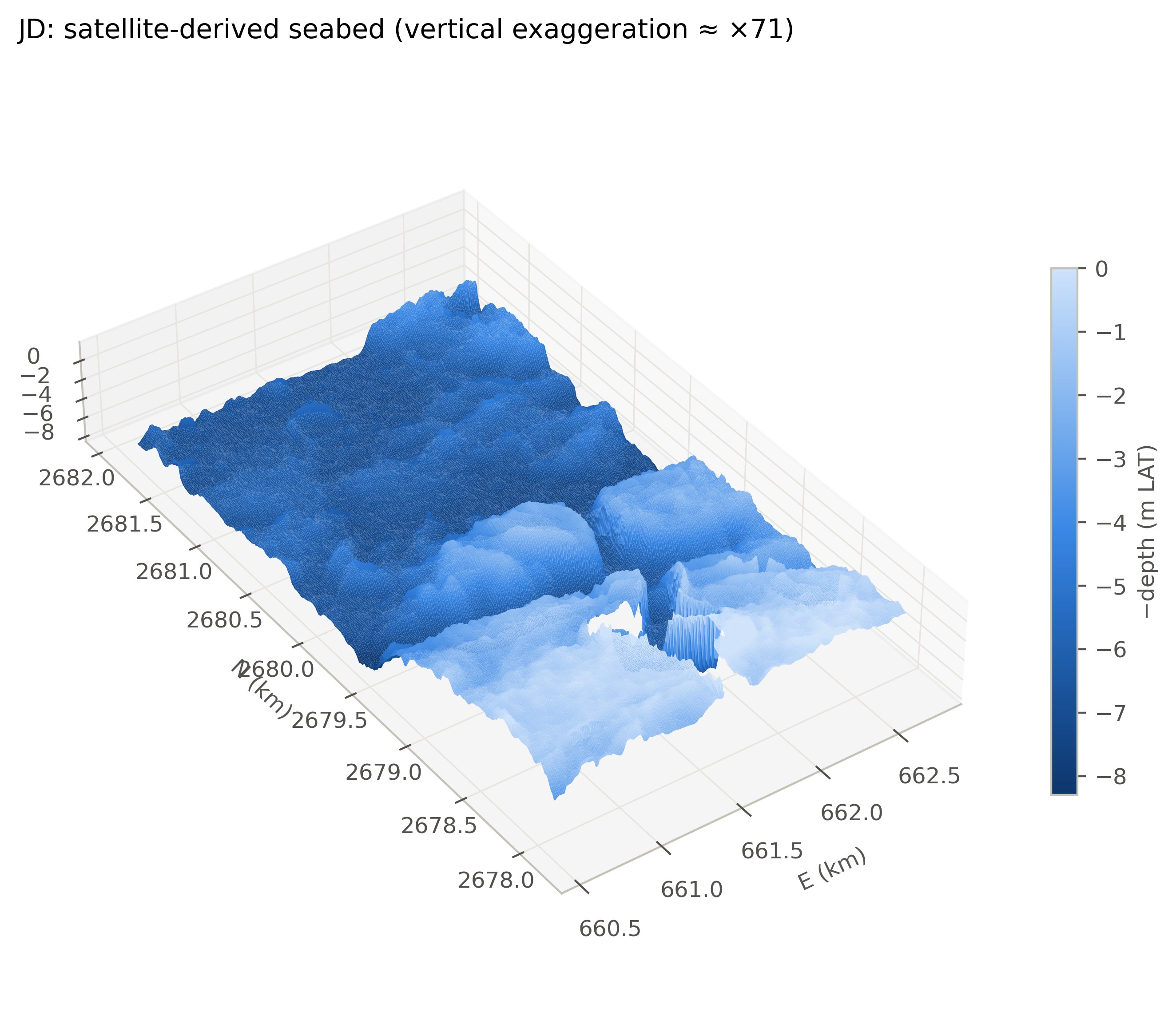}
\caption{Three-dimensional view of the satellite-derived seabed at Jebel Dhanna (vertical exaggeration about $\times$71).}
\label{fig:3d}
\end{figure}

Because no method was best everywhere, the map of each site was produced with the method that won its own held-out test among the candidates that covered at least half of the pixels seen within the visible depth (Table~\ref{tab:choice}). Local control won at the clear basins of Jebel Dhanna and Sir Bani Yas (hybrid), at the Sir Bani Yas logistics port (LightGBM-local) and at the Khalifa Port channel (Lyzenga-local), with coverage of 52--69\,\% of the pixels within the visible depth. Transfer won at the Mugharraq basin and channel (Stumpf and LightGBM, held-out RMSE 1.09 and 0.89\,m) and at Mirfa (Lyzenga, 2.13\,m), where it covered 100\,\%. At the Khalifa Port basin, at the December 2022 EGA basin survey and at Old Musaffah the best eligible held-out $R^2$ was negative ($-0.34$, $-0.71$ and $-10.6$) and no map was published. At the Khalifa Port limits the local methods, although accurate to 0.3--0.5\,m, were tested on 48\,\% of the visible pixels, just below the 50\,\% eligibility threshold of the rule; the rule therefore fell back on the transfer methods, which had no skill at this site, and withheld the map. The rule is designed to err on this side, but the case shows that a coverage threshold applied to a site where the images read the seabed over only half of its area removes a usable product; the local-control maps of this site are available with their held-out statistics, and the rule should be relaxed for sites of this type. The showcase products (Fig.~\ref{fig:maps}) illustrate the three columns of information delivered with every map: the depth, the 95\,\% uncertainty, which is wider where the images disagree and where the water is deeper, and the held-out error at the survey pixels. At Jebel Dhanna the dredged basin and approach channel are resolved with a held-out RMSE of 1.04\,m, and the three-dimensional rendering of the same product (Fig.~\ref{fig:3d}) shows the basin, the channel and the flats; the median filter of Section~\ref{sec:fusion} has removed most of the pixel-to-pixel noise that the vertical exaggeration would otherwise amplify. The full product panels of all twelve datasets, including the quality flag and the estimated S-44 Order, are given in the supplementary material (Figs.~\ref{fig:panels} and \ref{fig:panels2}).

\begin{table}[htbp]
\centering
\caption{Map method chosen at each dataset from its own held-out test, with the held-out RMSE and $R^2$ of the winner and the coverage within the visible depth. No map is published where no eligible method beats a constant depth ($R^2\leq0$).}
\label{tab:choice}
\small
\begin{adjustbox}{max width=\textwidth}
\begin{tabular}{@{}llrrrl@{}}
\toprule
Dataset & Method & RMSE (m) & $R^2$ & Coverage (\%) & Map mode \\
\midrule
JD & Hybrid-local & 1.04 & 0.55 & 52.3 & local control \\
MGP\_NB & Stumpf & 1.09 & 0.65 & 100.0 & transfer \\
MGP\_NC & LightGBM & 0.89 & 0.54 & 100.0 & transfer \\
SBY\_LCB & Hybrid-local & 1.10 & 0.80 & 69.2 & local control \\
SBY\_LP & LightGBM-local & 0.97 & 0.77 & 64.4 & local control \\
KP\_BASIN & (Stumpf) & 1.96 & $-0.34$ & 100.0 & withheld (no skill) \\
KP\_EGA\_DEC & (Lyzenga) & 1.76 & $-0.71$ & 100.0 & withheld (no skill) \\
MIRFA & Lyzenga & 2.13 & 0.33 & 100.0 & transfer \\
OMC & (Lyzenga) & 3.12 & $-10.6$ & 100.0 & withheld (no skill) \\
KP\_2021 & (LightGBM) & 2.95 & $-29.0$ & 100.0 & withheld (local methods below the coverage threshold) \\
KP\_CHANNEL & Lyzenga-local & 0.56 & 0.86 & 66.3 & local control \\
\bottomrule
\end{tabular}
\end{adjustbox}
\end{table}

\subsection{Calibration of the uncertainty and S-44 compliance}
\label{sec:uncertainty}

\begin{figure}[htbp]
\centering
\includegraphics[width=0.55\textwidth]{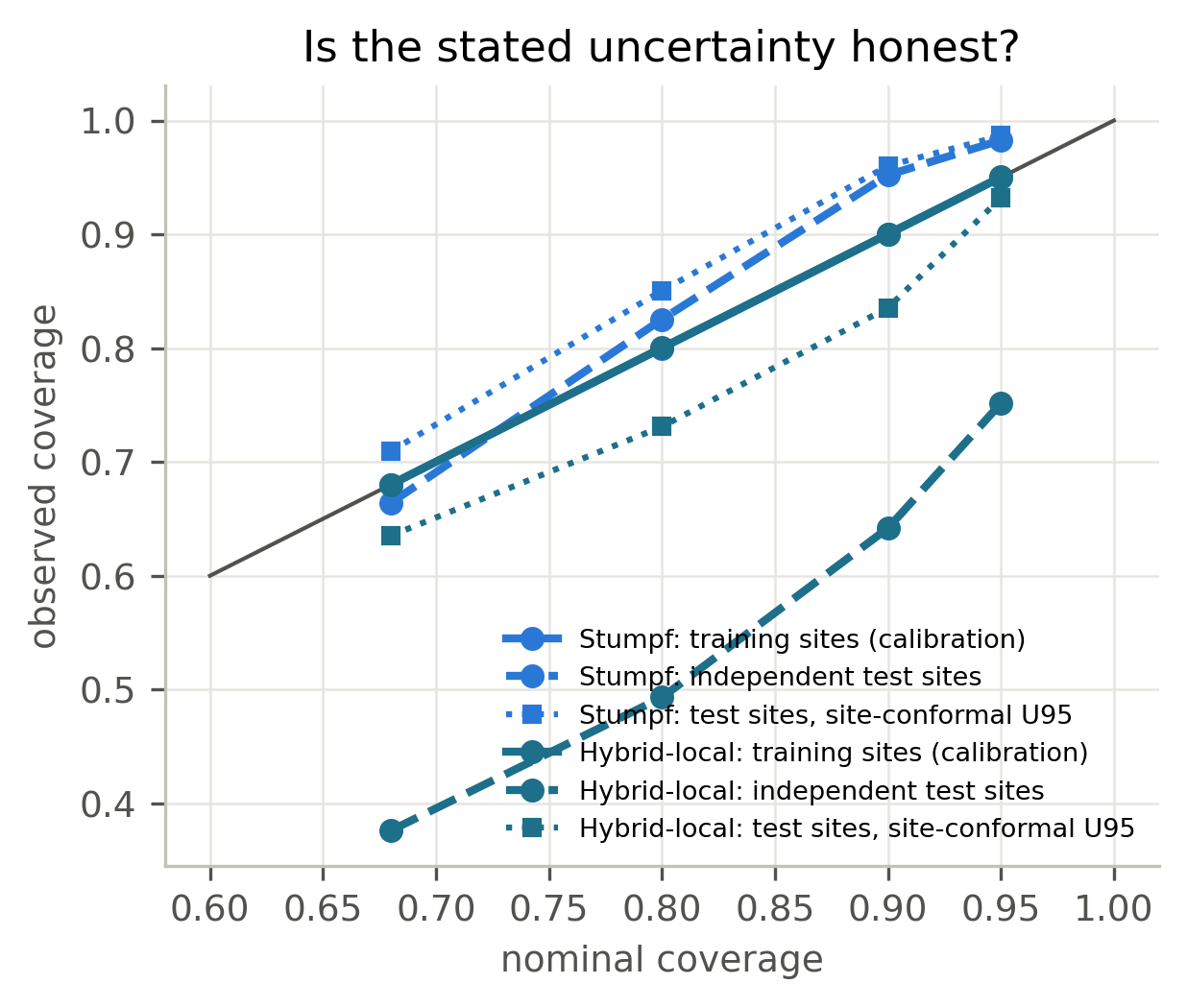}
\caption{Calibration of the conformal uncertainty: observed share of surveyed depths inside the interval against the nominal level, for the training sites (on which the pooled interval was calibrated) and for the independent test sites with the pooled and with the site-conformal interval.}
\label{fig:conformal}
\end{figure}

The pooled conformal 95\,\% intervals were well calibrated where they were constructed and where the water resembled the training data: on the training sites the observed coverage of the U95 interval was 95.0\,\% for every method (Table~\ref{tab:s44}), and it followed the nominal level at 68, 80 and 90\,\% to within 1\,\% (Fig.~\ref{fig:conformal}). At the temporal site the observed coverage was 92--100\,\% for the local methods and 99--100\,\% for the transfer methods. At the independent sites the pooled intervals of the transfer methods were conservative (98--100\,\% at Mirfa and Old Musaffah) but those of the local methods were too narrow at Mirfa, 74--85\,\% for the three ratio-based methods and 75\,\% for the hybrid. The site-conformal interval, which takes the quantile of the worst training site, repaired this: its observed coverage at Mirfa was 93\,\% for the hybrid and 99.9\,\% for the ratio, and it followed the nominal level at the lower levels (Fig.~\ref{fig:conformal}). The pooled interval thus reports the error faithfully within the population it was calibrated on, and the site-conformal interval is the one to attach to a product for an area that has not been surveyed.

By the S-44 grading, Sentinel-2 SDB does not reach survey standards in these waters without local control, and reaches Order~1a/1b only in favourable conditions with it. Pooled over the training sites, 32\,\% of the Stumpf transfer pixels and 39\,\% of the hybrid local-control pixels were within the Order~1a/1b TVU, and 57 and 67\,\% within Order~2; the Exclusive and Special Orders were met by 11--13 and 17--20\,\%. Only at the Khalifa Port limits, in clear water of 8.5--11\,m with local control, did a method approach the 95\,\% criterion for Order~1a/1b (92\,\% for Lyzenga-local, 76\,\% for the hybrid) and exceed it for Order~2 (99.8 and 98.9\,\%).

\begin{table}[htbp]
\centering
\caption{IHO S-44 compliance and uncertainty calibration of the map methods: share of held-out pixels within the TVU of each Order (an Order is met at $\geq$95\,\%) and observed coverage of the pooled conformal 95\,\% interval (U95) and, where it applies, of the site-conformal interval (U95$_{\mathrm{new}}$).}
\label{tab:s44}
\small
\begin{adjustbox}{max width=\textwidth}
\begin{tabular}{@{}llrrrrrrr@{}}
\toprule
Set & Method & $n$ & Exclusive (\%) & Special (\%) & Order 1a/1b (\%) & Order 2 (\%) & U95 (\%) & U95$_{\mathrm{new}}$ (\%) \\
\midrule
Training, leave-one-site-out & Stumpf & 35,668 & 10.5 & 16.6 & 31.6 & 56.5 & 95.0 & -- \\
Training, leave-one-site-out & LightGBM & 35,155 & 8.3 & 13.3 & 25.8 & 46.6 & 95.0 & -- \\
Training, leave-one-site-out & U-Net & 35,762 & 9.6 & 15.2 & 28.8 & 49.3 & 95.0 & -- \\
Training, local control & Hybrid-local & 17,280 & 12.7 & 20.2 & 38.8 & 66.7 & 95.0 & -- \\
Training, local control & LightGBM-local & 16,815 & 10.8 & 17.8 & 34.8 & 63.5 & 95.0 & -- \\
KP\_2021 (temporal) & Lyzenga-local & 8,649 & 45.2 & 64.7 & 92.2 & 99.8 & 100.0 & -- \\
KP\_2021 (temporal) & Hybrid-local & 8,649 & 27.6 & 42.3 & 76.2 & 98.9 & 100.0 & -- \\
KP\_CHANNEL (check) & Hybrid-local & 635 & 15.1 & 22.8 & 44.7 & 86.5 & 100.0 & -- \\
MIRFA (independent) & Stumpf & 5,659 & 2.7 & 4.7 & 10.0 & 19.6 & 99.8 & 99.9 \\
MIRFA (independent) & Lyzenga & 5,659 & 3.4 & 5.6 & 12.7 & 29.1 & 99.1 & -- \\
MIRFA (independent) & Hybrid-local & 1,513 & 5.4 & 9.4 & 19.2 & 34.6 & 75.1 & 93.2 \\
OMC (independent) & Stumpf & 22,729 & 0.1 & 0.1 & 0.2 & 0.7 & 97.9 & 98.4 \\
OMC (independent) & Lyzenga & 22,727 & 0.4 & 0.6 & 1.4 & 5.4 & 98.2 & -- \\
\bottomrule
\end{tabular}
\end{adjustbox}
\end{table}

\subsection{ICESat-2 lidar as an independent check and as a substitute for soundings}
\label{sec:atl24results}

\begin{figure}[htbp]
\centering
\includegraphics[width=\textwidth]{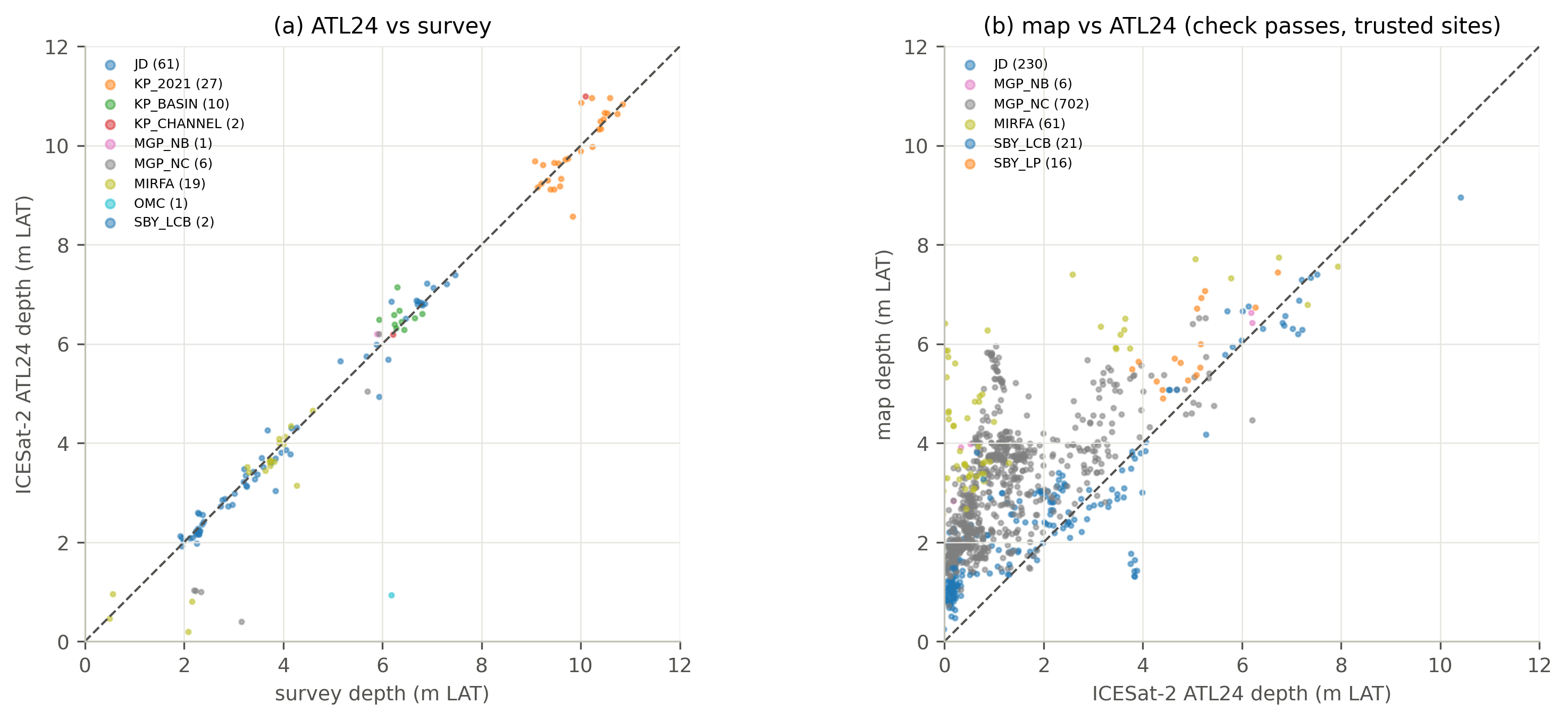}
\caption{Independent check with ICESat-2 ATL24: (a) offset-corrected lidar against survey on pixels with soundings; (b) map against the lidar of the check passes on pixels without soundings, at the trusted sites.}
\label{fig:atl24}
\end{figure}

Of the 170 site--pass combinations, 13 overlapped a survey by three or more pixels, ten were verified and three were rejected (Section~\ref{sec:atl24methods}). Before correction, the verified lidar read deeper than the MBES surveys at every site, by a median of $+0.36$ to $+0.64$\,m at Jebel Dhanna and up to $+0.77$\,m at the 10\,m depths of the Khalifa Port limits; the robust fit of this offset gave $+0.28$\,m $+3.5\,\%$ of depth, with a residual standard deviation of 0.16\,m. After its removal the lidar agreed with the surveys to an RMSE of 0.25\,m at Jebel Dhanna ($n=61$, bias $-0.01$\,m, $R^2=0.98$, 92\,\% within Order~1a/1b and 100\,\% within Order~2), 0.37\,m at the Khalifa Port basin ($n=10$), 0.61\,m at Mirfa ($n=19$) and 0.94\,m at the Khalifa Port limits ($n=30$, 2.0 years from the survey) (Table~\ref{tab:atl24}; Fig.~\ref{fig:atl24}a). The median absolute time between the lidar passes and the surveys was 339--730 days, so that part of the remaining difference at the dredged sites may be real seabed change.

The comparison of the maps with the lidar of the check passes, away from the soundings and never used for calibration (Fig.~\ref{fig:atl24}b), extends the validation to parts of the sites that the vessels did not cover, which are predominantly the shallow margins. At Jebel Dhanna the local-control map agreed with the lidar to 0.50\,m on the 61 pixels that also hold soundings and to 0.96\,m with a bias of $+0.74$\,m on 221 pixels without soundings; at the Sir Bani Yas sites the maps agreed to 1.1--1.8\,m on 16--20 pixels; at Mirfa the transfer map was 3.4\,m too deep on 58 shallow lidar pixels; and at the Mugharraq channel the transfer map was 1.9\,m too deep on 699 lidar pixels shallower than 2\,m on the channel margins, where the map coverage was 54\,\%. The lidar thus revealed, consistently with the depth-band analysis, that the transfer maps over-estimate the depth of water shallower than 2\,m that the surveys do not sample by 2--3\,m, and that the local-control map with the hinge and the lidar controls reduced this over-estimation to 0.7\,m at Jebel Dhanna.

\begin{figure}[htbp]
\centering
\includegraphics[width=0.9\textwidth]{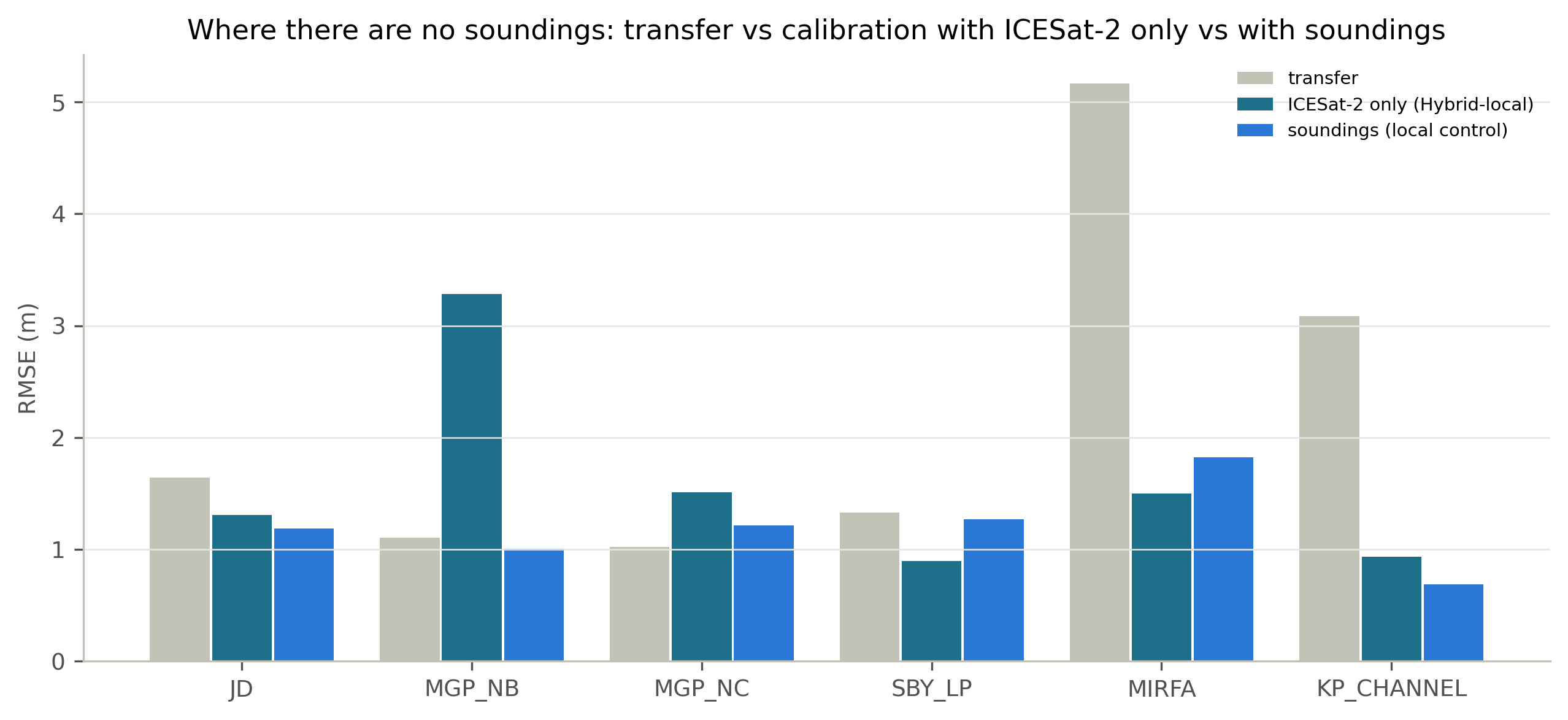}
\caption{Where there are no soundings: RMSE of the transfer method, of calibration with ICESat-2 only and of calibration with soundings, scored on survey pixels more than 225\,m from any lidar pixel.}
\label{fig:lidaronly}
\end{figure}

Calibrating the images with lidar alone (Table~\ref{tab:lidaronly}; Fig.~\ref{fig:lidaronly}) approached the sounding-based accuracy at the sites where the calibration passes crossed a representative depth range: 1.31\,m at Jebel Dhanna (756 lidar pixels over $-0.5$ to 8.3\,m, 94\,\% of the test pixels covered) against 1.19\,m with soundings, 0.90\,m at the Sir Bani Yas logistics port (49 pixels) against 1.27\,m, though on only 7\,\% of the pixels, and 0.93\,m at the Khalifa Port channel (583 pixels, 100\,\% coverage) against 0.69\,m. At Mirfa the lidar-only calibration (1.50\,m) was better than the sounding-based one (1.82\,m) but covered 2\,\% of the pixels, and at the Mugharraq basin and channel, where the lidar pixels lie almost entirely on the shallow margins, the lidar-only calibration was biased by $-1.1$ to $-2.9$\,m (RMSE 1.51--3.28\,m) and covered about a fifth of the pixels. At the Sir Bani Yas landing-craft basin the lidar failed the plausibility check and at the deep Khalifa Port basins no image could be calibrated with it.

\begin{table}[htbp]
\centering
\caption{ICESat-2 ATL24 checks. (a) Lidar against survey on pixels with soundings, before and after removal of the offset $+0.28$\,m $+3.5\,\%$; (b) map against the lidar of the independent check passes on pixels without soundings. $\Delta t$ = median absolute time between pass and survey.}
\label{tab:atl24}
\small
\begin{adjustbox}{max width=\textwidth}
\begin{tabular}{@{}lllrrrrr@{}}
\toprule
Site & Comparison & Pixels & $n$ & RMSE (m) & Bias (m) & Order 2 (\%) & $\Delta t$ (d) \\
\midrule
JD & (a) lidar vs survey, raw & with soundings & 61 & 0.50 & $+0.41$ & 96.7 & 524 \\
JD & (a) lidar vs survey, corrected & with soundings & 61 & 0.25 & $-0.01$ & 100.0 & 524 \\
KP\_BASIN & (a) lidar vs survey, corrected & with soundings & 10 & 0.37 & $+0.19$ & 100.0 & 339 \\
MIRFA & (a) lidar vs survey, corrected & with soundings & 19 & 0.61 & $-0.21$ & 84.2 & 490 \\
KP\_2021 & (a) lidar vs survey, corrected & with soundings & 30 & 0.94 & $+0.25$ & 86.7 & 730 \\
\midrule
JD & (b) map vs lidar & with soundings & 61 & 0.50 & $+0.04$ & 93.4 & 524 \\
JD & (b) map vs lidar & without soundings & 221 & 0.96 & $+0.74$ & 66.5 & 445 \\
MGP\_NC & (b) map vs lidar & without soundings & 699 & 2.12 & $+1.89$ & 14.3 & 215 \\
SBY\_LCB & (b) map vs lidar & without soundings & 20 & 1.82 & $-0.19$ & 35.0 & 73 \\
SBY\_LP & (b) map vs lidar & without soundings & 16 & 1.12 & $+0.98$ & 62.5 & 102 \\
MIRFA & (b) map vs lidar & without soundings & 58 & 3.65 & $+3.37$ & 5.2 & 727 \\
\bottomrule
\end{tabular}
\end{adjustbox}
\end{table}

\begin{table}[htbp]
\centering
\caption{Calibration with ICESat-2 alone (Hybrid-local), scored on survey pixels more than 225\,m from any lidar pixel, compared with the transfer method and with calibration on soundings at the same pixels. Coverage = share of those pixels given a depth by the lidar-only calibration.}
\label{tab:lidaronly}
\small
\begin{adjustbox}{max width=\textwidth}
\begin{tabular}{@{}lrlrrrrrrr@{}}
\toprule
Site & Lidar pixels & Lidar depths (m) & \multicolumn{2}{c}{Transfer (Stumpf)} & \multicolumn{2}{c}{Soundings (Hybrid-local)} & \multicolumn{3}{c}{ICESat-2 only (Hybrid-local)} \\
\cmidrule(lr){4-5}\cmidrule(lr){6-7}\cmidrule(lr){8-10}
 & & & RMSE & Bias & RMSE & Bias & RMSE & Bias & Coverage (\%) \\
\midrule
JD & 756 & $-0.5$ to 8.3 & 1.64 & $+0.65$ & 1.19 & $-0.10$ & 1.31 & $+0.11$ & 94.1 \\
MGP\_NB & 139 & $-0.3$ to 6.2 & 1.10 & $-0.11$ & 1.01 & $+0.04$ & 3.28 & $-2.87$ & 22.5 \\
MGP\_NC & 1,886 & $-0.5$ to 6.3 & 1.02 & $+0.12$ & 1.22 & $-0.71$ & 1.51 & $-1.10$ & 21.1 \\
SBY\_LP & 49 & $-0.1$ to 6.7 & 1.33 & $+0.84$ & 1.27 & $+0.05$ & 0.90 & $-0.31$ & 6.7 \\
MIRFA & 276 & $-0.5$ to 11.9 & 5.17 & $-4.67$ & 1.82 & $-1.08$ & 1.50 & $+0.15$ & 2.1 \\
KP\_CHANNEL & 583 & $-0.5$ to 12.0 & 3.09 & $-2.86$ & 0.69 & $-0.43$ & 0.93 & $-0.74$ & 100.0 \\
\bottomrule
\end{tabular}
\end{adjustbox}
\end{table}

\subsection{Very-high-resolution imagery}
\label{sec:vhr}

\begin{figure}[htbp]
\centering
\includegraphics[width=\textwidth]{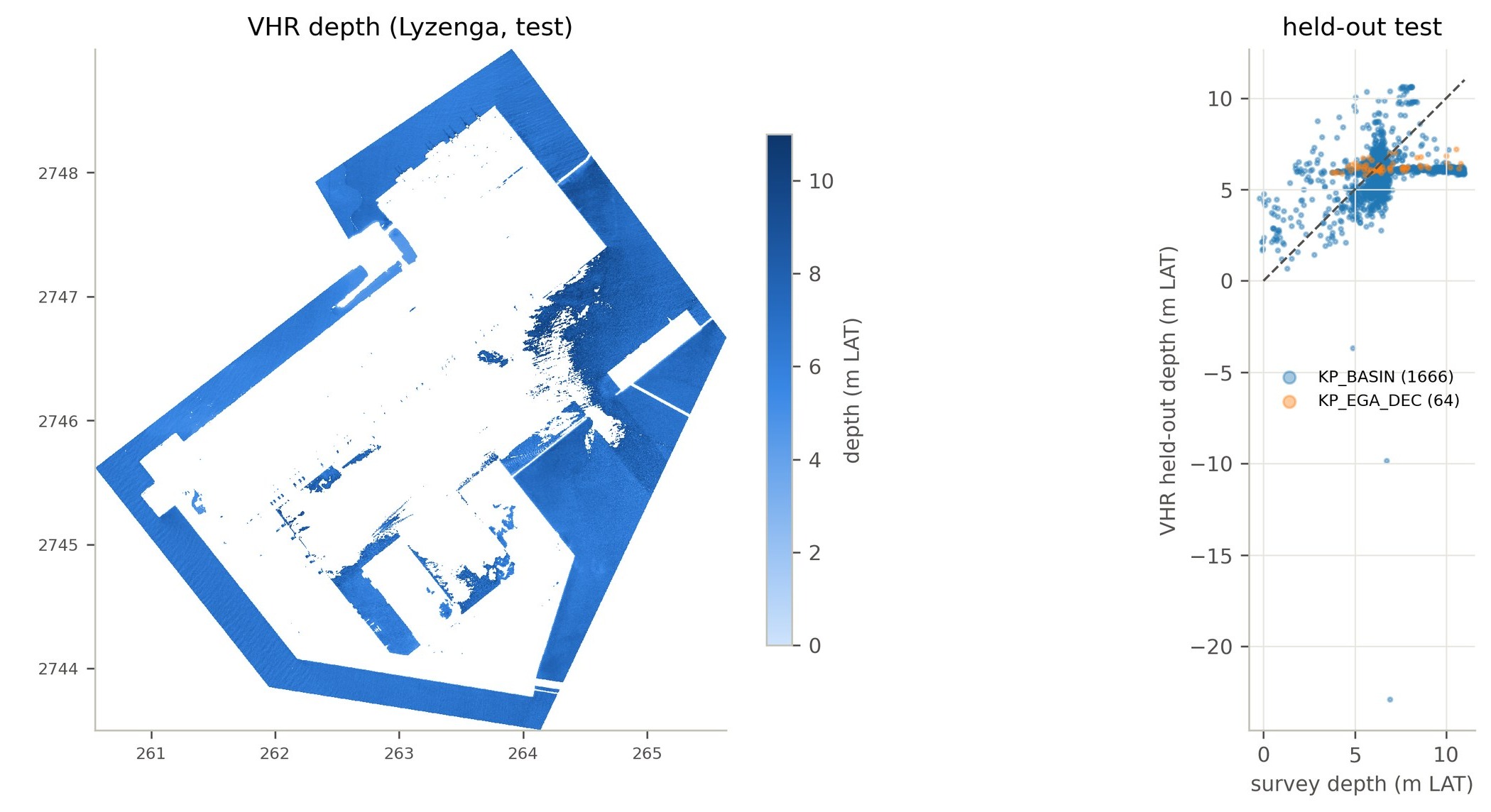}
\caption{Very-high-resolution test at Khalifa Port: the 2\,m depth map derived from a Pl\'eiades-1A image of 3 April 2026 (transfer) and the held-out comparison with the surveys of the basin and the EGA berths. Derived from Pl\'eiades data \copyright{}~CNES (2026), Distribution Airbus DS.}
\label{fig:vhr}
\end{figure}

The Pl\'eiades-1A image of Khalifa Port was tested with the three transfer models at 2\,m pixels on the survey pixels of the basin and the EGA berths that lie within the depth limit, 3.3--4.0 years after the surveys (Table~\ref{tab:vhr}; Fig.~\ref{fig:vhr}). At the basin, Lyzenga gave 2.19\,m RMSE (bias $-0.76$\,m, 50\,\% within Order~2) against 2.47\,m for LightGBM and 3.73\,m for Stumpf, whereas the Sentinel-2 local-control map at the same pixels gave 1.57\,m (bias $-0.75$\,m, 22\,\% within Order~1a/1b) on the 35\,\% of them that it covered. At the EGA berths the three models gave 1.64--1.69\,m on 64 pixels and the Sentinel-2 map covered none of them. Aggregating the Pl\'eiades image to 5 and 10\,m improved Lyzenga to 1.94\,m at 5\,m and Stumpf to 2.12\,m at 10\,m but did not bring any model below the Sentinel-2 value (Fig.~\ref{fig:resolution}c). Because the image is an 8-bit display product acquired four years after the surveys of a maintained basin, these results establish that a VHR image can be processed with the same chain and that its 2\,m depth map resolves the berths and the basin slopes, but they provide no evidence of an accuracy advantage of VHR over Sentinel-2 in these conditions (Section~\ref{sec:disc_vhr}).

\begin{table}[htbp]
\centering
\caption{Pl\'eiades-1A image of Khalifa Port (3 April 2026, 0.5\,m, 8-bit display product) at 2\,m pixels, tested with the transfer models on the survey pixels within the depth limit, next to the Sentinel-2 local-control map at the same pixels.}
\label{tab:vhr}
\small
\begin{adjustbox}{max width=\textwidth}
\begin{tabular}{@{}llrrrrrrr@{}}
\toprule
Site & Method & $n$ & RMSE (m) & Bias (m) & $R^2$ & Order 2 (\%) & Coverage (\%) & Years after survey \\
\midrule
KP\_BASIN & Stumpf & 1,674 & 3.73 & $-2.06$ & $-2.55$ & 28.6 & 53.2 & 4.0 \\
KP\_BASIN & Lyzenga & 1,666 & 2.19 & $-0.76$ & $-0.22$ & 50.1 & 53.0 & 4.0 \\
KP\_BASIN & LightGBM & 1,675 & 2.47 & $-1.38$ & $-0.56$ & 35.4 & 53.2 & 4.0 \\
KP\_BASIN & Sentinel-2, Hybrid-local & 824 & 1.57 & $-0.75$ & $-1.81$ & -- & 35.0 & -- \\
KP\_EGA\_DEC & Stumpf & 64 & 1.64 & $+0.06$ & 0.05 & 50.0 & 32.2 & 3.3 \\
KP\_EGA\_DEC & Lyzenga & 64 & 1.65 & $-0.46$ & 0.04 & 51.6 & 32.2 & 3.3 \\
KP\_EGA\_DEC & LightGBM & 64 & 1.69 & $+0.59$ & 0.00 & 45.3 & 32.2 & 3.3 \\
\bottomrule
\end{tabular}
\end{adjustbox}
\end{table}

\subsection{Seasonal variation of accuracy and coverage}
\label{sec:seasons}

\begin{figure}[htbp]
\centering
\includegraphics[width=\textwidth]{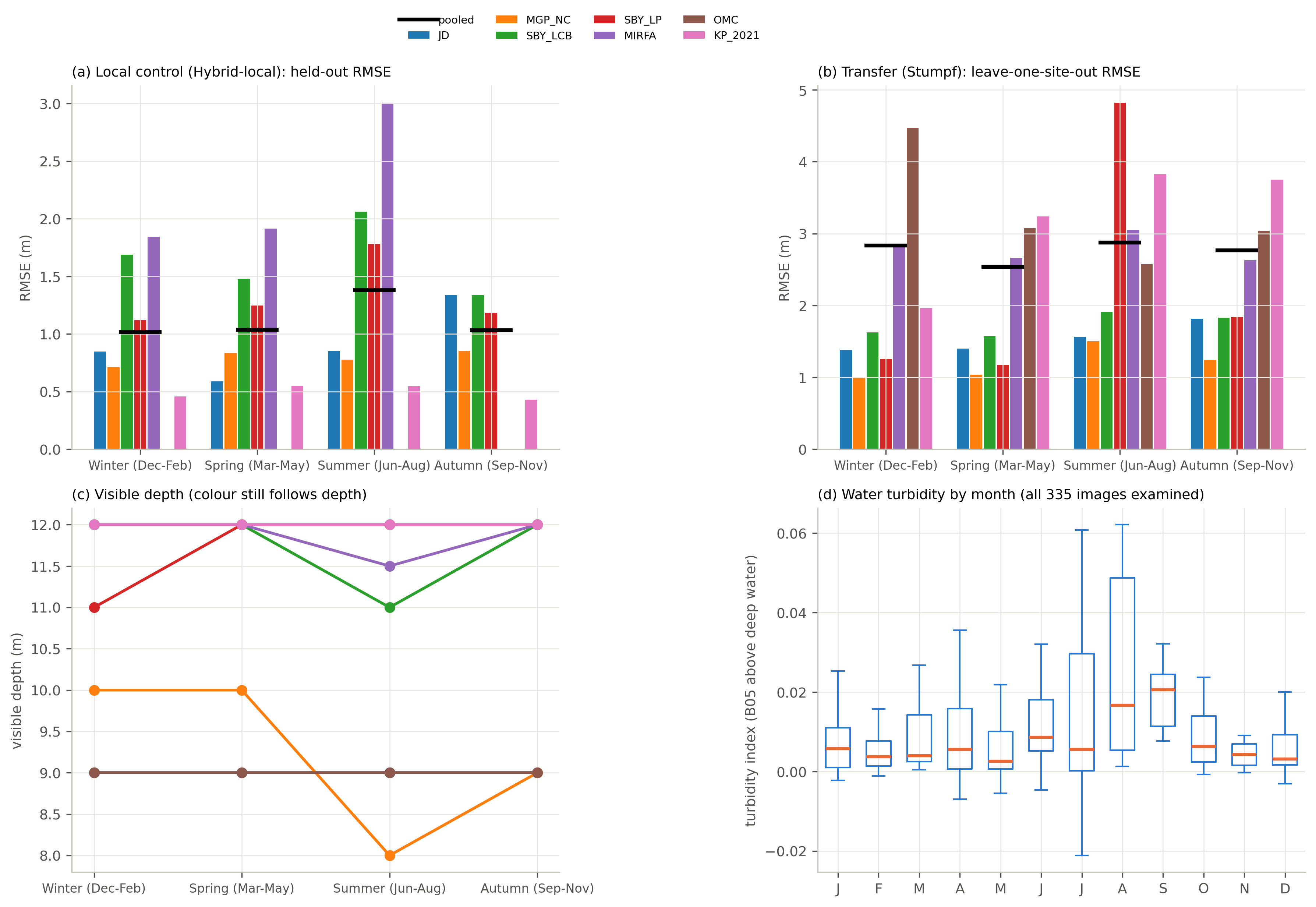}
\caption{Seasonal experiment: (a) local-control (Hybrid-local) and (b) transfer (Stumpf) RMSE per season and site, with the pooled value as a black bar; (c) visible depth per season; (d) turbidity index by month of all images examined.}
\label{fig:seasons}
\end{figure}

Table~\ref{tab:seasons} and Fig.~\ref{fig:seasons} summarise the seasonal experiment over the seven sites. The turbidity index of the 362 images examined in the main and seasonal experiments was lowest from November to March (monthly medians 0.002--0.006), rose in April--June and peaked from July to September (0.015--0.040), before returning to winter values in October (Fig.~\ref{fig:seasons}d). The pooled turbidity of the images kept per season was 0.003 in winter, 0.004 in spring, 0.017 in summer and 0.005 in autumn, i.e.\ five to six times higher in summer than in winter. The pooled local-control RMSE of the hybrid followed the turbidity: 1.02\,m in winter, 1.04\,m in spring, 1.03\,m in autumn and 1.38\,m in summer. The effect on coverage was larger than the effect on error: the median share of pixels that could be calibrated with the hybrid was 93--98\,\% in winter, spring and autumn but 70\,\% in summer, and with the Stumpf ratio 88--96\,\% against none at the median site in summer, when the ratio could be calibrated at three sites only. The transfer error (Stumpf 2.5--2.9\,m and LightGBM 2.4--3.0\,m pooled) was dominated by the biased sites and showed no clear seasonal pattern. The visible depth hardly changed with the season at the clear sites (12\,m at Sir Bani Yas, Mirfa and the Khalifa Port limits in nearly every season) but fell from 10 to 8\,m at the Mugharraq channel in summer.

Per site, the seasonal differences were large. At Jebel Dhanna the hybrid gave 0.59\,m in spring and 0.85\,m in winter and summer but 1.34\,m in autumn; at the Sir Bani Yas logistics port 1.12--1.25\,m in winter, spring and autumn but 1.78\,m on half of the pixels in summer (turbidity index 0.031), where the Stumpf transfer error reached 4.8\,m; at the landing-craft basin 1.34--1.48\,m in spring and autumn against 2.06\,m in summer; at Mirfa 1.85\,m in winter ($R^2=0.54$) and 1.92\,m in spring against 3.0\,m in summer and no calibrable image in autumn, so the winter images used for the main experiment at this site (Section~\ref{sec:independent}) were the right choice; and at the Khalifa Port limits 0.43--0.55\,m in every season, the clear offshore water being insensitive to the season. At Old Musaffah no image of any season could be calibrated, so that the failure of this site is not seasonal. The clear-water season of the Abu Dhabi coast, as seen by Sentinel-2, is therefore November--December to April--May, and the season term of the image selection (Section~\ref{sec:preproc}) is supported by the data.

\begin{table}[htbp]
\centering
\caption{Seasonal experiment pooled over the seven sites: turbidity index and visible depth of the images kept, pooled RMSE of the transfer and local-control methods and median coverage of the local methods. In summer the ratio could be calibrated at three sites only, and its pooled RMSE is given in brackets.}
\label{tab:seasons}
\small
\begin{adjustbox}{max width=\textwidth}
\begin{tabular}{@{}lrrrrrrrr@{}}
\toprule
Season & Turbidity index & Visible depth (m) & Stumpf RMSE (m) & LightGBM RMSE (m) & Stumpf-local RMSE (m) & Coverage (\%) & Hybrid-local RMSE (m) & Coverage (\%) \\
\midrule
Winter (Dec--Feb) & 0.003 & 11.0 & 2.84 & 2.74 & 1.21 & 95.7 & 1.02 & 98.4 \\
Spring (Mar--May) & 0.004 & 12.0 & 2.54 & 2.40 & 1.21 & 95.0 & 1.04 & 93.6 \\
Summer (Jun--Aug) & 0.017 & 11.0 & 2.88 & 2.78 & (1.00) & 0.0 & 1.38 & 69.9 \\
Autumn (Sep--Nov) & 0.005 & 12.0 & 2.77 & 3.00 & 1.15 & 87.6 & 1.03 & 98.1 \\
\bottomrule
\end{tabular}
\end{adjustbox}
\end{table}

\subsection{Number of fused images}
\label{sec:nimages}

\begin{figure}[htbp]
\centering
\includegraphics[width=0.9\textwidth]{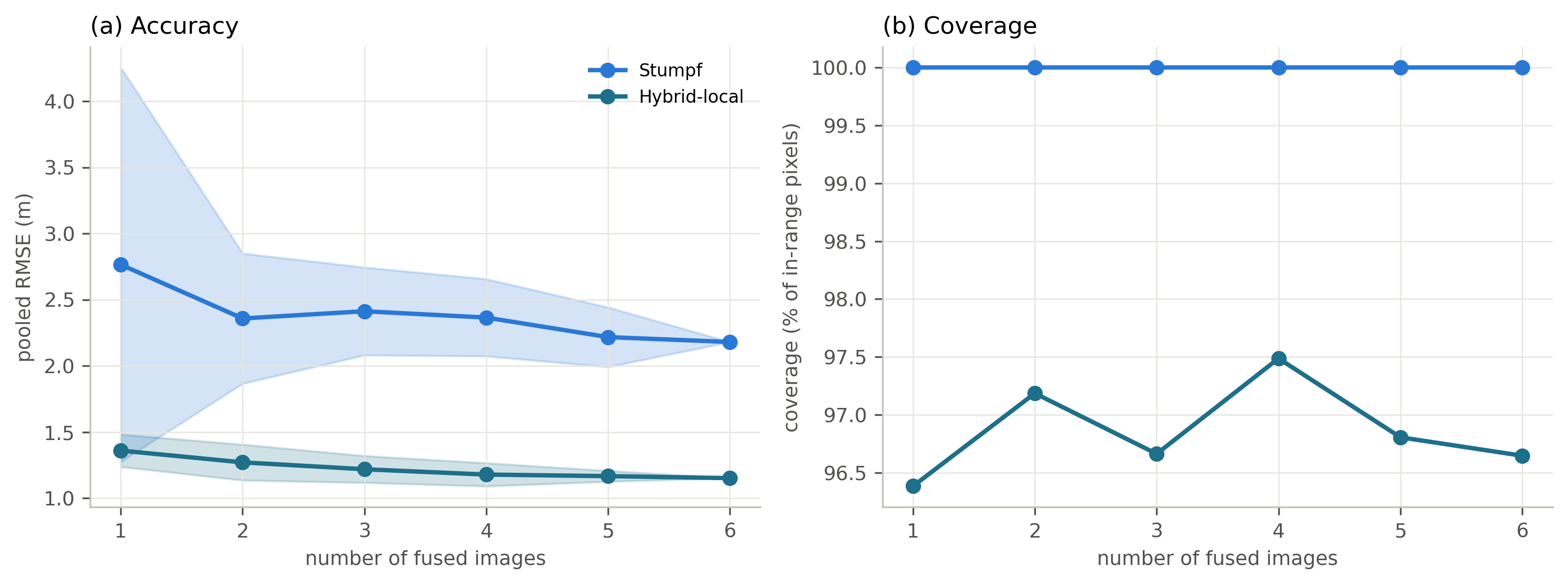}
\caption{Pooled held-out RMSE (a) and coverage (b) against the number of fused Sentinel-2 images (random subsets; shading: $\pm$1 SD over subsets).}
\label{fig:fused}
\end{figure}

Fusing more images improved both modes monotonically once the fusion weights accounted for the calibration quality of each image (Fig.~\ref{fig:fused}; Table~\ref{tab:experiments}). For the Stumpf transfer the pooled RMSE fell from 2.76\,m with one image (standard deviation over subsets 1.49\,m) to 2.41\,m with three and 2.18\,m with six, and the spread between subsets fell from 1.49 to 0.23\,m. For the hybrid the RMSE fell from 1.36\,m with one image to 1.22\,m with three, 1.17\,m with five and 1.15\,m with six, while the coverage remained between 96 and 97\,\%. Most of the gain was obtained with three images (two thirds of the total improvement for the hybrid), and the fourth to sixth images, which are those with the lowest scene scores, still contributed because their weight was reduced in proportion to their calibration residual.

\subsection{Number of control soundings}
\label{sec:nsoundings}

\begin{figure}[htbp]
\centering
\includegraphics[width=0.9\textwidth]{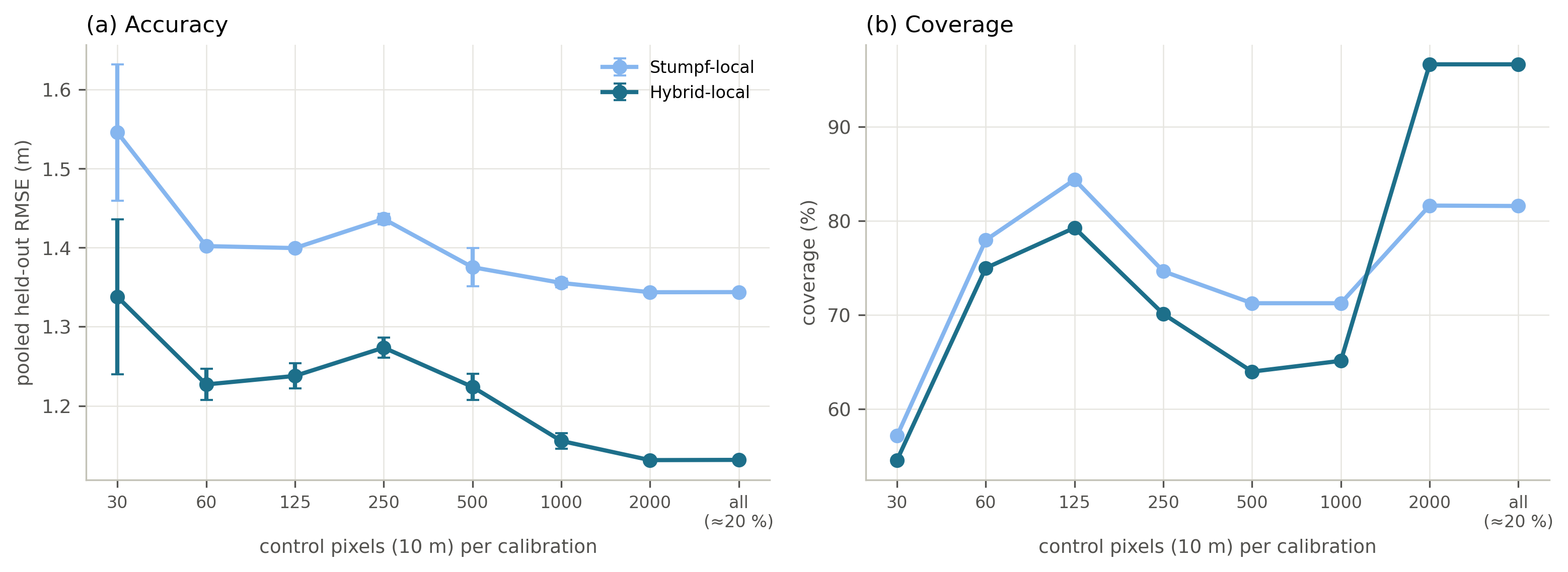}
\caption{Pooled held-out RMSE (a) and coverage (b) of local calibration against the number of control pixels per image (error bars: $\pm$1 SD over seeds).}
\label{fig:soundings}
\end{figure}

The local-control accuracy saturated at a few tens to a few hundred control pixels per image (Fig.~\ref{fig:soundings}; Table~\ref{tab:experiments}). With 30 control pixels (10\,m) per image the pooled held-out RMSE was 1.55\,m for Stumpf-local and 1.34\,m for the hybrid, and only 55--57\,\% of the pixels could be calibrated, because many images failed the fit criteria with so few controls. With 60 pixels the RMSE reached 1.40\,m for Stumpf-local and 1.23\,m for the hybrid, with 75--78\,\% coverage, and from 125 to 1,000 pixels it remained between 1.36 and 1.44\,m for Stumpf-local and between 1.16 and 1.27\,m for the hybrid; with 2,000 pixels and with all controls (about 20\,\% of each site, up to 5,000 pixels) the hybrid reached 1.13\,m with 97\,\% coverage. The hybrid, with four coefficients and depth-balanced weights, gained more from large samples than the two-parameter ratio, whose error with all controls (1.34\,m) was within 0.06\,m of its error with 60 pixels. About 60--125 well-distributed control pixels per image, equivalent to a single-beam line of about 1\,km across the depth range, therefore gave most of the accuracy of the method, and more controls improved coverage and, for the hybrid, a further decimetre of accuracy.

\subsection{Pixel size}
\label{sec:resolution}

\begin{figure}[htbp]
\centering
\includegraphics[width=\textwidth]{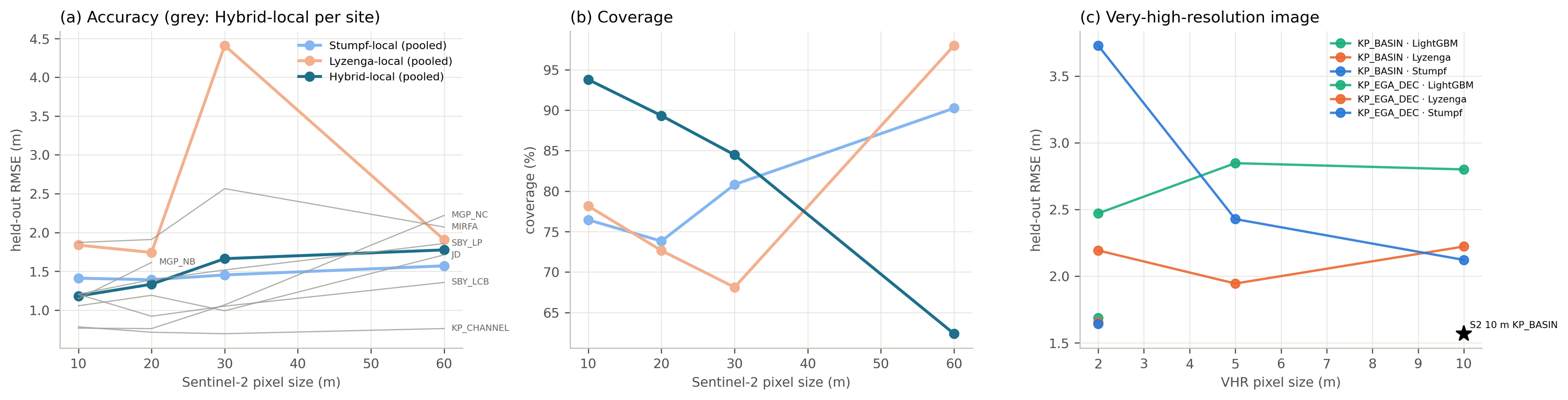}
\caption{Effect of pixel size: (a) pooled held-out RMSE of the local-control methods for Sentinel-2 aggregated from 10 to 60\,m (grey: Hybrid-local per site); (b) coverage; (c) the Pl\'eiades image at 2, 5 and 10\,m, with the Sentinel-2 result at the Khalifa Port basin for reference.}
\label{fig:resolution}
\end{figure}

Aggregating the Sentinel-2 features from 10 to 20\,m left the ratio method unchanged (1.41 to 1.39\,m for Stumpf-local) and cost the hybrid 0.15\,m (1.18 to 1.33\,m), because the hybrid exploits the $7\times7$ context features that are coarsened by the aggregation (Fig.~\ref{fig:resolution}a; Table~\ref{tab:experiments}). At 30 and 60\,m the RMSE rose to 1.66 and 1.78\,m for the hybrid and to 1.45 and 1.57\,m for Stumpf-local, and the coverage fell because the coarser pixels straddle the edges of the surveyed areas (Fig.~\ref{fig:resolution}b). The Lyzenga-local method was erratic beyond 20\,m (4.41\,m at 30\,m). For the Pl\'eiades image (Fig.~\ref{fig:resolution}c) the ratio and Lyzenga models improved from 2 to 5--10\,m (3.73 to 2.12\,m and 2.19 to 1.94\,m) whereas LightGBM did not, so the pixel noise of the 8-bit product, rather than the seabed, limited the 2\,m result.

\begin{table}[htbp]
\centering
\caption{Sensitivity experiments at the training sites: pooled held-out RMSE (m) and coverage (\%) against the number of fused images (mean and SD over random subsets), the number of control pixels per image (mean and SD over seeds) and the Sentinel-2 pixel size.}
\label{tab:experiments}
\small
\begin{adjustbox}{max width=\textwidth}
\begin{tabular}{@{}lrrrrrr@{}}
\toprule
\multicolumn{7}{@{}l}{\textit{Number of fused images}} \\
Images & 1 & 2 & 3 & 4 & 5 & 6 \\
\midrule
Stumpf (transfer) & 2.76 $\pm$ 1.49 & 2.36 $\pm$ 0.49 & 2.41 $\pm$ 0.33 & 2.36 $\pm$ 0.29 & 2.22 $\pm$ 0.23 & 2.18 \\
Hybrid-local & 1.36 $\pm$ 0.12 & 1.27 $\pm$ 0.14 & 1.22 $\pm$ 0.10 & 1.18 $\pm$ 0.09 & 1.17 $\pm$ 0.04 & 1.15 \\
Hybrid-local coverage & 96.4 & 97.2 & 96.7 & 97.5 & 96.8 & 96.6 \\
\midrule
\multicolumn{7}{@{}l}{\textit{Control pixels per image}} \\
Pixels & 30 & 60 & 125 & 250 & 500 & 1,000 / 2,000 / all \\
\midrule
Stumpf-local & 1.55 $\pm$ 0.09 & 1.40 $\pm$ 0.00 & 1.40 $\pm$ 0.00 & 1.44 $\pm$ 0.01 & 1.38 $\pm$ 0.02 & 1.36 / 1.34 / 1.34 \\
Stumpf-local coverage & 57.1 & 78.0 & 84.4 & 74.6 & 71.2 & 71.2 / 81.6 / 81.6 \\
Hybrid-local & 1.34 $\pm$ 0.10 & 1.23 $\pm$ 0.02 & 1.24 $\pm$ 0.02 & 1.27 $\pm$ 0.01 & 1.22 $\pm$ 0.02 & 1.16 / 1.13 / 1.13 \\
Hybrid-local coverage & 54.5 & 75.0 & 79.3 & 70.1 & 64.0 & 65.1 / 96.7 / 96.7 \\
\midrule
\multicolumn{7}{@{}l}{\textit{Sentinel-2 pixel size (m)}} \\
Pixel size & 10 & 20 & 30 & 60 & & \\
\midrule
Stumpf-local & 1.41 & 1.39 & 1.45 & 1.57 & & \\
Lyzenga-local & 1.84 & 1.74 & 4.41 & 1.91 & & \\
Hybrid-local & 1.18 & 1.33 & 1.66 & 1.78 & & \\
\bottomrule
\end{tabular}
\end{adjustbox}
\end{table}

\subsection{Feature contributions}
\label{sec:shap}

\begin{figure}[htbp]
\centering
\includegraphics[width=\textwidth]{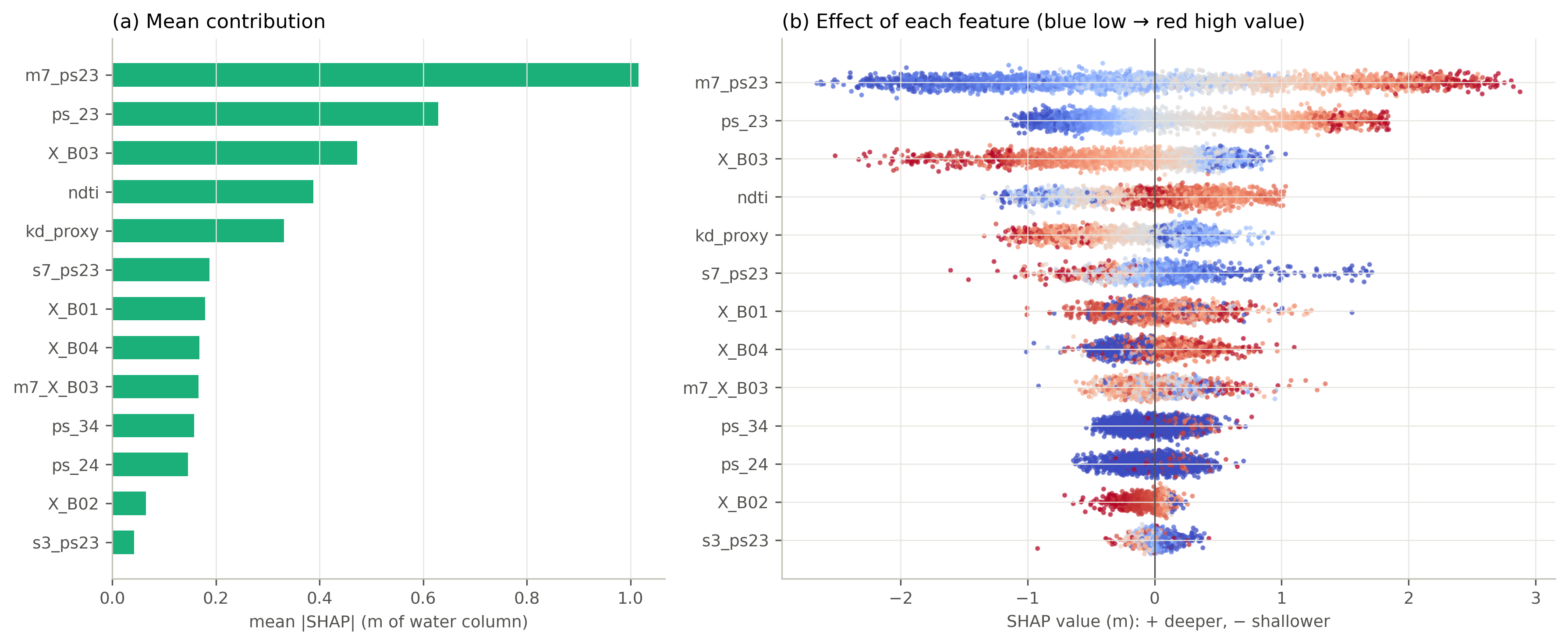}
\caption{SHAP feature contributions of the LightGBM model: (a) mean absolute contribution in metres of water column; (b) distribution of the contribution of each feature, coloured by the feature value (blue low, red high); positive values deepen the prediction.}
\label{fig:shap}
\end{figure}

The SHAP analysis of the LightGBM model (Fig.~\ref{fig:shap}) attributed the largest mean absolute contribution, 1.02\,m of water column, to the $7\times7$ mean of the blue/green log-ratio $m_7(p_{23})$, followed by the pixel-level ratio $p_{23}$ (0.63\,m), the green Lyzenga transform $X_{B03}$ (0.47\,m), the turbidity index NDTI (0.39\,m) and the $K_d$ proxy (0.33\,m). The directions of the effects are physical: a high blue/green ratio and a low green transform push the prediction deeper, a high turbidity index (bright shallow bottom or sediment) pushes it shallower, and the spread of the $7\times7$ ratio, $s_7(p_{23})$ (0.19\,m), which is large on slopes and edges, pushes it shallower. The coastal-aerosol and red transforms (0.17--0.18\,m), the smoothed green transform and the remaining ratios (0.15--0.17\,m) contributed less, and the blue transform and the $3\times3$ spread less than 0.1\,m each. The model therefore relied first on the smoothed and the pixel-level Stumpf ratio, i.e.\ on the same signal as the classical method, and used the water-column descriptors to modulate it for turbidity, which is consistent with the small difference between the two methods in transfer mode and the larger advantage of the tree model within surveyed areas.

\subsection{Outlook for the waters of the UAE}
\label{sec:outlook}

\subsubsection{Coast-wide feasibility from one year of Sentinel-2}
\label{sec:coast}

\begin{figure}[htbp]
\centering
\includegraphics[width=\textwidth]{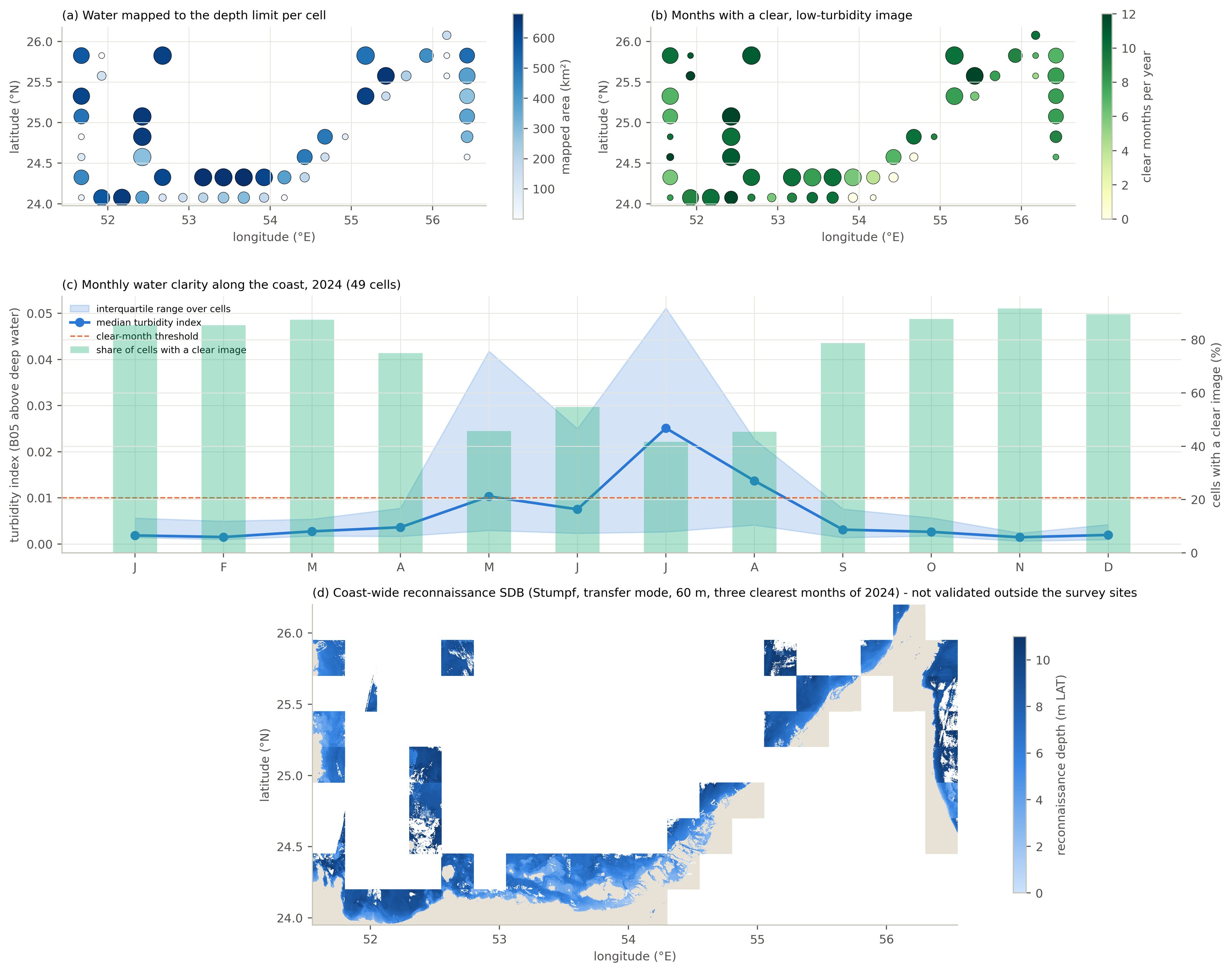}
\caption{Coast-wide feasibility of SDB along the UAE coast from the Sentinel-2 archive of 2024: (a) water given a reconnaissance depth to the 11\,m limit per $0.25\degree$ cell; (b) number of months with a clear, low-turbidity image; (c) monthly turbidity index over the cells (median and interquartile range) and share of cells with a clear image; (d) reconnaissance depth mosaic in transfer mode at 60\,m, checked against ICESat-2 in Section~\ref{sec:coastatl24} and including deep water that the ratio saturates at 8--10\,m (Section~\ref{sec:coastatl24}).}
\label{fig:coast}
\end{figure}

Of the 200 cells of the $0.25\degree$ grid, 60 passed the water-fraction probe and 49 contained at least one clear month in 2024; the eleven others were inland cells of the Hajar foothills in which mountain shadow had been classed as water. The 49 coastal cells hold 19,263\,km$^2$ of water: 8,049\,km$^2$ in 19 cells of the Al Dhafra coast west of 53\degree E, 5,543\,km$^2$ in 14 cells of the Abu Dhabi coast (53--54.7\degree E), 2,958\,km$^2$ in 7 cells from Dubai to Ras Al Khaimah (54.7--56\degree E) and 2,713\,km$^2$ in 9 cells of the Gulf of Oman coast and the mountains east of 56.1\degree E (Table~\ref{tab:coast}; Fig.~\ref{fig:coast}). The number of clear months per cell, i.e.\ months in which the least cloudy image was at least 70\,\% cloud-free and had a turbidity index at or below 0.010, had a median of 9 and ranged from 0 to 12 (Fig.~\ref{fig:coast}b). The Al Dhafra coast was the clearest (median 10, 6--12 months); the Dubai--Ras Al Khaimah coast had 9 (6--12) and the east coast 8 (5--10); and the Abu Dhabi coast had 7.5 (0--10), with four cells, those of the urban coast of Abu Dhabi island, the Musaffah channel and Khalifa Port (53.9--54.7\degree E), in which no month of 2024 was clear (median turbidity index 0.035--0.15). The monthly climatology over the 49 cells (Fig.~\ref{fig:coast}c; Table~\ref{tab:coastclim}) is the coast-wide form of the seasonal result of Section~\ref{sec:seasons}: the median turbidity index was 0.001--0.004 from September to April, rose to 0.010 in May and 0.025 in July, and the share of cells with a clear image was 75--92\,\% from September to April against 42--55\,\% from May to August. Cloud was never the limiting factor (median cloud fraction below 6\,\% in every month); turbidity was.

Applying the transfer method to the three clearest months of each cell gave a reconnaissance depth for 17,233\,km$^2$, 89\,\% of the water examined, with 1,002\,km$^2$ beyond the 11\,m limit, mainly offshore of Al Dhafra and on the east coast. The two tests of optically deep water classed almost the same area as shallow, 18,861\,km$^2$ by the signal-to-noise test and 18,571\,km$^2$ by the spread test, and their check at the surveyed sites (Table~\ref{tab:deeptest}) explains why: of the survey pixels deeper than the 11\,m limit, the signal-to-noise test flagged 2.3\,\% and the spread test 5.9\,\%, while they wrongly flagged 1.4 and 3.2\,\% of the pixels within the limit. Deep water in the Gulf, and the water of the turbid basins, is brighter than the darkest 2\,\% of a context window, so that no test on the brightness of a single image can separate it from water of 8--10\,m, and the Stumpf ratio, which saturates, gives it a depth of 8--10\,m; within the surveyed sites this is caught by the seabed-not-readable test, which needs control depths that do not exist offshore. The mapped area of this experiment is therefore an upper bound that includes an unknown share of deep water, and the median reconnaissance depth per cell of 2.6--10.5\,m (median 6.8\,m) and the median site-conformal U95 of 1.7--10.6\,m (median 4.4\,m) describe the product only where the seabed is within reach. That share was measured against ICESat-2 (Section~\ref{sec:coastatl24}).

\begin{table}[htbp]
\centering
\caption{Check of the two tests of optically deep water at the surveyed sites (pixel $\times$ image rows of the main experiment): share of survey pixels flagged as optically deep by the signal-to-noise test of Section~\ref{sec:preproc} and by the spread test of the coast product ($k=2$), for pixels deeper and shallower than the 11\,m depth limit.}
\label{tab:deeptest}
\small
\begin{tabular}{@{}lrrr@{}}
\toprule
Survey pixels & $n$ & Flagged, SNR test (\%) & Flagged, $\sigma_\infty$ test (\%) \\
\midrule
Deeper than the limit (should be flagged) & 886,278 & 2.3 & 5.9 \\
Within the limit (should not be flagged) & 459,372 & 1.4 & 3.2 \\
\bottomrule
\end{tabular}
\end{table}

\begin{table}[htbp]
\centering
\caption{Coast-wide feasibility of SDB in 2024, summarised by coastal region from the 49 cells of $0.25\degree$ (the full table per cell is given in the supplementary material). Water = area of water examined; shallow = area classed optically shallow by the signal-to-noise (SNR) and by the spread ($\sigma_\infty$) test; mapped = area given a reconnaissance depth to the 11\,m limit; clear months = median (range) over the cells; turbidity, depth and U95 = medians over the cells.}
\label{tab:coast}
\small
\begin{adjustbox}{max width=\textwidth}
\begin{tabular}{@{}lrrrrrrrrr@{}}
\toprule
Region & Cells & Water (km$^2$) & Shallow, SNR (km$^2$) & Shallow, $\sigma_\infty$ (km$^2$) & Mapped (km$^2$) & Clear months & Turbidity index & Depth (m) & U95 (m) \\
\midrule
Al Dhafra (51.5--53.0\degree E) & 19 & 8,049 & 7,983 & 7,670 & 7,146 & 10 (6--12) & 0.002 & 8.0 & 4.8 \\
Abu Dhabi (53.0--54.7\degree E) & 14 & 5,543 & 5,523 & 5,283 & 5,280 & 7.5 (0--10) & 0.005 & 4.7 & 3.7 \\
Dubai to Ras Al Khaimah (54.7--56.0\degree E) & 7 & 2,958 & 2,920 & 2,917 & 2,704 & 9 (6--12) & 0.003 & 6.3 & 4.4 \\
East coast and mountains ($>$56.1\degree E) & 9 & 2,713 & 2,436 & 2,701 & 2,102 & 8 (5--10) & 0.003 & 7.4 & 4.9 \\
\midrule
All coastal cells & 49 & 19,263 & 18,861 & 18,571 & 17,233 & 9 (0--12) & 0.003 & 6.8 & 4.4 \\
\bottomrule
\end{tabular}
\end{adjustbox}
\end{table}

\begin{table}[htbp]
\centering
\caption{Monthly water-clarity climatology of the 49 coastal cells in 2024: median and interquartile range of the turbidity index of the least cloudy image of the month, share of cells with a clear image (cloud-free $\geq$70\,\%, turbidity $\leq$0.010) and median cloud fraction.}
\label{tab:coastclim}
\small
\begin{tabular}{@{}lrrrrr@{}}
\toprule
Month & Cells & Turbidity (median) & $Q_{25}$--$Q_{75}$ & Clear cells (\%) & Cloud (\%) \\
\midrule
January & 48 & 0.002 & 0.001--0.006 & 85 & 3.0 \\
February & 48 & 0.002 & 0.001--0.005 & 85 & 4.5 \\
March & 48 & 0.003 & 0.002--0.005 & 88 & 5.4 \\
April & 44 & 0.004 & 0.002--0.008 & 75 & 2.0 \\
May & 35 & 0.010 & 0.003--0.042 & 46 & 2.7 \\
June & 42 & 0.008 & 0.002--0.025 & 55 & 3.1 \\
July & 36 & 0.025 & 0.003--0.051 & 42 & 4.9 \\
August & 44 & 0.014 & 0.004--0.023 & 45 & 2.8 \\
September & 47 & 0.003 & 0.001--0.008 & 79 & 1.7 \\
October & 49 & 0.003 & 0.002--0.006 & 88 & 3.2 \\
November & 48 & 0.001 & 0.001--0.002 & 92 & 4.2 \\
December & 48 & 0.002 & 0.001--0.004 & 90 & 2.6 \\
\bottomrule
\end{tabular}
\end{table}

\subsubsection{ICESat-2 check of the coast product}
\label{sec:coastatl24}

\begin{figure}[htbp]
\centering
\includegraphics[width=\textwidth]{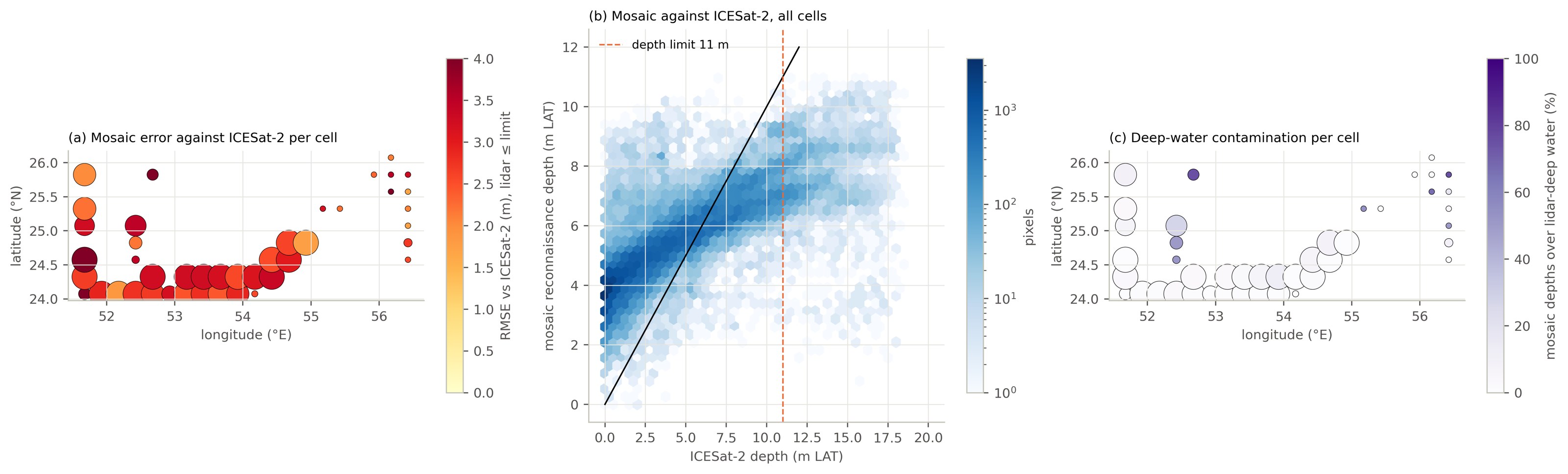}
\caption{ICESat-2 ATL24 check of the coast-wide reconnaissance product: (a) RMSE of the mosaic against the offset-corrected lidar on lidar-shallow water per cell (symbol size: number of lidar pixels); (b) mosaic depth against lidar depth at all 110,039 lidar pixels (line 1:1; dashed: 11\,m depth limit); (c) share of the mosaic depths that lie over water the lidar finds deeper than the limit, per cell.}
\label{fig:coastatl24}
\end{figure}

The 291 ATL24 passes obtained over the coastal cells gave 110,039 lidar pixels of 60\,m in 47 cells, 36 of which had at least 50 pixels (Table~\ref{tab:coastatl24}; Fig.~\ref{fig:coastatl24}). They cover the Al Dhafra and Abu Dhabi coasts densely (33 of the 36 cells; up to 18,300 pixels per cell) but only three cells east of 54.7\degree E with a few passes each, so that the check describes mainly the western two thirds of the coast. On the lidar pixels shallower than the 11\,m limit ($n=97,271$), the reconnaissance mosaic had an RMSE of 2.91\,m and a bias of $+2.02$\,m, and 72\,\% of the lidar depths lay inside the site-conformal U95. The error depended strongly on depth: 3.4\,m (bias $+3.3$\,m) in the 0--2\,m band, which holds a third of the lidar pixels, 2.1\,m in the 2--5\,m band, 0.91\,m (bias $-0.22$\,m) in the 5--8\,m band and 2.5\,m (bias $-2.3$\,m) in the 8--11\,m band. In 1\,m bins the error fell from 3.6\,m at 0--1\,m to a minimum of 0.73\,m at 6--7\,m and rose again to 3.3\,m at 10--11\,m. This curve reproduces, almost bin for bin, the error of the same Stumpf transfer against the multibeam surveys at the training sites (Table~\ref{tab:bands} and Fig.~\ref{fig:depthlimits}a: 4.15\,m at 0--1\,m, 0.68\,m at 6--7\,m, 2.33\,m at 10--11\,m), although the lidar is an entirely independent reference, measured over 36 cells of open coast rather than six port areas, and reduced to LAT through its own tide and offset. The compression of the depth range by the transfer ratio, which over-estimates the flats and under-estimates the deeper water, is therefore a property of the method on this coast and not of the survey sites.

The deep-water contamination was small along the coast and large offshore. Over all lidar pixels that received a mosaic depth, 4.6\,\% lay where the lidar found the seabed deeper than 11\,m, and only 18\,\% of the lidar-deep pixels had been left without a depth by the mosaic, consistent with the weakness of both deep-water tests (Table~\ref{tab:deeptest}). In the 29 cells whose lidar was mostly shallow the contamination was 4.1\,\%; in the seven cells where the lidar found more than 30\,\% of the water deeper than 11\,m, mostly offshore cells of Al Dhafra and the Dubai coast, it was 59\,\%, with the mosaic giving a median depth of 8.6\,m to water that the lidar measured beyond 11\,m (Fig.~\ref{fig:coastatl24}b, right of the dashed line). Removing the contamination share cell by cell reduced the mapped area of the 36 checked cells from 13,462 to 11,428\,km$^2$ (85\,\%); in five offshore cells (c043, c063, c114, c121 and c144) the lidar-verified area was 50--100\,\% smaller than the mapped area. Applied to the coast as a whole, the check indicates that about 15\,\% of the area given a reconnaissance depth in transfer mode lies over water beyond the depth limit, concentrated offshore, and that the shallow coastal water that dominates the area is correctly placed within the limit but with a depth error of 2--3.5\,m on the flats.

\begin{table}[htbp]
\centering
\caption{ICESat-2 check of the coast-wide reconnaissance mosaic by lidar depth band (110,039 lidar pixels of 60\,m in 47 cells; 291 passes, October 2022 to December 2025): lidar pixels, share given a mosaic depth, and RMSE and bias of the mosaic against the offset-corrected lidar (positive: mosaic too deep).}
\label{tab:coastatl24}
\small
\begin{tabular}{@{}lrrrr@{}}
\toprule
Lidar depth (m) & Lidar pixels & Given a depth (\%) & RMSE (m) & Bias (m) \\
\midrule
0--2 & 33,062 & 91.6 & 3.43 & $+3.32$ \\
2--5 & 24,813 & 94.2 & 2.08 & $+1.84$ \\
5--8 & 15,530 & 93.6 & 0.91 & $-0.22$ \\
8--11 & 9,253 & 95.3 & 2.52 & $-2.31$ \\
11--15 & 4,381 & 87.1 & 5.03 & $-4.80$ \\
15--20 & 1,286 & 64.7 & 8.80 & $-8.56$ \\
\midrule
All $\leq$11 m & 97,271 (compared) & -- & 2.91 & $+2.02$ \\
\bottomrule
\end{tabular}
\end{table}

\subsubsection{Coast product calibrated with ICESat-2}
\label{sec:coastcal}

Because the lidar reproduced the survey-site error curve, it was used to calibrate the mosaic cell by cell (Section~\ref{sec:outlookmethods}). Of the 49 coastal cells, 24 had enough lidar and a calibration that passed the quality gate; 18 had fewer than 50 lidar pixels, almost all of them east of 54.7\degree E or far offshore, and in seven the lidar depth did not follow the mosaic depth (four fits with $R^2<0.3$ and three with a negative slope, all in cells where the lidar is mostly deeper than 11\,m or the water is turbid) and the transfer depth was kept (supplementary Table~\ref{tab:coastcalcells}). The 24 calibrated cells hold 8,584\,km$^2$ of mapped water, half of the coast product, including the whole Al Dhafra coast and the coast of Abu Dhabi from Mirfa to Ras Ghanada. In nine of them the lidar set the depth limit below 11\,m (3--10\,m), which removes the saturated depths that the transfer mosaic had assigned to deeper water.

\begin{figure}[htbp]
\centering
\includegraphics[width=\textwidth]{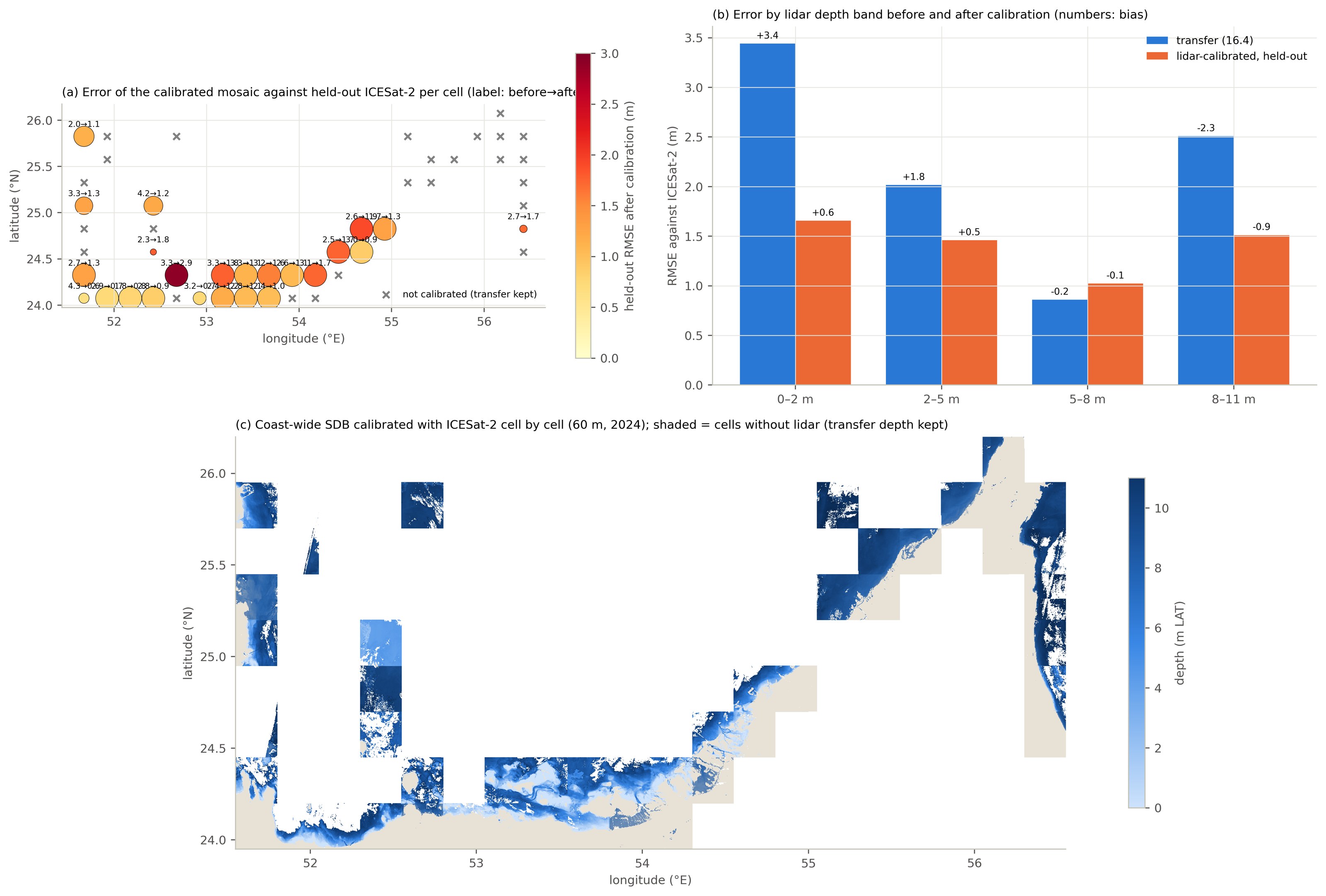}
\caption{Coast product calibrated with ICESat-2 cell by cell: (a) held-out RMSE against the lidar per calibrated cell after calibration (labels: before $\to$ after; crosses: cells that keep the transfer depth); (b) error by lidar depth band before and after calibration (numbers: bias); (c) the calibrated mosaic at 60\,m (shaded: cells without enough lidar, transfer depth kept).}
\label{fig:coastcal}
\end{figure}

On the held-out blocks of the 24 cells ($n=87,982$ lidar pixels within the limit) the error of the mosaic fell from 2.89\,m to 1.42\,m and its bias from $+2.01$ to $+0.24$\,m (Table~\ref{tab:coastcal}; Fig.~\ref{fig:coastcal}). The gain is largest where the transfer error was a bias: on the flats (0--2\,m) the RMSE fell from 3.44 to 1.66\,m and the bias from $+3.4$ to $+0.6$\,m, in the 2--5\,m band from 2.02 to 1.46\,m, and in the 8--11\,m band from 2.51 to 1.51\,m with the bias reduced from $-2.3$ to $-0.9$\,m. In the 5--8\,m band, where the transfer mosaic was already nearly unbiased, the calibration cost 0.16\,m (0.86 to 1.02\,m), because a single straight line with a shallow hinge cannot fit the flats and the mid-depths at the same time. By 1\,m bin the calibrated error lay between 0.96 and 1.8\,m from 0 to 10\,m, against 0.7--3.6\,m before, and the share of pixels within the S-44 Order~2 tolerance rose from 17 to 69\,\% (from 0 to 65\,\% on the flats), with 40\,\% within Order~1a/1b. Per cell the held-out error after calibration had a median of 1.21\,m (0.63--2.90\,m); six cells were below 1\,m, among them the Al Dhafra cells c001--c003 (0.74--0.86\,m) and c052 east of Mirfa (0.87\,m), and the densest cell, c027 off the Mirfa--Abu Al Abyad coast with 18,000 lidar pixels, reached 1.12\,m. The conformal interval of the calibrated product, taken as the largest quantile over the cells, covered 99\,\% of the held-out lidar depths at the nominal 95\,\%, with a median half-width of 4.4\,m; it is conservative, and a per-cell quantile would halve it where the lidar is dense.

The calibrated product is thus a map of about half the coast with the accuracy of the locally controlled port maps (1.1--1.5\,m), unbiased on the flats, with a depth limit set by the lidar in each cell, obtained without any vessel survey; the other half keeps the transfer depth and its error budget. Its remaining limits are those of the lidar coverage: the cells east of 54.7\degree E had too few passes within the download budget, and the error at 8--11\,m, although halved, is still the largest, because the ratio saturates there and the lidar can only re-scale what the image still resolves.

\begin{table}[htbp]
\centering
\caption{Coast product before and after calibration on ICESat-2, by lidar depth band, on held-out blocks of the 24 calibrated cells (87,982 lidar pixels within the depth limit): RMSE and bias (positive: too deep), share of pixels within the S-44 Order~2 tolerance before and after, share within Order~1a/1b after, and observed coverage of the conformal 95\,\% interval of the calibrated product.}
\label{tab:coastcal}
\small
\begin{adjustbox}{max width=\textwidth}
\begin{tabular}{@{}lrrrrrrrrr@{}}
\toprule
 & & \multicolumn{2}{c}{Transfer} & \multicolumn{2}{c}{Lidar-calibrated} & \multicolumn{2}{c}{Order 2 (\%)} & Order 1a/1b (\%) & U95 (\%) \\
Lidar depth (m) & $n$ & RMSE (m) & Bias (m) & RMSE (m) & Bias (m) & before & after & after & after \\
\midrule
0--2 & 26,830 & 3.44 & $+3.35$ & 1.66 & $+0.55$ & 0.2 & 64.7 & 37.1 & 98.4 \\
2--5 & 21,419 & 2.02 & $+1.82$ & 1.46 & $+0.47$ & 16.3 & 67.8 & 39.6 & 99.4 \\
5--8 & 13,829 & 0.86 & $-0.21$ & 1.02 & $-0.06$ & 76.9 & 79.1 & 50.3 & 99.3 \\
8--11 & 7,938 & 2.51 & $-2.31$ & 1.51 & $-0.93$ & 7.9 & 58.2 & 30.8 & 99.2 \\
\midrule
All $\leq$11 m & 87,982 & 2.89 & $+2.01$ & 1.42 & $+0.24$ & 16.9 & 68.7 & 39.6 & 99.0 \\
\bottomrule
\end{tabular}
\end{adjustbox}
\end{table}

\subsubsection{Unsurveyed areas mapped with ICESat-2 alone}
\label{sec:unsurveyed}

\begin{figure}[htbp]
\centering
\includegraphics[width=\textwidth]{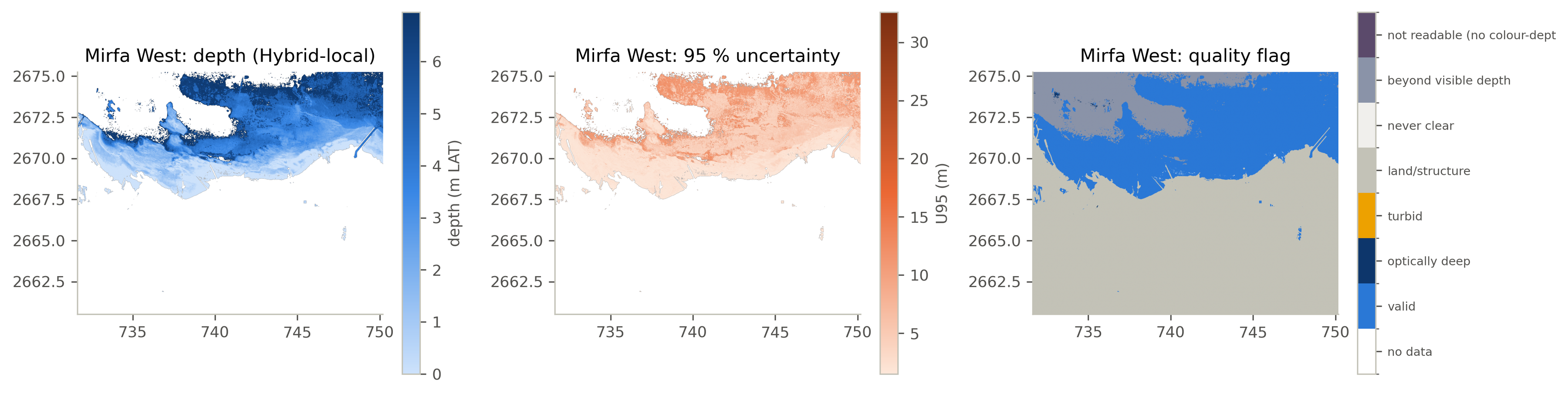}\\[1mm]
\includegraphics[width=\textwidth]{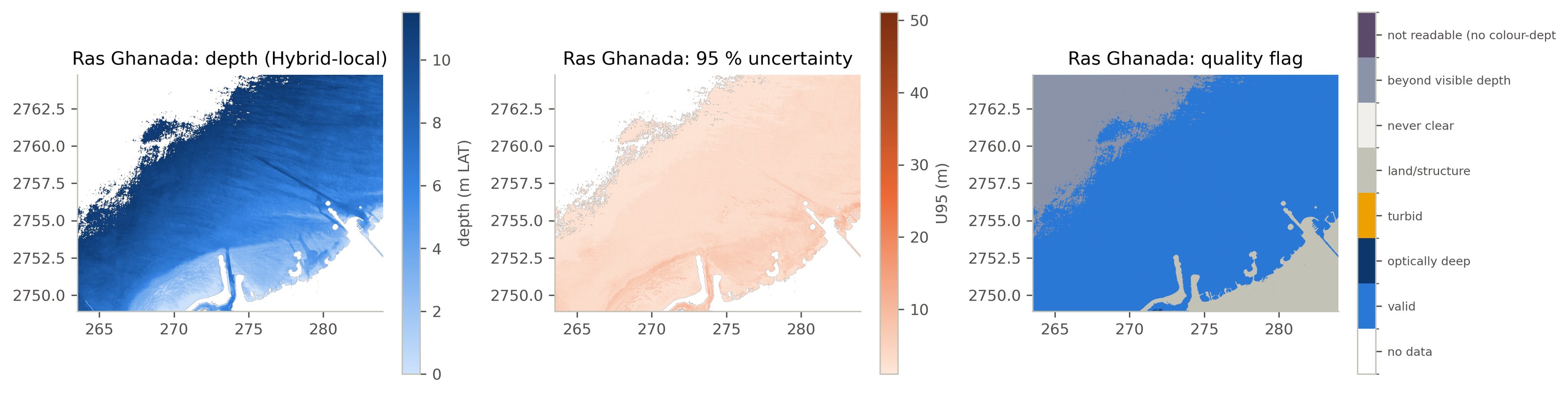}
\caption{Unsurveyed areas mapped from Sentinel-2 calibrated with ICESat-2 ATL24 alone: Mirfa West (top) and Ras Ghanada (bottom). Columns: depth below LAT, site-conformal 95\,\% uncertainty and quality flag. Coordinates in km (UTM 39N and 40N).}
\label{fig:unsurveyed}
\end{figure}

The two areas without any survey were mapped from four clear-season images each, calibrated with the ATL24 lidar alone (Table~\ref{tab:unsurveyed}; Fig.~\ref{fig:unsurveyed}). At Mirfa West, 5,159 lidar pixels over 113\,km$^2$ of water, the visible depth was 7\,m, three of the four images were calibrated (held-in $R^2$ of 0.55--0.72 and residual RMSE of 0.81--1.05\,m; the fourth, with $R^2=0.24$, was rejected), and 83\,km$^2$ (73\,\% of the water) received a depth, with a median of 3.6\,m and a median site-conformal U95 of 4.0\,m. At Ras Ghanada, 11,670 lidar pixels over 291\,km$^2$, the visible depth was 12\,m, all four images were calibrated ($R^2$ of 0.88--0.94, residual RMSE 0.79--1.09\,m) and 233\,km$^2$ (80\,\%) received a depth, with a median of 8.0\,m and a median U95 of 3.3\,m. The quality flag of the products (Fig.~\ref{fig:unsurveyed}) shows the two reasons for which the remaining water was not mapped: beyond the visible depth in the offshore part of each area, and never clear on a few turbid patches. These two maps are the first bathymetric information of any kind for the two areas that the authors are aware of beyond the chart; they have the accuracy that the lidar-only experiment of Section~\ref{sec:atl24results} suggests for areas where the passes cross the depth range (about 1--1.5\,m), which the dense ICESat-2 coverage of both areas (several thousand pixels across the full depth range) makes likely, but they have not been validated against soundings and should be read with their uncertainty layer.

\begin{table}[htbp]
\centering
\caption{Unsurveyed areas mapped with ICESat-2 ATL24 controls only (Hybrid-local, December 2024 to May 2025): lidar control pixels, images calibrated, visible depth, area of water and area mapped, and median depth and site-conformal U95 of the product.}
\label{tab:unsurveyed}
\small
\begin{adjustbox}{max width=\textwidth}
\begin{tabular}{@{}lrrrrrrrr@{}}
\toprule
Area & Lidar pixels & Images (calibrated) & Visible depth (m) & Water (km$^2$) & Mapped (km$^2$) & Coverage (\%) & Depth (m) & U95 (m) \\
\midrule
Mirfa West (53.28--53.46\degree E, 24.04--24.17\degree N) & 5,159 & 4 (3) & 7 & 113.4 & 82.6 & 72.9 & 3.6 & 4.0 \\
Ras Ghanada (54.66--54.86\degree E, 24.84--24.98\degree N) & 11,670 & 4 (4) & 12 & 291.2 & 232.9 & 80.0 & 8.0 & 3.3 \\
\bottomrule
\end{tabular}
\end{adjustbox}
\end{table}

\subsubsection{Year-to-year repeatability}
\label{sec:repeat}

\begin{figure}[htbp]
\centering
\includegraphics[width=\textwidth]{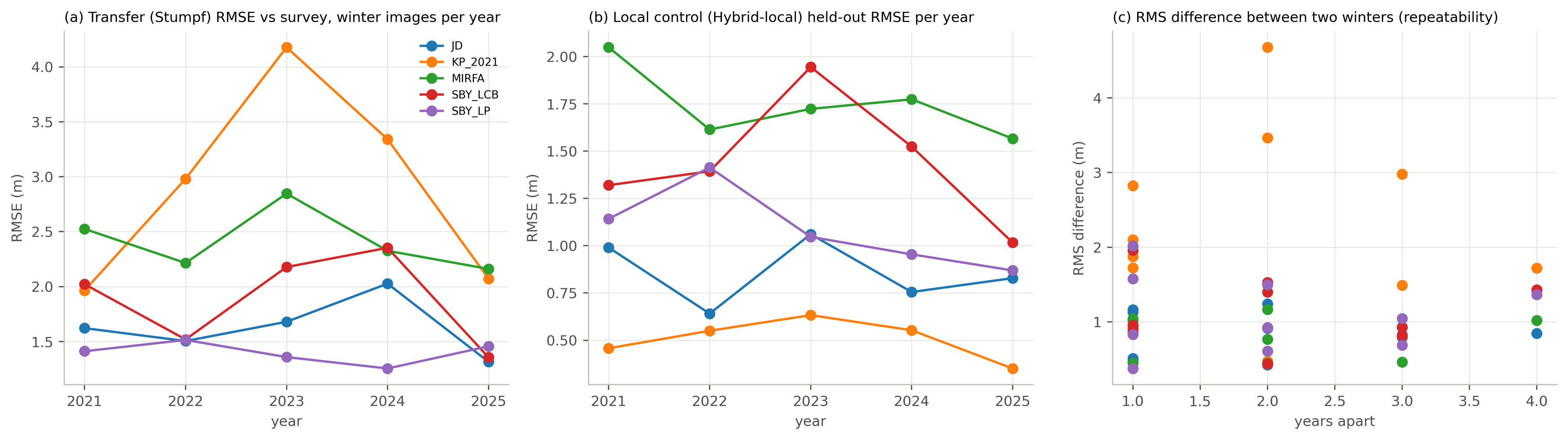}
\caption{Year-to-year behaviour of the maps made from the three clearest images of each winter: (a) transfer (Stumpf) RMSE against the survey; (b) held-out local-control (Hybrid-local) RMSE; (c) RMS difference between the transfer maps of two winters at their common pixels, against the number of years between them.}
\label{fig:repeat}
\end{figure}

Table~\ref{tab:repeat} and Fig.~\ref{fig:repeat} give the behaviour of the maps made from the three clearest images of each of the five winters from 2020/21 to 2024/25. At the three natural shallow sites the transfer error against the survey varied little from year to year: 1.26--1.52\,m at the Sir Bani Yas logistics port, 1.36--2.35\,m at the landing-craft basin and 2.16--2.85\,m at Mirfa, without a trend over the three to four years that separate the last winter from the survey; neither the seabed nor the water of these bays changed by more than the method can resolve. At the Khalifa Port limits the transfer error was 1.97\,m in the winter of the survey, 2.98 and 4.18\,m in the following two winters, 3.34\,m in the fourth and 2.07\,m in the fifth, with a bias that moved from $-1.9$ to $-4.1$\,m and back to $-2.0$\,m: in transfer mode this clear offshore site is read through a calibration from the port basins, and the year-to-year change of its water colour appears as a change of several metres in the map. With local control the same winters gave 0.35--0.63\,m at this site, 0.87--1.41\,m at the Sir Bani Yas sites, 1.57--2.05\,m at Mirfa and 0.64--1.06\,m at Jebel Dhanna (Fig.~\ref{fig:repeat}b), which shows that a few control depths remove most of the inter-annual variation. The repeatability of the transfer maps themselves, measured as the RMS difference between the fused maps of two winters at their common pixels (ten pairs per site; Fig.~\ref{fig:repeat}c; Table~\ref{tab:repeatpairs}), had a median of 0.84\,m at Jebel Dhanna, 0.93--0.98\,m at the three natural shallow sites and 1.99\,m at the Khalifa Port limits for the ratio, and 1.0--1.4\,m for LightGBM, with mean differences between winters of $-0.6$ to $+0.8$\,m. Two maps of the same bay made in different winters therefore agree to about 1\,m, which sets the smallest seabed change that can be detected by differencing two transfer products; with local control the detectable change is correspondingly smaller.

\begin{table}[htbp]
\centering
\caption{Year-to-year experiment: maps from the three clearest clear-season images of each winter (December--May), transfer RMSE and bias (Stumpf) and LightGBM RMSE against the survey, and held-out local-control RMSE and coverage of the hybrid (m, \%). Years from survey = time from the survey to the middle of the winter.}
\label{tab:repeat}
\small
\begin{adjustbox}{max width=\textwidth}
\begin{tabular}{@{}llrrrrrr@{}}
\toprule
Site & Winter & Years from survey & Stumpf RMSE & Stumpf bias & LightGBM RMSE & Hybrid-local RMSE & Coverage (\%) \\
\midrule
SBY\_LP & 2020/21 & $-1.0$ & 1.41 & $-0.22$ & 1.43 & 1.14 & 90.4 \\
 & 2021/22 & 0.0 & 1.52 & $+1.08$ & 1.86 & 1.41 & 88.1 \\
 & 2022/23 & 1.0 & 1.36 & $+0.28$ & 1.42 & 1.05 & 99.8 \\
 & 2023/24 & 2.0 & 1.26 & $+0.31$ & 1.23 & 0.95 & 99.9 \\
 & 2024/25 & 3.0 & 1.46 & $+0.98$ & 2.06 & 0.87 & 99.1 \\
SBY\_LCB & 2020/21 & $-1.0$ & 2.02 & $-1.02$ & 2.17 & 1.32 & 96.8 \\
 & 2021/22 & 0.0 & 1.52 & $-0.42$ & 1.80 & 1.39 & 92.6 \\
 & 2022/23 & 1.0 & 2.18 & $-1.01$ & 2.02 & 1.94 & 93.5 \\
 & 2023/24 & 2.0 & 2.35 & $-1.43$ & 2.09 & 1.52 & 75.5 \\
 & 2024/25 & 3.0 & 1.36 & $+0.17$ & 1.16 & 1.02 & 99.4 \\
MIRFA & 2020/21 & $-1.7$ & 2.53 & $-1.69$ & 1.87 & 2.05 & 96.5 \\
 & 2021/22 & $-0.7$ & 2.22 & $-1.07$ & 1.74 & 1.61 & 98.2 \\
 & 2022/23 & 0.4 & 2.85 & $-1.98$ & 2.24 & 1.72 & 100.0 \\
 & 2023/24 & 1.4 & 2.33 & $-1.06$ & 1.98 & 1.77 & 92.1 \\
 & 2024/25 & 2.4 & 2.16 & $-0.91$ & 1.74 & 1.57 & 97.2 \\
KP\_2021 & 2020/21 & $-0.1$ & 1.97 & $-1.92$ & 2.05 & 0.46 & 80.7 \\
 & 2021/22 & 0.9 & 2.98 & $-2.94$ & 2.88 & 0.55 & 100.0 \\
 & 2022/23 & 1.9 & 4.18 & $-4.14$ & 3.50 & 0.63 & 28.8 \\
 & 2023/24 & 2.9 & 3.34 & $-3.31$ & 3.12 & 0.55 & 55.6 \\
 & 2024/25 & 3.9 & 2.07 & $-2.02$ & 2.30 & 0.35 & 98.2 \\
JD (dredged) & 2020/21 & $-1.5$ & 1.62 & $+1.09$ & 1.89 & 0.99 & 99.0 \\
 & 2021/22 & $-0.5$ & 1.51 & $+1.12$ & 1.34 & 0.64 & 98.1 \\
 & 2022/23 & 0.5 & 1.68 & $+1.28$ & 1.92 & 1.06 & 97.2 \\
 & 2023/24 & 1.5 & 2.03 & $+1.19$ & 1.92 & 0.76 & 98.3 \\
 & 2024/25 & 2.5 & 1.32 & $+0.69$ & 1.02 & 0.83 & 99.9 \\
\bottomrule
\end{tabular}
\end{adjustbox}
\end{table}

\begin{table}[htbp]
\centering
\caption{Repeatability of the transfer maps: median RMS and mean difference (m) between the fused maps of two winters at their common pixels, over the ten pairs of winters per site.}
\label{tab:repeatpairs}
\small
\begin{tabular}{@{}lrrrr@{}}
\toprule
 & \multicolumn{2}{c}{Stumpf} & \multicolumn{2}{c}{LightGBM} \\
\cmidrule(lr){2-3}\cmidrule(lr){4-5}
Site & RMS difference & Mean difference & RMS difference & Mean difference \\
\midrule
JD & 0.84 & $-0.03$ & 1.03 & $-0.19$ \\
SBY\_LP & 0.98 & $+0.50$ & 1.38 & $+0.81$ \\
SBY\_LCB & 0.94 & $+0.29$ & 1.03 & $+0.56$ \\
MIRFA & 0.93 & $+0.37$ & 1.01 & $+0.28$ \\
KP\_2021 & 1.99 & $-0.60$ & 1.34 & $-0.34$ \\
\bottomrule
\end{tabular}
\end{table}

\section{Discussion}
\label{sec:discussion}

\subsection{Why the accuracy is about 1\,m, and why published values are often lower}
\label{sec:disc_accuracy}

The central quantitative result of this study is that, under a validation design in which no test pixel has a calibration pixel within 225\,m, the accuracy of Sentinel-2 SDB in Gulf port waters is 1.1--1.25\,m with local control and 1.45\,m in transfer, over 0--11\,m, with a further 0.3--0.5\,m achievable in clear water of uniform optical properties (the Khalifa Port limits). These values are considerably larger than the 0.3--0.6\,m frequently reported for Sentinel-2 at a single site \cite{Caballero2019,Traganos2018,Evagorou2022}, and also larger than the 0.29\,m obtained with a convolutional network on part of the same Jebel Dhanna survey in earlier work of our group. The leakage experiment of Section~\ref{sec:leakage} explains most of the difference: on the same data, the same model and the same settings, a random pixel split reports 0.57\,m where buffered spatial blocks report 1.30\,m and an unseen site 1.66\,m. The random split does not measure the error of a depth that the model has to infer from colour; it measures the error of interpolating between soundings 10--20\,m away, which the spatial autocorrelation of depth makes small. The problem is well documented in ecological and soil mapping \cite{Roberts2017,Ploton2020} and applies with full force to SDB, where the reference depths are dense survey grids. SDB accuracies in the literature can therefore be compared with the present ones only when the validation design is stated, and a sub-decimetre or even sub-half-metre accuracy obtained from a random split should not be used to decide whether an SDB product meets a hydrographic standard.

The remaining 1\,m has several sources that can be named. The seabed signal fades with depth, so the error grows beyond 9--10\,m in every method, and the shallowest water, which vessel surveys sample poorly, is over-estimated by models trained on 2--11\,m water; the shallow hinge and the lidar controls of the present version lowered the 0--2\,m RMSE of the local-control method from 1.17\,m in a preliminary run to 1.04\,m, and the over-estimation seen by the lidar at Jebel Dhanna from 0.8 to 0.7\,m, but a bias of $+0.5$\,m remains in that band. The water of these ports is not optically uniform: the tree model gains about 10\,\% on the ratio method inside a surveyed area and loses that advantage at an unseen site, so the gain comes from site-specific combinations of bottom brightness and water colour that do not carry over. Each Sentinel-2 pixel integrates 100\,m$^2$ of seabed, whereas the surveys resolve slopes, berth pockets and dredged edges that the pixel averages; aggregation to 20\,m cost the ratio method nothing, which says that the seabed of these sites is smooth at that scale, not that the 10\,m pixel resolves everything the survey does. The images were acquired 0--20 days from the surveys at eleven datasets, but the dredged basins are maintained and may have changed between the dates, and at Mirfa the clear-season images are 53 days from the survey. The tide at overpass, 0.1--2.2\,m above LAT, came from a model that agrees with the gauges to within a few decimetres (supplementary Fig.~\ref{fig:tidecheck}), and an error of 0.2\,m in $\eta$ enters the depth directly. The conformal intervals, whose observed coverage matched the nominal 95\,\% at the training sites, show that these error sources are at least well quantified where the water resembles the calibration population; the site-conformal interval extends that property to new areas at the cost of a wider interval.

\subsection{Turbidity, visible depth and the limits of the method}
\label{sec:disc_turbidity}

Two depth limits emerged from the data and both should be stated with any SDB product for these waters. The first is the visible depth of an image, which the colour--depth walk of Section~\ref{sec:depthlimit} measured at 8--9\,m in the dredged basins of Jebel Dhanna and Old Musaffah, 6--10\,m at Mugharraq and 11--12\,m in the clearer water of Sir Bani Yas, Mirfa and the Khalifa Port limits, and which the images of the 18--21\,m basins did not have at all. The 95th percentile of the usable training depths, 11\,m, was adopted as the transfer limit, and the results by 1\,m bin confirm that the local-control error is roughly constant to 10\,m and nearly doubles in the 10--11\,m bin. The second limit is the one that the seabed-not-readable test enforces: in turbid dredged basins the water is brighter than the offshore reference and the pixels pass the signal-to-noise test, so that a model unaware of the limit publishes depths of about 10\,m for 18--21\,m basins. The pipeline refused these depths, withheld the Khalifa Port basin, EGA basin and Old Musaffah maps, and flagged 75--78\,\% of the pixels deeper than the limit, which is the behaviour that an operational product must have. The coast-wide experiment showed that the same problem arises offshore, where deep water can be brighter than the darkest 2\,\% of a context window, and that no brightness test, whether based on the noise or on the spread of the deep-water set, removes it; only the lidar check located the affected area.

The seasonal experiment at the seven sites, together with the coast-wide climatology, describes how the clarity of UAE coastal water as seen by Sentinel-2 varies through the year and what that variation costs in SDB accuracy and coverage. The turbidity index of the images was five to six times higher in the summer images than in the winter ones at the survey sites, and the coast-wide climatology of 2024 showed the median index of the 49 coastal cells rising from 0.001--0.004 in September--April to 0.008--0.025 in May--August, with the share of cells holding a clear image falling from 75--92\,\% to 42--55\,\%; the pattern matches the summer increase in phytoplankton and wind-driven resuspension in the southern Gulf \cite{Sheppard2010}. The effect on the error of local control was moderate (1.02--1.04 to 1.38\,m pooled) because calibration partly compensates for a brighter water column, but the effect on coverage was severe: in summer only two or three of the seven sites could be calibrated with the ratio method and the hybrid covered 70\,\% of the pixels against 93--98\,\% in the other seasons. Images for SDB on this coast should therefore be chosen from November to April, when two or three clear months are available in nearly every cell of the coast, and the choice should rest on a measured turbidity index rather than on a seasonal prior. Mirfa shows what the season is worth: calibrated on winter images the site gave 1.84\,m, where a preliminary run with its autumn images, those closest to the survey, had given about 3.9\,m, and in the seasonal experiment the autumn images could not be calibrated at all.

\subsection{Transfer versus local control, and the ranking of the models}
\label{sec:disc_transfer}

On the second objective of the study the answer is plain: a model trained on seven surveys of the same coast did not transfer to an unseen port with an accuracy better than the two-parameter Stumpf ratio, and neither the gradient-boosted trees nor the U-Net transferred better than the ratio. The tree model was the best within surveyed areas (1.30\,m against 1.45\,m), the U-Net was the best at Mugharraq and the EGA basin, the Lyzenga transform was the best at Old Musaffah, Mirfa and the Khalifa Port channel, and the ratio failed at Mirfa, whose clear winter water lies outside the range of the training ratios; none of these advantages was systematic. The SHAP analysis suggests why: the learned models rely on the same smoothed ratio signal as the classical method and use the remaining features to correct it for local conditions that the sites do not share. The machine-learning and deep-learning models thus pay for their complexity when the training data cover the area to be mapped, i.e.\ when the problem is interpolation within a port, and not when the problem is extrapolation to a new port. The same was found in the review of \cite{Kutser2020}, in the multi-site study of \cite{Sagawa2019}, who obtained 1.4\,m across five areas with a random forest, and in the cross-regional test of \cite{Hsu2026} in clear reef waters, where a random forest and four deep-learning encoders reached 1.2--1.9\,m within a region but 2.5--3.8\,m at reefs never used in training. These results argue against the expectation that a large enough training set will produce a universal SDB model for a coast as optically variable as this one. The year-to-year experiment shows the same thing in time: at the Khalifa Port limits the transfer bias wandered between $-1.9$ and $-4.1$\,m over five winters, while the local-control error stayed between 0.35 and 0.63\,m.

Local control is a different matter. With about 20\,\% of a site's soundings used to calibrate each image, the hybrid of the tree output, the ratio and the shallow hinge reached 1.11\,m pooled and 0.3--0.5\,m at the temporal site, which is the accuracy that matters to a port authority that surveys regularly and wants to update the chart between surveys. The advantage of the hybrid over the pure ratio (1.24\,m) and the pure re-scaled tree (1.25\,m) lay in the shallowest band, where the ratio and hinge terms let it follow controls below 2\,m, and in the depth-balanced error (1.25 against 1.47--1.55\,m). The experiments on the number of controls and images give the operating envelope of this mode: about 60--125 control pixels per image for most of the accuracy, 2,000 or more for the last decimetre, and three images for two thirds of the gain of fusion. The fusion weights of Eq.~\eqref{eq:fusion}, which include the calibration residual of each image, were introduced after a preliminary run had shown the error rising beyond three images; with them the fusion improved monotonically, so an operator no longer needs to limit the number of images.

\subsection{ICESat-2 as a check, as a substitute for soundings and as the way into unsurveyed areas}
\label{sec:disc_atl24}

The agreement of the offset-corrected ATL24 depths with the MBES surveys, 0.25\,m at Jebel Dhanna and 0.37\,m at the Khalifa Port basin, is of the same order as the lidar's own stated uncertainty and as the values reported by \cite{Parrish2019} and \cite{Babbel2021}, and it validates both the lidar and the tide reduction of this study, since the two were reduced to LAT independently. The systematic offset of $+0.28$\,m $+3.5\,\%$ of depth, with the lidar reading deeper, was found at every verified pass before correction. Its depth-proportional part is of the size of a refraction-index error, and its constant part of the size of a sea-surface determination error; both are plausible residuals of the ATL24 processing in the warm, hypersaline Gulf water, whose refractive index differs from that of standard seawater, but the present data cannot separate the causes. Users of ATL24 for calibration in similar waters should expect and check for such an offset. The three rejected passes over the deep Khalifa Port waters, with false seabed photons at 1--6\,m, show that the product must be verified before use in turbid or deep water.

Calibration with lidar alone approached the sounding-based accuracy at the sites where the passes crossed a representative depth range, and failed where the passes lay only on the shallow margins. ICESat-2 is therefore a usable source of control for SDB in these waters when a pass crosses the area of interest at several depths, and the lidar-only mode of the chain delivers a map with its own uncertainty in that case; where the tracks miss the deeper water, the calibration extrapolates and the error grows. The independent check of the maps against the check passes revealed a bias of $+0.7$ to $+3.4$\,m in water shallower than 2\,m, which the vessel surveys could not reveal because they do not sample it, and the addition of the verified lidar pixels to the control set of the local-control test is the first step towards removing it. The two unsurveyed areas show what this means in practice: with several thousand lidar pixels across the depth range, Mirfa West and Ras Ghanada received depth maps with $R^2$ of 0.55--0.94 against the lidar and a stated uncertainty of 3--4\,m at 95\,\%, a reconnaissance product obtained without a vessel for 315\,km$^2$ of water that has never been surveyed. A few check lines across these areas would make them the first independent test of the lidar-only mode outside the survey sites.

\subsection{Very-high-resolution imagery}
\label{sec:disc_vhr}

The Pl\'eiades test must be read with its two caveats: the image was a display product without radiometric calibration, and it was acquired four years after the surveys of a maintained basin. Under these conditions the best of the three transfer models gave 2.19\,m at 2\,m pixels and 1.94\,m at 5\,m, against 1.57\,m for the Sentinel-2 local-control map at the same pixels, and the 2\,m map resolved the berth pockets and slopes that the 10\,m map could not. This shows that the chain can ingest VHR imagery and that VHR adds spatial detail; it does not show an accuracy advantage. The ratio and Lyzenga models improved when the image was aggregated from 2 to 5--10\,m, which means the 8-bit product is noise-limited at full resolution, and the Sentinel-2 map, with its six calibrated images, was the more accurate product. The fair comparison, a calibrated VHR image over a clear 0--10\,m site acquired within days of a survey, was not possible with the imagery available and remains the main gap of this study.

\subsection{What SDB can deliver for the waters of the UAE}
\label{sec:disc_outlook}

The outlook experiments place the site results in the context of the country's waters. The coast-wide climatology shows that the Sentinel-2 archive offers, in a single year, a median of nine clear months per $0.25\degree$ cell and at least five in every cell of the Al Dhafra coast, the northern emirates and the east coast; only the urban coast of Abu Dhabi island, the Musaffah channel and Khalifa Port, where dredging, reclamation and shipping keep the water turbid, had no clear month in 2024. Since three clear images are enough for a product, the archive is not the limiting resource anywhere outside those four cells. The mosaic of reconnaissance depths covers the shallow banks, channels and reefs of the whole coast at 60\,m from one year of data, 17,233\,km$^2$ of the 19,263\,km$^2$ examined. It is a transfer product, however, that carries the uncertainty of the independent sites and, because no single-image brightness test separates deep bright water from 8--10\,m water, an admixture of deep water mapped at 8--10\,m. The ICESat-2 check put that admixture at about 15\,\% of the mapped area, nearly all of it offshore, and showed that on the coastal shallows the mosaic is within about 1\,m only between 5 and 8\,m and 2--3.5\,m too deep on the flats, which is what the transfer test against the surveys had predicted. The agreement matters in its own right: an error model derived at six port areas held along some 350\,km of open coast against an independent reference, so the survey-site error budget can be used for products outside the sites. The lidar then did what the soundings do at the ports. Calibrated on it cell by cell and tested on held-out blocks, the mosaic went from 2.9 to 1.4\,m and from a bias of $+2.0$\,m to $+0.2$\,m over the 24 cells with enough lidar, the flats from 3.4 to 1.7\,m, and the share of pixels within Order~2 from 17 to 69\,\%; the best cells reached 0.6--0.9\,m. That is the accuracy of the locally controlled port maps, obtained over half of the coast product from public data alone. The calibration cannot recover what the image does not resolve: the 8--11\,m band stays the weakest at 1.5\,m because the ratio saturates there, and the cells east of 54.7\degree E wait for more passes. The value of the uncalibrated mosaic is in locating the shallow water and in planning surveys; the calibrated one can also serve as a first chart-scale reconnaissance of the flats. Where ICESat-2 passes cross an area, the lidar-only mode delivers a calibrated map with an uncertainty layer for areas that have never been surveyed, as the Mirfa West and Ras Ghanada products show. The year-to-year experiment shows that two transfer maps of the same bay made in different winters agree to about 1\,m, and local-control maps to a few decimetres in clear water; this sets the change that can be detected by differencing products of successive years: a dredged basin that silts by 1\,m or a bank that migrates by a pixel can be seen, a change of a few decimetres cannot without control depths. Together these results describe an SDB service for the UAE that is feasible with public imagery and public lidar: a coast-wide layer refreshed every winter and calibrated on ICESat-2 wherever it passes, lidar-calibrated maps for unsurveyed areas, and locally controlled maps to about 1\,m for the ports, with a known seasonal window and a known detection threshold for change.

\subsection{Operational guidance for a port authority}
\label{sec:disc_operational}

The results support the following recommendations for the use of SDB by port authorities in the Gulf. (i)~SDB from Sentinel-2 is appropriate for reconnaissance, for the monitoring of change between surveys and for the planning of surveys, to a depth of about 10--11\,m in clear water and about 8--9\,m in dredged basins, and it is not appropriate for the 18--21\,m basins. (ii)~Local control should be used whenever any soundings exist: a single-beam line of about a hundred 10\,m pixels across the depth range, or an older survey, raises the accuracy from 1.45 to 1.1--1.25\,m on average and to 0.3--0.5\,m in clear water, and the calibration of every image should be checked by its $R^2$. (iii)~Images should be selected from November to April by a measured turbidity index, three images per site giving most of the benefit of fusion. (iv)~Every product should carry the depth limit, the quality flag and a calibrated uncertainty, the site-conformal one for areas without soundings, and the estimated S-44 Order derived from it. On the present evidence, Sentinel-2 SDB with local control can be expected to meet Order~2 at the 95\,\% level only in clear water of uniform optical properties (as at the Khalifa Port limits), to approach Order~1a/1b there, and to meet neither Order in dredged basins or on turbid coasts; its depths are therefore to be used as supporting information and not as charted depths, consistent with the guidance of the IHO on SDB \cite{IHO2019}. (v)~ICESat-2 ATL24 should be downloaded for every area of interest, verified against any available survey, corrected for the offset and used both as an independent check, as a source of shallow-water control and, where no soundings exist, as the sole calibration. (vi)~For areas that are re-mapped every year, a change smaller than about 1\,m between two transfer products should not be interpreted as seabed change.

\subsection{Limitations}
\label{sec:limitations}

Several limitations qualify these conclusions. The surveys are of ports and their immediate surroundings, which are dredged, turbid and deep by the standards of SDB; the results do not describe the accuracy achievable on the clear carbonate platforms and reefs of the western Gulf, where visible depths of 15--20\,m may be expected, and the coast-wide mosaic was checked and calibrated against ICESat-2 mainly on the western two thirds of the coast and against a reference (the lidar) whose own accuracy, 0.25--0.94\,m against the surveys, is not negligible on the flats; the calibrated cells have no check independent of the lidar. Only one independent site of each type (natural bay, turbid channel) was available, so that the independent-site results are indicative rather than statistical. The atmospheric correction was that of Sen2Cor; a comparison with a coastal processor such as ACOLITE was not made, although the deep-water reference removes most of the residual path reflectance for the ratio and Lyzenga features. The seasonal experiment used three images per season from one year per site and the coast-wide climatology one year, 2024, so that the inter-annual variability of the clear season is not known; the image selection in the main experiment was constrained to within 60 days of the survey at the dredged sites, so that the seasonal and the main results are not fully comparable. The VHR test is qualified as discussed. The U-Net was trained with a modest budget (two seeds per region, 40 epochs) on the same hand-crafted features as the tree model, and a network trained on raw bands with a larger budget might transfer differently. The lidar comparisons rest on 10--61 overlap pixels per site, and the lidar offset, fitted on 91 pixels, should be re-estimated as more passes accumulate; the maps of the unsurveyed areas are calibrated on all the lidar of each area and have no independent check. The per-site map rule withheld the map of the temporal site on a coverage threshold although the local methods were accurate there, which shows that the rule needs a provision for sites that the images read over only part of their area. Finally, the pooled conformal intervals are calibrated on the training population and, as the independent sites showed, do not anticipate a site whose water lies outside it; the site-conformal interval is the one to attach to a product for a new site, and a few check depths remain advisable before its uncertainty is relied upon.

\section{Conclusions}
\label{sec:conclusions}

Satellite-derived bathymetry was benchmarked in the turbid, tidal port waters of the Abu Dhabi coast of the Arabian Gulf against 285,619 multibeam soundings from twelve surveys, with four model families, two operating modes and a validation protocol in which no test pixel is predicted by a fit that used it or its neighbours, and the same chain was then applied to the whole coast of the UAE, to two unsurveyed areas and to five successive winters. The following conclusions are drawn.

\begin{enumerate}[leftmargin=*]
\item The validation design dominates the reported accuracy. The same LightGBM model gave 0.57\,m RMSE under a random pixel split, 1.30\,m under buffered spatial blocks and 1.66\,m at an unseen site. SDB accuracies are comparable between studies only when the design is stated, and claims of Order-level accuracy from random splits are not transferable to operational use.
\item In transfer to an unseen site, the two-parameter Stumpf ratio (1.45\,m) was not outperformed by the gradient-boosted trees (1.66\,m), the U-Net ensemble (1.77\,m) or the Lyzenga transform (2.00\,m), although the ratio failed at the one site whose clear water lay outside the range of the training ratios. The learned models were better only within surveyed areas (LightGBM 1.30\,m against 1.45\,m), so their advantage comes from interpolation and not from a transferable understanding of the water.
\item With local control, i.e.\ about 20\,\% of a site's soundings and the verified ICESat-2 pixels used to calibrate each image, the hybrid of the tree output, the band ratio and a shallow-water hinge reached 1.11\,m pooled over 0--11\,m, nearly uniform across depth bands to 10\,m, with 67\,\% of pixels within the S-44 Order~2 TVU and a conformal 95\,\% interval whose observed coverage was 95.0\,\%. At the temporal test site, in clear water of 8.5--11\,m, local control gave 0.29--0.53\,m with 76--92\,\% of pixels within Order~1a/1b, and at the independent natural bay 1.84\,m.
\item The seabed was readable to 8--9\,m in dredged basins and 11--12\,m in clear water; a data-driven limit of 11\,m was derived, and the chain correctly withheld depths for the 18--21\,m turbid basins, where every method failed. No method met any S-44 Order at the 95\,\% level in any 1\,m depth bin at the training sites.
\item Water turbidity was five to six times higher in the summer images than in the winter ones at the survey sites, and over the 49 coastal cells of the UAE the median turbidity index rose from 0.001--0.004 in September--April to 0.008--0.025 in May--August; the summer images degraded the local-control error from 1.02--1.04 to 1.38\,m and the calibrated coverage from 93--98 to 70\,\%. November to April is the season for SDB on this coast, and the Sentinel-2 archive offers a median of nine clear months a year per cell, with none only on the urban coast of Abu Dhabi and at Khalifa Port.
\item Accuracy saturated at about 60--125 control pixels per image, and fusion with calibration-weighted images improved monotonically to six images, three of which gave two thirds of the gain; aggregation to 20\,m cost the ratio method nothing and the hybrid 0.15\,m, whereas 30--60\,m pixels raised the error by 10--50\,\%.
\item ICESat-2 ATL24 agreed with the surveys to 0.25--0.37\,m after removal of an offset of $+0.28$\,m $+3.5\,\%$ of depth, revealed a $+0.7$ to $+3.4$\,m over-estimation of the transfer maps in water shallower than 2\,m that the surveys do not sample, calibrated the images nearly as well as soundings where its passes crossed a representative depth range, and delivered the first depth maps, with an uncertainty layer, of two unsurveyed areas of 113 and 291\,km$^2$.
\item Checked against 110,039 ICESat-2 pixels in 47 coastal cells, the coast-wide transfer mosaic had 2.91\,m RMSE on lidar-shallow water, with an error curve by depth that reproduced the one found against the multibeam surveys (minimum 0.7\,m at 6--7\,m, 3--3.5\,m on the flats), and about 15\,\% of its mapped area lay over water deeper than the limit, mostly offshore; the survey-site error budget therefore transfers to the open coast. Calibrated on the lidar cell by cell and tested on held-out blocks, the mosaic reached 1.42\,m with a bias of $+0.24$\,m over the 24 cells with enough lidar (8,600\,km$^2$, half of the product), 69\,\% of pixels within Order~2 and 0.6--0.9\,m in the best cells, without any vessel survey.
\item Maps of the same bay made from the clearest images of different winters agreed to about 1\,m in transfer mode, and the local-control error at the clear temporal site stayed between 0.35 and 0.63\,m over five winters, which sets the seabed change that can be detected by differencing successive products.
\item A single 0.5\,m Pl\'eiades-1A display image acquired four years after the survey gave 1.9--2.2\,m at Khalifa Port against 1.57\,m for the locally controlled Sentinel-2 map at the same pixels; a calibrated VHR image over a clear shallow site remains to be tested.
\end{enumerate}

For a port authority, Sentinel-2 SDB with about a hundred control soundings per image is recommended as a reconnaissance and change-monitoring tool to about 10\,m, delivered with its depth limit, quality flag and calibrated uncertainty; its depths should not be charted. For the waters of the UAE as a whole, the results describe a feasible service built on public imagery and public lidar: a coast-wide layer refreshed every winter and calibrated on ICESat-2 wherever it passes, lidar-calibrated maps for unsurveyed areas and locally controlled maps for the ports.

\paragraph{Future work.}
\label{sec:future}
Five extensions follow directly from the limitations of the present study. (i)~The lidar calibration of the coast product will be extended to the cells east of 54.7\degree E, for which only a few passes were obtained within the download budget of this run, and improved with a second hinge at the deep end, so that the flats and the 5--8\,m band are fitted at the same time, and with calibration per image rather than of the fused mosaic; a per-cell conformal quantile would also tighten the interval of the calibrated product, which the largest-quantile rule leaves at twice the width it needs where the lidar is dense. (ii)~The maps of the two unsurveyed areas will be validated against single-beam check lines, which would make them the first independent test of the lidar-only mode outside the survey sites. (iii)~The deep-learning branch will be retrained on raw bands with a larger network and ensemble, which the in-house computing resources now available to the authors (two NVIDIA H100 GPUs and 256 CPU cores) make practical, to test whether a deep model trained without hand-crafted features transfers better than the band ratio. (iv)~A radiometrically calibrated VHR image will be acquired over a clear 0--10\,m site within days of a survey, to settle the question of the VHR accuracy advantage. (v)~The per-site map rule will be revised for sites that the images read over only part of their area, and the benchmark extended to the reef and carbonate-platform waters of the western Gulf, where visible depths of 15--20\,m may be expected.

\section*{Data and code availability}
The multibeam surveys are the property of the port operator and cannot be redistributed; the validation points (predicted depth, surveyed depth, uncertainty and flags at every held-out pixel) can be provided to reviewers on request. The Sentinel-2 L2A imagery is available from the Microsoft Planetary Computer (the maps in this paper contain modified Copernicus Sentinel data, 2020--2025), the ICESat-2 ATL24 granules from the NASA Earthdata archive and the EOT20 tide model from SEANOE. The coast-wide reconnaissance mosaic, its lidar-calibrated version and the maps of the two unsurveyed areas (GeoTIFF, with uncertainty and quality layers) will be deposited with the notebook. The complete processing notebook (ADMA-SDB version 1.7), which reproduces every table and figure of this paper from the raw soundings and the public data, together with the run manifest, will be made publicly available on publication.

\section*{CRediT authorship contribution statement}
\textbf{Muhammad Saeed:} Conceptualization, Methodology, Software, Validation, Formal analysis, Investigation, Data curation, Writing -- original draft, Writing -- review and editing, Visualization, Project administration. \textbf{Mohamed H. Alhosani:} Writing -- review and editing, Visualization, Project administration. \textbf{Dmitry Mikhaylov:} Writing -- review and editing, Visualization, Project administration. \textbf{Yasser F. Al Wahedi:} Writing -- review and editing, Visualization, Project administration, Supervision, Resources.

\section*{Declaration of competing interest}
The authors declare that they have no known competing financial interests or personal relationships that could have appeared to influence the work reported in this paper.

\section*{Acknowledgements}
The authors thank the hydrographic survey team of the Abu Dhabi ports for the multibeam surveys and tide-gauge records, the European Union's Copernicus programme and the European Space Agency for the Sentinel-2 data, the Microsoft Planetary Computer for access to the Sentinel-2 archive, and NASA's National Snow and Ice Data Center for the ICESat-2 ATL24 product. This research received no specific grant from any funding agency.

\bibliographystyle{unsrtnat}
\bibliography{refs}

\appendix
\section{Supplementary figures}
\label{sec:supp}

\begin{figure}[htbp]
\centering
\includegraphics[width=0.9\textwidth]{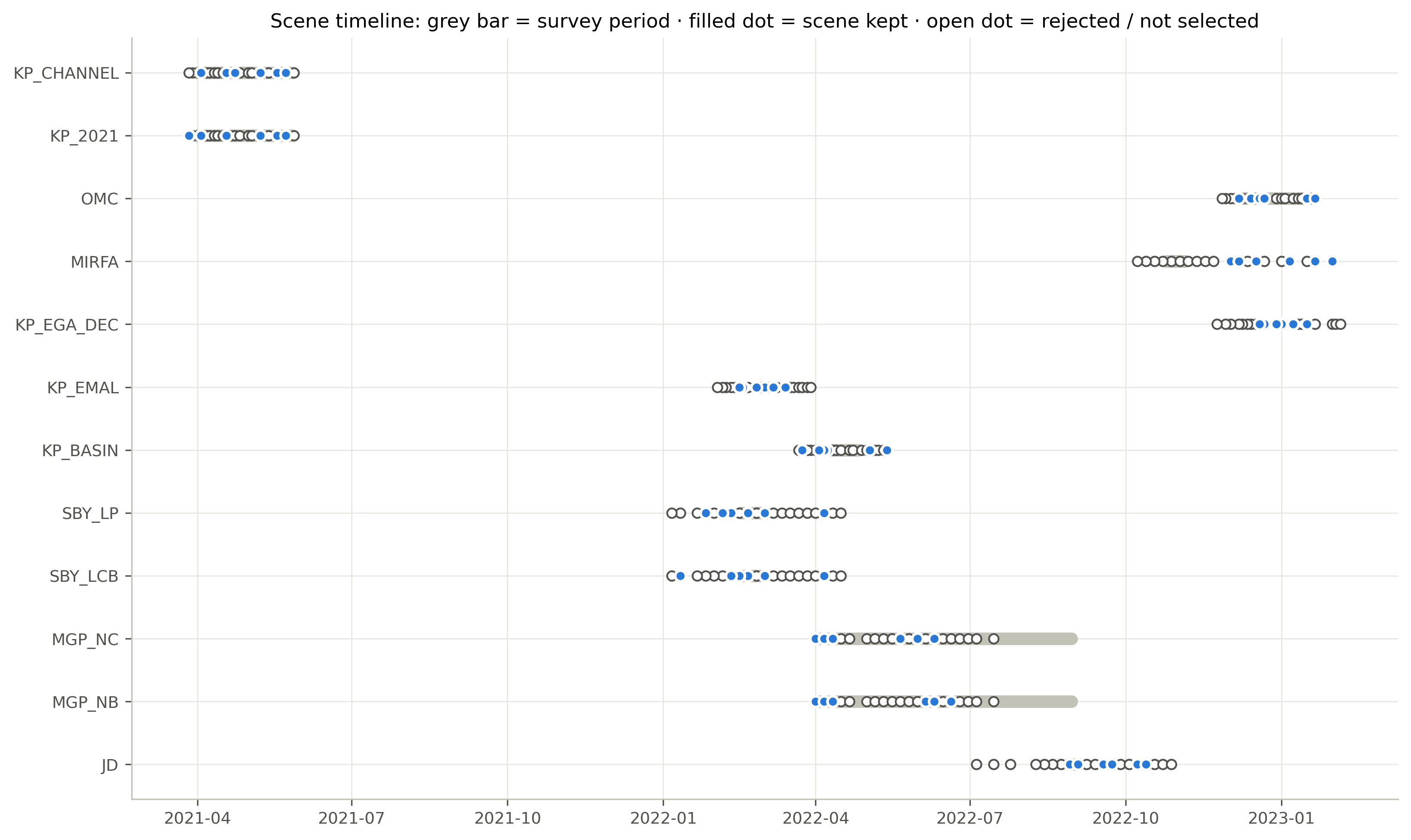}
\caption{Sentinel-2 images examined (open dots) and kept (filled dots) for each survey; grey bars give the survey period.}
\label{fig:timeline}
\end{figure}

\begin{figure}[htbp]
\centering
\includegraphics[width=0.6\textwidth]{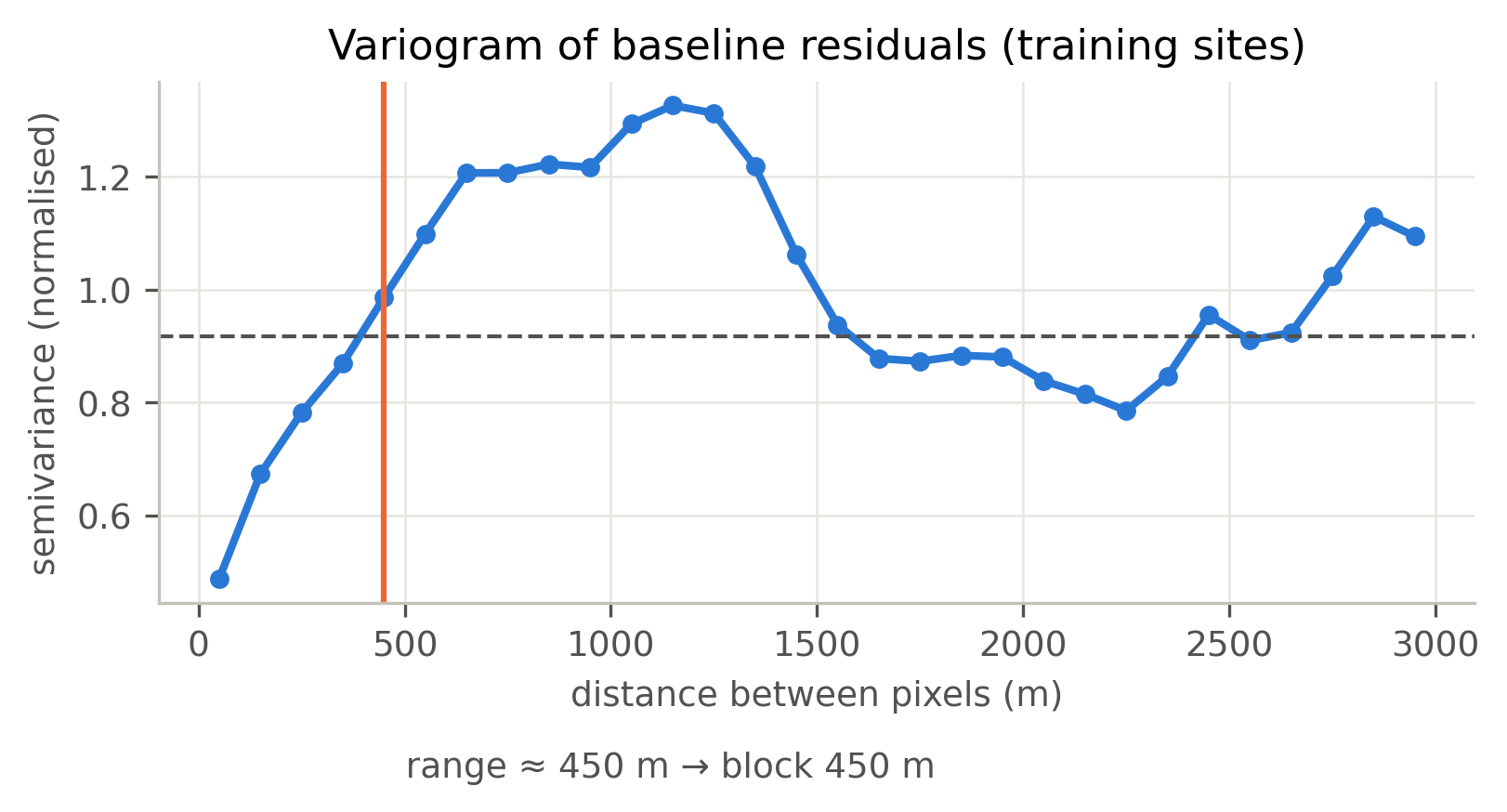}
\caption{Empirical variogram of the residuals of a preliminary Stumpf fit at the training sites, used to size the spatial validation blocks (range about 450\,m).}
\label{fig:variogram}
\end{figure}

\begin{figure}[htbp]
\centering
\includegraphics[width=\textwidth]{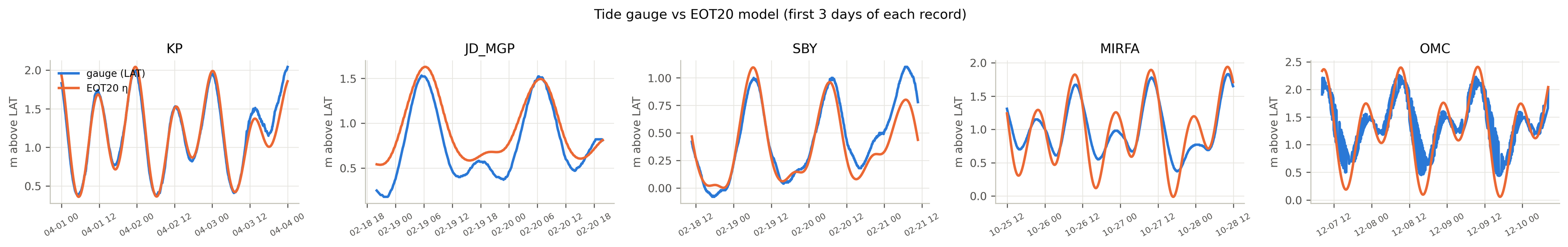}
\caption{EOT20 tide prediction against the survey tide gauges (first three days of each record, after the datum offsets of Section~\ref{sec:tides}).}
\label{fig:tidecheck}
\end{figure}

\begin{figure}[p]
\centering
\includegraphics[width=\textwidth]{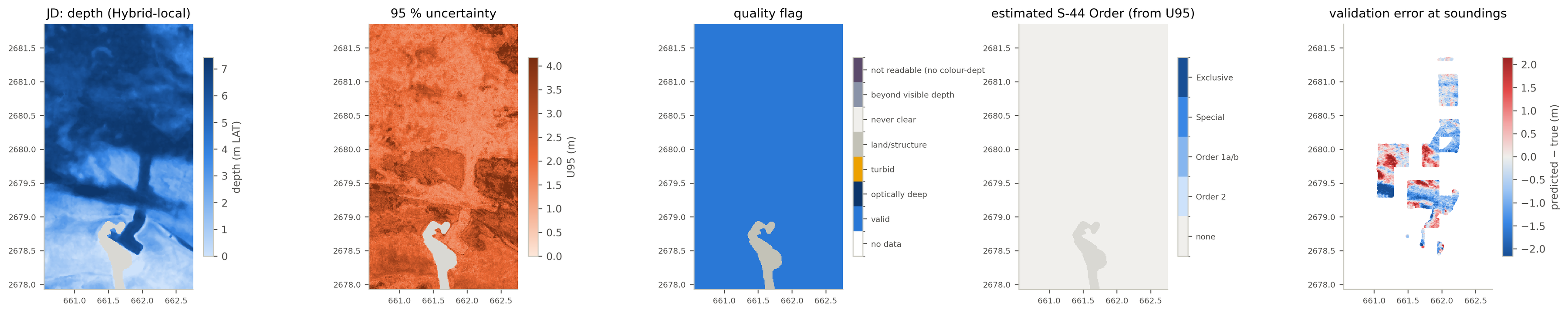}\\[1mm]
\includegraphics[width=\textwidth]{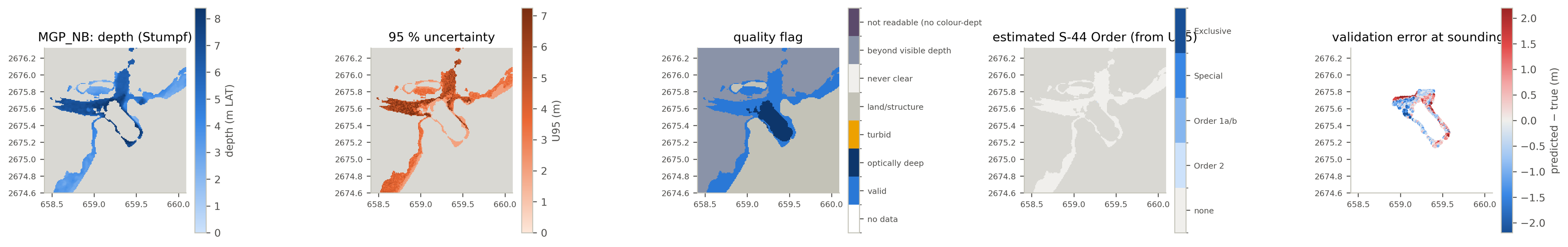}\\[1mm]
\includegraphics[width=\textwidth]{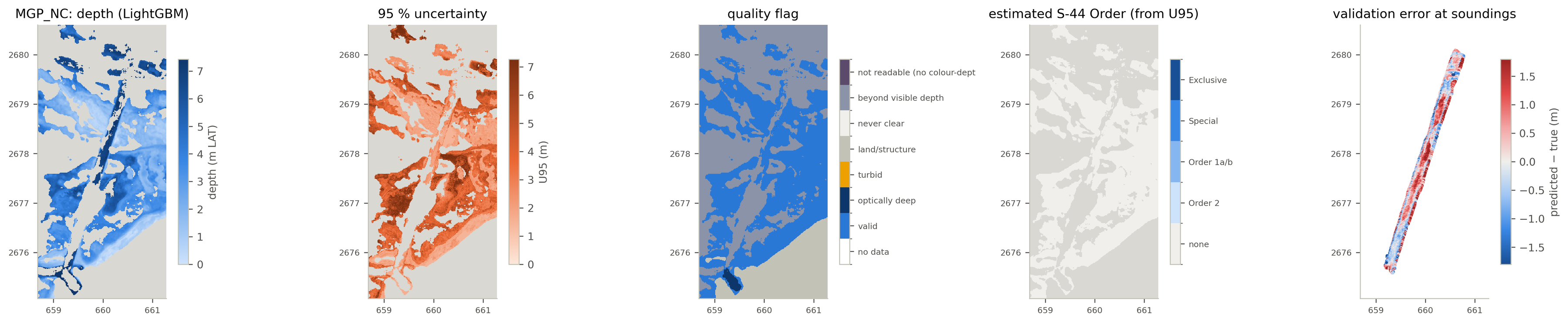}\\[1mm]
\includegraphics[width=\textwidth]{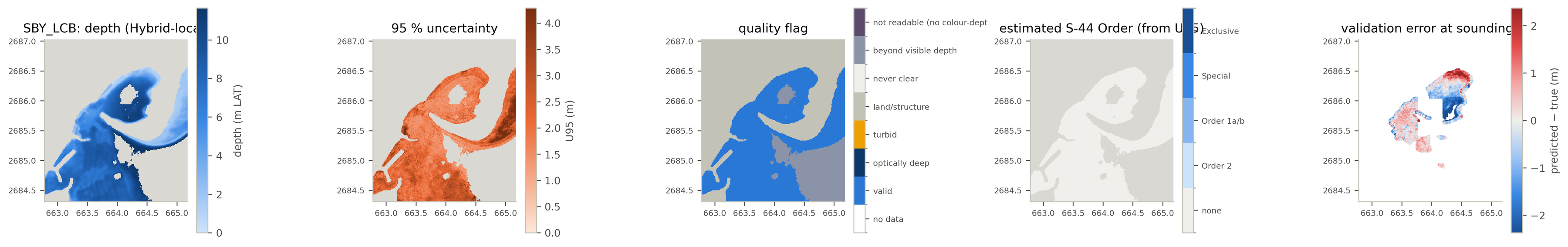}\\[1mm]
\includegraphics[width=\textwidth]{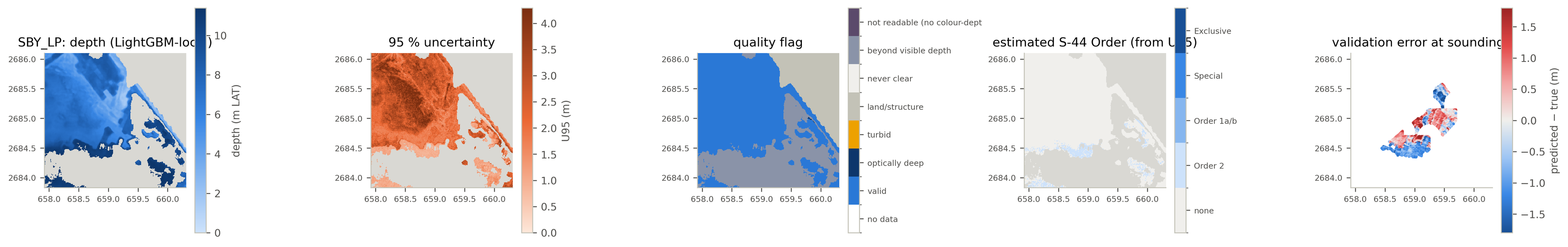}\\[1mm]
\includegraphics[width=\textwidth]{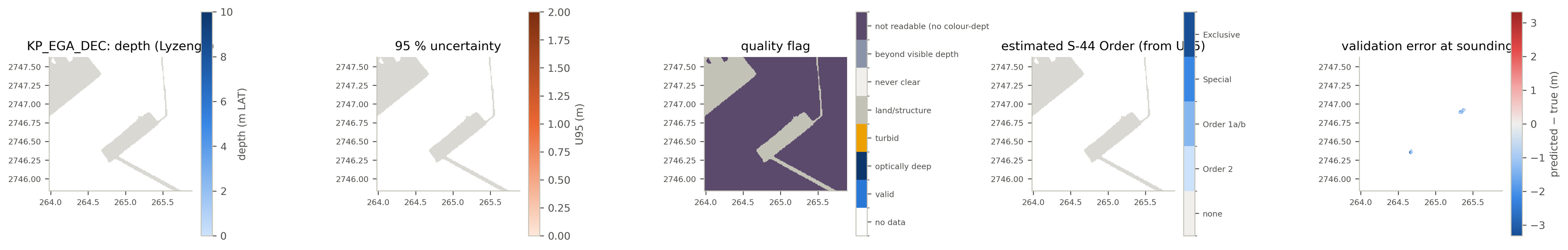}
\caption{Full product panels (depth, 95\,\% uncertainty, quality flag, estimated S-44 Order from U95, held-out error at the soundings), part 1: JD, MGP\_NB, MGP\_NC, SBY\_LCB, SBY\_LP and the withheld map of KP\_EGA\_DEC.}
\label{fig:panels}
\end{figure}

\begin{figure}[p]
\centering
\includegraphics[width=\textwidth]{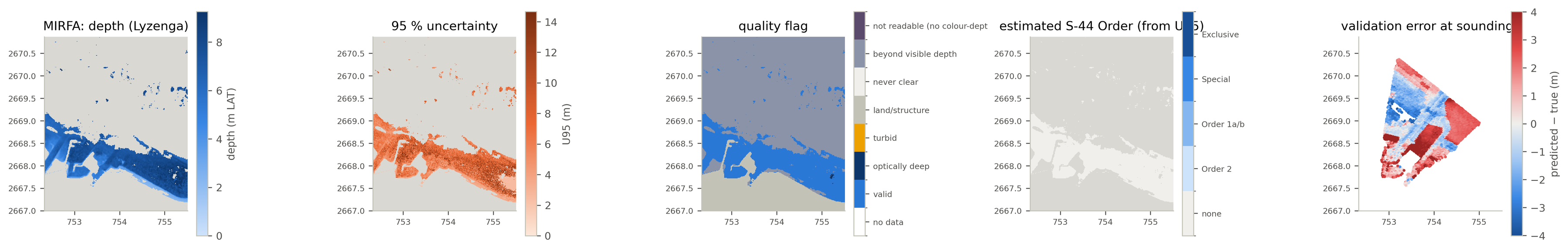}\\[1mm]
\includegraphics[width=\textwidth]{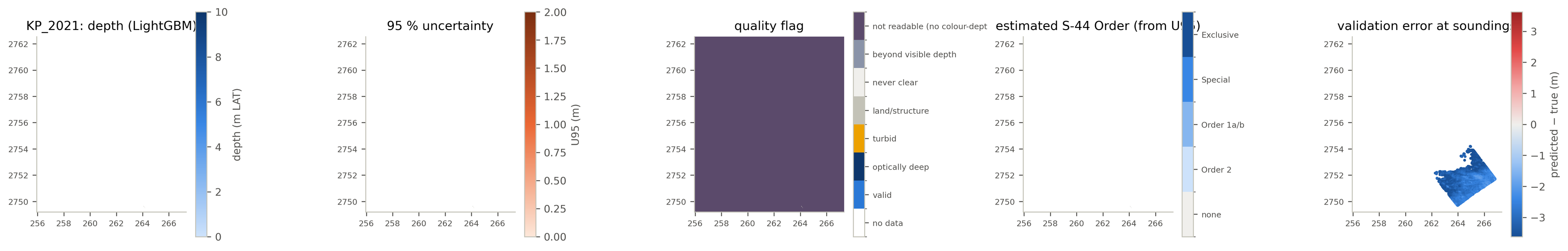}\\[1mm]
\includegraphics[width=\textwidth]{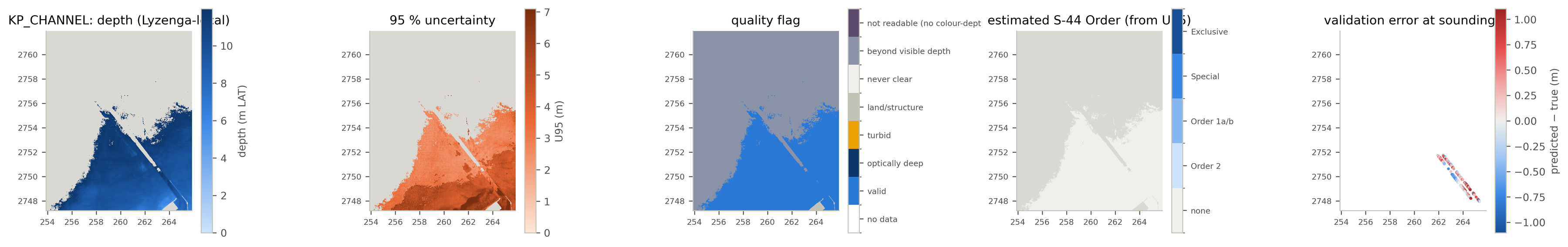}\\[1mm]
\includegraphics[width=\textwidth]{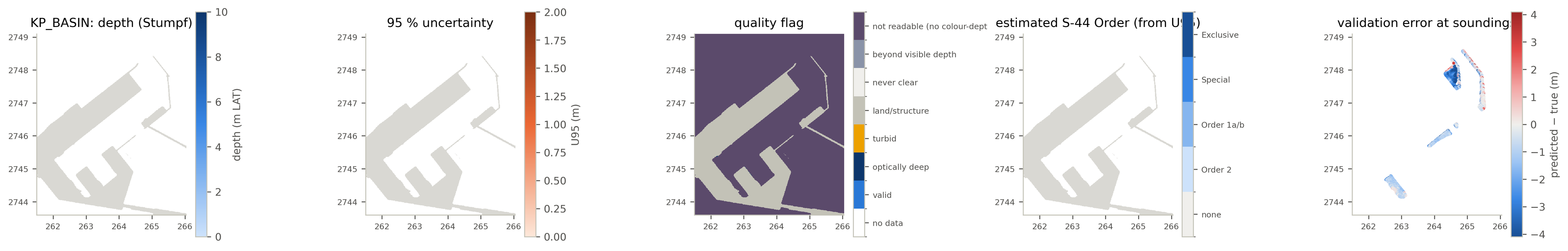}\\[1mm]
\includegraphics[width=\textwidth]{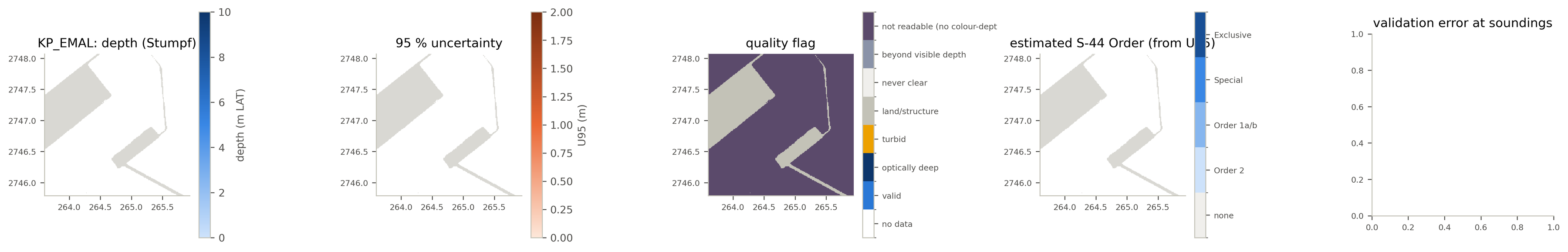}\\[1mm]
\includegraphics[width=\textwidth]{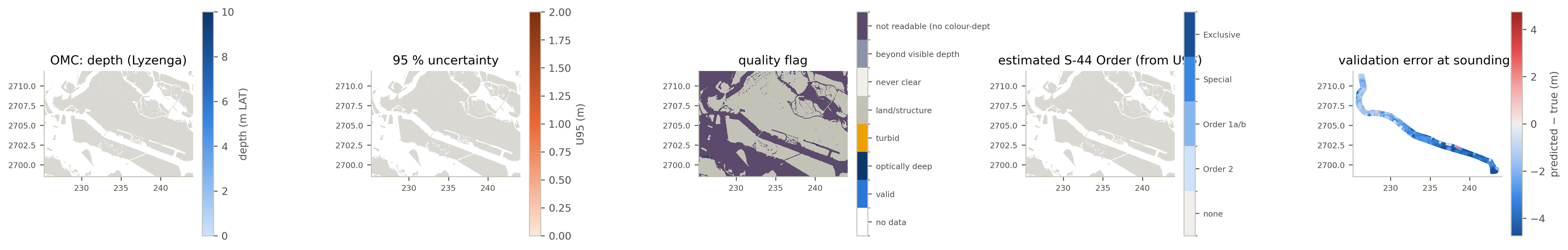}
\caption{Full product panels, part 2: MIRFA, KP\_CHANNEL, and the withheld or empty maps of KP\_2021, KP\_BASIN, KP\_EMAL and OMC, where the quality flag shows why no depth is published.}
\label{fig:panels2}
\end{figure}

This appendix holds the figures referred to in the main text that support the data selection and the validation design: the timeline of the Sentinel-2 images examined and kept for each survey (Fig.~\ref{fig:timeline}), the semivariogram from which the 450\,m block size was derived (Fig.~\ref{fig:variogram}), the comparison of the EOT20 tide model with the survey gauges (Fig.~\ref{fig:tidecheck}) and the full product panels of the twelve datasets (Figs.~\ref{fig:panels} and \ref{fig:panels2}).

\clearpage
\section{Site-level results of the seasonal experiment}
\label{sec:supp_seasons}

Table~\ref{tab:seasons_site} gives, for each of the seven sites of the seasonal experiment (Section~\ref{sec:seasons}), the turbidity index of the images of each season, the colour--depth rank correlation, the Stumpf transfer error, and the error and coverage of the hybrid with local control. The pooled values are given in the main text.

\begin{table}[htbp]
\centering
\caption{Seasonal experiment per site: turbidity index of the three images kept, colour--depth rank correlation $\rho$, and held-out RMSE (m) of the Stumpf transfer and Hybrid-local methods with the Hybrid-local coverage (\%). ``--'': no image could be calibrated.}
\label{tab:seasons_site}
\small
\begin{adjustbox}{max width=\textwidth}
\begin{tabular}{@{}llrrrrr@{}}
\toprule
Site & Season & Turbidity & $\rho$ & Stumpf RMSE & Hybrid-local RMSE & Coverage \\
\midrule
JD & Winter & 0.003 & 0.73 & 1.38 & 0.85 & 98.4 \\
JD & Spring & 0.005 & 0.82 & 1.40 & 0.59 & 99.4 \\
JD & Summer & 0.017 & 0.58 & 1.57 & 0.85 & 84.3 \\
JD & Autumn & 0.009 & 0.67 & 1.82 & 1.34 & 96.9 \\
MGP\_NC & Winter & 0.003 & 0.45 & 1.00 & 0.71 & 99.6 \\
MGP\_NC & Spring & 0.004 & 0.33 & 1.04 & 0.83 & 83.4 \\
MGP\_NC & Summer & 0.012 & 0.26 & 1.50 & 0.78 & 69.9 \\
MGP\_NC & Autumn & 0.007 & 0.54 & 1.24 & 0.85 & 98.7 \\
SBY\_LCB & Winter & 0.001 & 0.83 & 1.63 & 1.69 & 37.1 \\
SBY\_LCB & Spring & 0.001 & 0.74 & 1.58 & 1.48 & 93.2 \\
SBY\_LCB & Summer & 0.031 & 0.63 & 1.91 & 2.06 & 95.6 \\
SBY\_LCB & Autumn & 0.005 & 0.64 & 1.83 & 1.34 & 98.1 \\
SBY\_LP & Winter & 0.008 & 0.77 & 1.26 & 1.12 & 99.8 \\
SBY\_LP & Spring & 0.004 & 0.80 & 1.17 & 1.25 & 97.7 \\
SBY\_LP & Summer & 0.031 & 0.45 & 4.82 & 1.78 & 50.6 \\
SBY\_LP & Autumn & 0.002 & 0.67 & 1.84 & 1.18 & 100.0 \\
MIRFA & Winter & 0.002 & 0.68 & 2.81 & 1.85 & 100.0 \\
MIRFA & Spring & 0.003 & 0.60 & 2.66 & 1.92 & 93.6 \\
MIRFA & Summer & 0.007 & 0.42 & 3.05 & 3.01 & 57.4 \\
MIRFA & Autumn & 0.002 & 0.47 & 2.63 & -- & 0 \\
OMC & Winter & 0.008 & 0.13 & 4.48 & -- & 0 \\
OMC & Spring & 0.016 & 0.18 & 3.08 & -- & 0 \\
OMC & Summer & 0.029 & 0.18 & 2.57 & -- & 0 \\
OMC & Autumn & 0.018 & 0.20 & 3.04 & -- & 0 \\
KP\_2021 & Winter & 0.000 & 0.40 & 1.97 & 0.46 & 80.7 \\
KP\_2021 & Spring & 0.002 & 0.46 & 3.24 & 0.55 & 100.0 \\
KP\_2021 & Summer & 0.002 & 0.41 & 3.83 & 0.55 & 100.0 \\
KP\_2021 & Autumn & 0.000 & 0.33 & 3.75 & 0.43 & 100.0 \\
\bottomrule
\end{tabular}
\end{adjustbox}
\end{table}

\clearpage
\section{Coast-wide feasibility per cell}
\label{sec:supp_coast}

Table~\ref{tab:coastcells} lists the 49 coastal cells of the feasibility experiment (Section~\ref{sec:coast}) with their water area, number of clear months in 2024, median turbidity index, optically shallow and mapped areas, and the median reconnaissance depth and uncertainty of the mosaic.

\begin{table}[htbp]
\centering
\caption{Coast-wide feasibility of SDB per $0.25\degree$ coastal cell in 2024 (cell centre; water, optically shallow area by the signal-to-noise and by the spread test, and mapped area in km$^2$; months with a clear image; median turbidity index; median reconnaissance depth and site-conformal U95).}
\label{tab:coastcells}
\scriptsize
\begin{adjustbox}{max width=\textwidth}
\begin{tabular}{@{}lrrrrrrrrrr@{}}
\toprule
Cell & Lon (\degree E) & Lat (\degree N) & Water & Shallow (SNR) & Shallow ($\sigma_\infty$) & Mapped & Clear months & Turbidity & Depth (m) & U95 (m) \\
\midrule
c000 & 51.675 & 24.075 & 48 & 48 & 45 & 45 & 8 & 0.005 & 6.8 & 4.7 \\
c001 & 51.925 & 24.075 & 603 & 600 & 578 & 578 & 9 & 0.005 & 8.2 & 5.1 \\
c002 & 52.175 & 24.075 & 667 & 656 & 646 & 646 & 10 & 0.003 & 8.7 & 5.7 \\
c003 & 52.425 & 24.075 & 395 & 395 & 383 & 383 & 12 & 0.002 & 8.4 & 4.3 \\
c004 & 52.675 & 24.075 & 126 & 124 & 118 & 118 & 9 & 0.005 & 5.7 & 5.2 \\
c005 & 52.925 & 24.075 & 171 & 168 & 151 & 151 & 6 & 0.012 & 3.9 & 2.8 \\
c006 & 53.175 & 24.075 & 204 & 202 & 195 & 195 & 9 & 0.002 & 3.8 & 3.3 \\
c007 & 53.425 & 24.075 & 269 & 268 & 259 & 259 & 10 & 0.002 & 5.3 & 3.8 \\
c008 & 53.675 & 24.075 & 308 & 306 & 297 & 297 & 10 & 0.002 & 4.0 & 3.2 \\
c009 & 53.925 & 24.075 & 200 & 197 & 179 & 179 & 0 & 0.054 & 2.6 & 2.7 \\
c010 & 54.175 & 24.075 & 11 & 10 & 10 & 10 & 0 & 0.152 & 3.7 & 2.4 \\
c020 & 51.675 & 24.325 & 507 & 499 & 473 & 450 & 6 & 0.009 & 8.0 & 3.2 \\
c024 & 52.675 & 24.325 & 631 & 629 & 618 & 618 & 10 & 0.003 & 7.9 & 4.8 \\
c026 & 53.175 & 24.325 & 696 & 695 & 670 & 670 & 10 & 0.002 & 6.6 & 5.8 \\
c027 & 53.425 & 24.325 & 694 & 694 & 668 & 668 & 8 & 0.006 & 5.0 & 3.8 \\
c028 & 53.675 & 24.325 & 686 & 685 & 680 & 680 & 10 & 0.002 & 6.7 & 5.3 \\
c029 & 53.925 & 24.325 & 660 & 655 & 625 & 625 & 6 & 0.006 & 5.8 & 3.5 \\
c030 & 54.175 & 24.325 & 407 & 406 & 380 & 380 & 4 & 0.030 & 4.4 & 5.3 \\
c031 & 54.425 & 24.325 & 204 & 202 & 189 & 189 & 0 & 0.042 & 3.2 & 3.7 \\
c040 & 51.675 & 24.575 & 122 & 121 & 114 & 98 & 12 & 0.001 & 8.4 & 5.2 \\
c043 & 52.425 & 24.575 & 693 & 689 & 652 & 288 & 11 & 0.002 & 10.3 & 3.1 \\
c051 & 54.425 & 24.575 & 531 & 530 & 487 & 487 & 7 & 0.005 & 6.0 & 5.2 \\
c052 & 54.675 & 24.575 & 158 & 157 & 150 & 150 & 0 & 0.035 & 3.1 & 3.4 \\
c059 & 56.425 & 24.575 & 36 & 33 & 35 & 33 & 7 & 0.000 & 7.0 & 6.1 \\
c060 & 51.675 & 24.825 & 4 & 4 & 4 & 3 & 10 & 0.000 & 10.0 & 2.5 \\
c063 & 52.425 & 24.825 & 719 & 716 & 651 & 647 & 10 & 0.001 & 8.5 & 6.4 \\
c072 & 54.675 & 24.825 & 515 & 515 & 492 & 491 & 10 & 0.003 & 7.4 & 4.9 \\
c073 & 54.925 & 24.825 & 81 & 81 & 80 & 80 & 9 & 0.003 & 6.3 & 3.6 \\
c079 & 56.425 & 24.825 & 322 & 321 & 322 & 320 & 9 & 0.002 & 4.9 & 9.4 \\
c080 & 51.675 & 25.075 & 515 & 514 & 502 & 502 & 7 & 0.004 & 6.9 & 4.2 \\
c083 & 52.425 & 25.075 & 716 & 708 & 680 & 653 & 12 & 0.001 & 9.6 & 6.2 \\
c099 & 56.425 & 25.075 & 521 & 420 & 517 & 371 & 8 & 0.002 & 9.9 & 2.7 \\
c100 & 51.675 & 25.325 & 630 & 630 & 624 & 623 & 7 & 0.003 & 5.1 & 5.5 \\
c114 & 55.175 & 25.325 & 655 & 649 & 637 & 637 & 8 & 0.003 & 7.9 & 5.7 \\
c115 & 55.425 & 25.325 & 168 & 168 & 165 & 165 & 6 & 0.012 & 5.9 & 4.6 \\
c119 & 56.425 & 25.325 & 513 & 512 & 513 & 281 & 8 & 0.003 & 10.5 & 1.7 \\
c121 & 51.925 & 25.575 & 181 & 176 & 140 & 135 & 12 & 0.001 & 8.0 & 4.8 \\
c135 & 55.425 & 25.575 & 663 & 662 & 661 & 660 & 12 & 0.002 & 8.5 & 5.6 \\
c136 & 55.675 & 25.575 & 225 & 225 & 224 & 224 & 8 & 0.004 & 5.7 & 4.2 \\
c138 & 56.175 & 25.575 & 26 & 26 & 26 & 25 & 5 & 0.007 & 5.9 & 7.1 \\
c139 & 56.425 & 25.575 & 597 & 430 & 594 & 380 & 8 & 0.002 & 9.1 & 2.7 \\
c140 & 51.675 & 25.825 & 589 & 587 & 574 & 573 & 10 & 0.002 & 6.8 & 3.2 \\
c141 & 51.925 & 25.825 & 12 & 11 & 0 & 0 & 11 & 0.000 & 5.5 & 2.7 \\
c144 & 52.675 & 25.825 & 721 & 708 & 718 & 634 & 11 & 0.001 & 8.8 & 10.6 \\
c154 & 55.175 & 25.825 & 719 & 697 & 716 & 507 & 10 & 0.002 & 10.2 & 2.8 \\
c157 & 55.925 & 25.825 & 447 & 438 & 433 & 432 & 9 & 0.003 & 6.3 & 4.4 \\
c158 & 56.175 & 25.825 & 11 & 11 & 11 & 11 & 7 & 0.004 & 7.3 & 5.8 \\
c159 & 56.425 & 25.825 & 522 & 520 & 520 & 519 & 7 & 0.004 & 8.8 & 4.9 \\
c178 & 56.175 & 26.075 & 165 & 163 & 163 & 162 & 10 & 0.005 & 7.4 & 4.2 \\
\bottomrule
\end{tabular}
\end{adjustbox}
\end{table}

\clearpage
\section{Lidar calibration of the coast product per cell}
\label{sec:supp_coastcal}

Table~\ref{tab:coastcalcells} lists, for each coastal cell, the lidar pixels available, the depth limit derived from the lidar, the number of 480\,m blocks of the held-out test, the RMSE and bias of the mosaic against the held-out lidar before and after calibration, the weighted $R^2$ of the calibration, the mapped area of the calibrated product and the status of the cell (Section~\ref{sec:coastcal}).

\begin{table}[htbp]
\centering
\caption{Lidar calibration of the coast product per $0.25\degree$ cell: lidar pixels (60\,m), depth limit from the lidar (m), blocks of the held-out test, held-out RMSE and bias (m) before (transfer) and after calibration, weighted $R^2$ of the calibration, mapped area of the calibrated product (km$^2$) and status.}
\label{tab:coastcalcells}
\scriptsize
\begin{adjustbox}{max width=\textwidth}
\begin{tabular}{@{}lrrrrrrrrrrrl@{}}
\toprule
Cell & Lon & Lat & Lidar px & Limit & Blocks & RMSE$_\mathrm{tr}$ & Bias$_\mathrm{tr}$ & RMSE$_\mathrm{cal}$ & Bias$_\mathrm{cal}$ & $R^2$ & Mapped & Status \\
\midrule
c000 & 51.67 & 24.07 & 303 & 4.3 & 30 & 4.31 & 4.29 & 0.63 & 0.34 & 0.61 & 18 & calibrated \\
c001 & 51.92 & 24.07 & 2728 & 10.0 & 270 & 2.91 & 2.58 & 0.74 & 0.05 & 0.89 & 232 & calibrated \\
c002 & 52.17 & 24.07 & 2567 & 11.0 & 331 & 1.85 & 1.15 & 0.79 & 0.05 & 0.90 & 278 & calibrated \\
c003 & 52.42 & 24.07 & 4204 & 11.0 & 384 & 2.78 & 2.38 & 0.86 & 0.09 & 0.91 & 338 & calibrated \\
c004 & 52.67 & 24.07 & 1122 & 6.0 & 139 & 2.70 & 2.26 & 1.63 & 0.18 & 0.19 & -- & fit too weak ($R^2<0.3$) \\
c005 & 52.92 & 24.07 & 485 & 3.0 & 57 & 3.24 & 3.17 & 0.75 & 0.10 & 0.54 & 147 & calibrated \\
c006 & 53.17 & 24.07 & 1916 & 10.0 & 306 & 2.45 & 1.23 & 1.20 & 0.03 & 0.87 & 195 & calibrated \\
c007 & 53.42 & 24.07 & 4629 & 9.0 & 487 & 2.80 & 2.30 & 1.05 & 0.36 & 0.82 & 252 & calibrated \\
c008 & 53.67 & 24.07 & 4177 & 6.0 & 426 & 2.41 & 2.18 & 1.03 & 0.22 & 0.59 & 295 & calibrated \\
c009 & 53.92 & 24.07 & 936 & 11.0 & 170 & -- & -- & -- & -- & 0.00 & -- & wrong slope sign \\
c010 & 54.17 & 24.07 & 11 & -- & -- & -- & -- & -- & -- & -- & -- & too few lidar pixels \\
c020 & 51.67 & 24.32 & 1573 & 11.0 & 253 & 2.67 & 2.24 & 1.28 & 0.07 & 0.80 & 327 & calibrated \\
c024 & 52.67 & 24.32 & 3555 & 11.0 & 546 & 3.28 & 2.54 & 2.90 & 1.35 & 0.35 & 470 & calibrated \\
c026 & 53.17 & 24.32 & 7902 & 11.0 & 929 & 3.28 & 2.41 & 1.77 & 0.30 & 0.70 & 650 & calibrated \\
c027 & 53.42 & 24.32 & 18077 & 11.0 & 1376 & 3.27 & 2.72 & 1.12 & 0.16 & 0.84 & 649 & calibrated \\
c028 & 53.67 & 24.32 & 4984 & 11.0 & 627 & 3.19 & 1.94 & 1.58 & 0.20 & 0.81 & 628 & calibrated \\
c029 & 53.92 & 24.32 & 5906 & 11.0 & 739 & 2.56 & 1.30 & 1.06 & 0.11 & 0.90 & 544 & calibrated \\
c030 & 54.17 & 24.32 & 4999 & 11.0 & 651 & 3.11 & 2.24 & 1.73 & 0.52 & 0.74 & 372 & calibrated \\
c031 & 54.42 & 24.32 & 2203 & 3.0 & 326 & -- & -- & -- & -- & 0.25 & -- & fit too weak ($R^2<0.3$) \\
c040 & 51.67 & 24.57 & 330 & 11.0 & 39 & 3.22 & 2.84 & 3.13 & -2.35 & 0.13 & -- & fit too weak ($R^2<0.3$) \\
c043 & 52.42 & 24.57 & 112 & 11.0 & 17 & 2.32 & 1.78 & 1.79 & -1.16 & 0.55 & 288 & calibrated \\
c051 & 54.42 & 24.57 & 8240 & 11.0 & 946 & 2.52 & 0.75 & 1.75 & 0.24 & 0.73 & 429 & calibrated \\
c052 & 54.67 & 24.57 & 3736 & 5.0 & 387 & 3.01 & 2.91 & 0.87 & 0.36 & 0.57 & 150 & calibrated \\
c059 & 56.42 & 24.57 & 43 & -- & -- & -- & -- & -- & -- & -- & -- & too few lidar pixels \\
c060 & 51.67 & 24.82 & 0 & -- & -- & -- & -- & -- & -- & -- & -- & too few lidar pixels \\
c063 & 52.42 & 24.82 & 376 & 11.0 & 48 & -- & -- & -- & -- & 0.00 & -- & wrong slope sign \\
c072 & 54.67 & 24.82 & 10068 & 11.0 & 891 & 2.60 & 1.31 & 1.90 & 0.35 & 0.68 & 373 & calibrated \\
c073 & 54.92 & 24.82 & 1744 & 9.0 & 157 & 1.72 & 0.82 & 1.29 & -0.07 & 0.76 & 79 & calibrated \\
c079 & 56.42 & 24.82 & 164 & 11.0 & 47 & 2.69 & -1.44 & 1.71 & 0.43 & 0.76 & 216 & calibrated \\
c080 & 51.67 & 25.07 & 898 & 11.0 & 169 & 3.26 & 2.15 & 1.27 & 0.12 & 0.88 & 460 & calibrated \\
c083 & 52.42 & 25.07 & 1032 & 6.0 & 127 & 4.24 & 3.93 & 1.21 & -0.36 & 0.39 & 653 & calibrated \\
c099 & 56.42 & 25.07 & 26 & -- & -- & -- & -- & -- & -- & -- & -- & too few lidar pixels \\
c100 & 51.67 & 25.32 & 1199 & 7.0 & 170 & 2.15 & 1.31 & 1.95 & 0.89 & 0.28 & -- & fit too weak ($R^2<0.3$) \\
c114 & 55.17 & 25.32 & 46 & -- & -- & -- & -- & -- & -- & -- & -- & too few lidar pixels \\
c115 & 55.42 & 25.32 & 1 & -- & -- & -- & -- & -- & -- & -- & -- & too few lidar pixels \\
c119 & 56.42 & 25.32 & 37 & -- & -- & -- & -- & -- & -- & -- & -- & too few lidar pixels \\
c121 & 51.92 & 25.57 & 21 & -- & -- & -- & -- & -- & -- & -- & -- & too few lidar pixels \\
c135 & 55.42 & 25.57 & 0 & -- & -- & -- & -- & -- & -- & -- & -- & too few lidar pixels \\
c136 & 55.67 & 25.57 & 0 & -- & -- & -- & -- & -- & -- & -- & -- & too few lidar pixels \\
c138 & 56.17 & 25.57 & 3 & -- & -- & -- & -- & -- & -- & -- & -- & too few lidar pixels \\
c139 & 56.42 & 25.57 & 23 & -- & -- & -- & -- & -- & -- & -- & -- & too few lidar pixels \\
c140 & 51.67 & 25.82 & 1187 & 11.0 & 181 & 2.01 & 0.62 & 1.15 & 0.26 & 0.83 & 541 & calibrated \\
c141 & 51.92 & 25.82 & 0 & -- & -- & -- & -- & -- & -- & -- & -- & too few lidar pixels \\
c144 & 52.67 & 25.82 & 256 & 11.0 & 55 & -- & -- & -- & -- & 0.00 & -- & wrong slope sign \\
c154 & 55.17 & 25.82 & 23 & -- & -- & -- & -- & -- & -- & -- & -- & too few lidar pixels \\
c157 & 55.92 & 25.82 & 6 & -- & -- & -- & -- & -- & -- & -- & -- & too few lidar pixels \\
c158 & 56.17 & 25.82 & 1 & -- & -- & -- & -- & -- & -- & -- & -- & too few lidar pixels \\
c159 & 56.42 & 25.82 & 28 & -- & -- & -- & -- & -- & -- & -- & -- & too few lidar pixels \\
c178 & 56.17 & 26.07 & 41 & -- & -- & -- & -- & -- & -- & -- & -- & too few lidar pixels \\
\bottomrule
\end{tabular}
\end{adjustbox}
\end{table}

\end{document}